# What Balanced Accuracy Cannot Identify in Neural-Signal Decoding under Test-Time Adaptation

Jiangshuo Xie Zonghan Wu

*Xiamen University Malaysia*

## Abstract

**Objective.** Adaptation for neural-signal classification is usually transductive: each batch is scored after updating on it, and the gain depends on the adapted model [4]. This paper asks what an aggregate change fails to characterise inside the cohort.

**Approach.** Full test-time adaptation (`full_ttt`) was evaluated in 11 dataset–encoder cells (four each on Siena/EEG and PTB-XL/ECG, three on Sleep-EDF/EEG), with splits and source-model evaluation fixed. The subject was the unit; exact sign tests were Holm-corrected and conditional on the shared stream, with bootstrap intervals and a dependence audit.

**Main results.** This paper reports two results of different kinds. (i) *Evaluation*: in the binary case balanced accuracy gives a one-class decoder the same value as one reporting both classes at 50%, so the aggregate cannot identify a class that stopped being reported. A class covering a tenth of a subject's samples went unpredicted in 26–47 of 112 subject-cells, against 7–9 for the controls; this is a descriptive count. (ii) *Comparison*: with cohort, split and update rule fixed, the within-subject class contrast follows the encoder. This study tested it directly instead of inferring it from tests against zero: two of six Siena encoder pairs survive the dependence readings at their nominal reference and one at a fold-level reference, while two of ten Sleep-EDF stage pairs pass the run-clustered gate at its nominal reference. Aggregate and contrast named different encoders in 10 of 11. The protocol bounds both results, since a stream-interleaving effect reverses sign with state sharing (+3.5 to −2.4 pp).

**Significance.** An aggregate score can hide individual harm, as [1] showed, but it cannot distinguish a uniform loss from a deleted class. Class-level change should therefore be reported alongside the aggregate and subject-level change alongside it; both results remain conditional on the cell, the rule and the stream construction.

## 1 Introduction

Neural signals are increasingly classified online. Sleep staging scores a night of EEG as it arrives, seizure detectors monitor scalp recordings continuously, and motor-imagery decoders turn cortical activity into control signals. In each case the model meets a stream it was not trained on: subjects differ in anatomy, electrode placement and state, and recording conditions drift. Test-time adaptation (TTA) is a common response: the model is updated on unlabeled incoming batches so that it tracks the stream [2, 5, 7]. No labels are needed, but an update that helps on average can still hurt particular subjects or particular classes, and an online system has no held-out set against which to check this.

Evaluations usually summarise adaptation with one aggregate: BA before and after, averaged over a test set. That answers whether the model improved, not who improved, and a cohort mean can conceal individual harm: in episodic segmentation the mean change was indistinguishable from zero while 58.7% of cases were individually worse [1]. Per-class scores have also been reported, for sleep staging among others [2]. Neither settles the structure of that concealment when the decoder keeps updating along one stream instead of being reset between cases. A cohort-level per-class score reports the average level of each class, and the difference between two of them gives the average contrast. It cannot say how much of that contrast the aggregate change already accounts for, nor how consistent the contrast is across subjects (§4.3.6). Earlier work has documented collapse [3], sensitivity to stream construction [1, 6], the pitfalls of TTA benchmarking [2, 3], the failure of adaptation gains to transfer across clinical shifts at the dataset level [4], and label-revealed online updates in EEG under a time-ordered protocol [5]. Nor is the design new: [4] already crosses four foundation models with three adaptation methods on five datasets at the dataset level of balanced accuracy.

This paper therefore asks one question about EEG-based classification in two parts: with the cohort, the partition and the adaptation protocol held fixed, does a class that is present in the stream still receive no prediction, and does the direction of the within-subject class contrast survive a change of encoder? Two properties of the run set the conditions under which either answer holds: whether adaptation state is shared across subjects, and where along the stream the change happens. PTB-XL, an ECG dataset, enters as a cross-modal comparison; the neural evidence comes from Siena and Sleep-EDF, with BCI-IV-2a and CHB-MIT as boundary cases. A secondary question — whether unlabeled predictions can stand in for the contrast — is answered within the second result as a label-free version of the same decision (§4.3.7). Whether those predictions reveal deterioration before it is scored is a check on the evaluation, not a contribution of its own (S20).

The paper rests on two results of different kinds. The first is a failure mode rather than a contrast: the adapted decoder omits represented classes from its predictions. Consider a class that made up at least a tenth of a subject's samples. It received no prediction at all, in all four seeds, in 44 of the 112 subject-cells whose ground truth contains two such classes when all parameters were updated. The count is 47 under entropy minimisation of the normalisation affine parameters and 26 under information maximisation with a diversity term; the source model did so in 9 of the same cells and an adapter on a frozen backbone in 7. The unevenness is part of the result: Sleep-EDF with CNN deletes a class for all 20 subjects, with EEGNet for 13 and with Mamba for 1; Siena with MedFormer for 6 of 13, with CNN for 4, and for none with EEGNet or Mamba. On Sleep-EDF the class that disappears is the short one: REM recall falls to 0.00% under CNN, 10.5% under EEGNet and 10.9% under Mamba, while the share of predictions given to N3 rises to 45.0%, 62.9% and 59.5% (§4.1). This paper claims no mechanism here, only that the decoder no longer reports what it was trained to report.

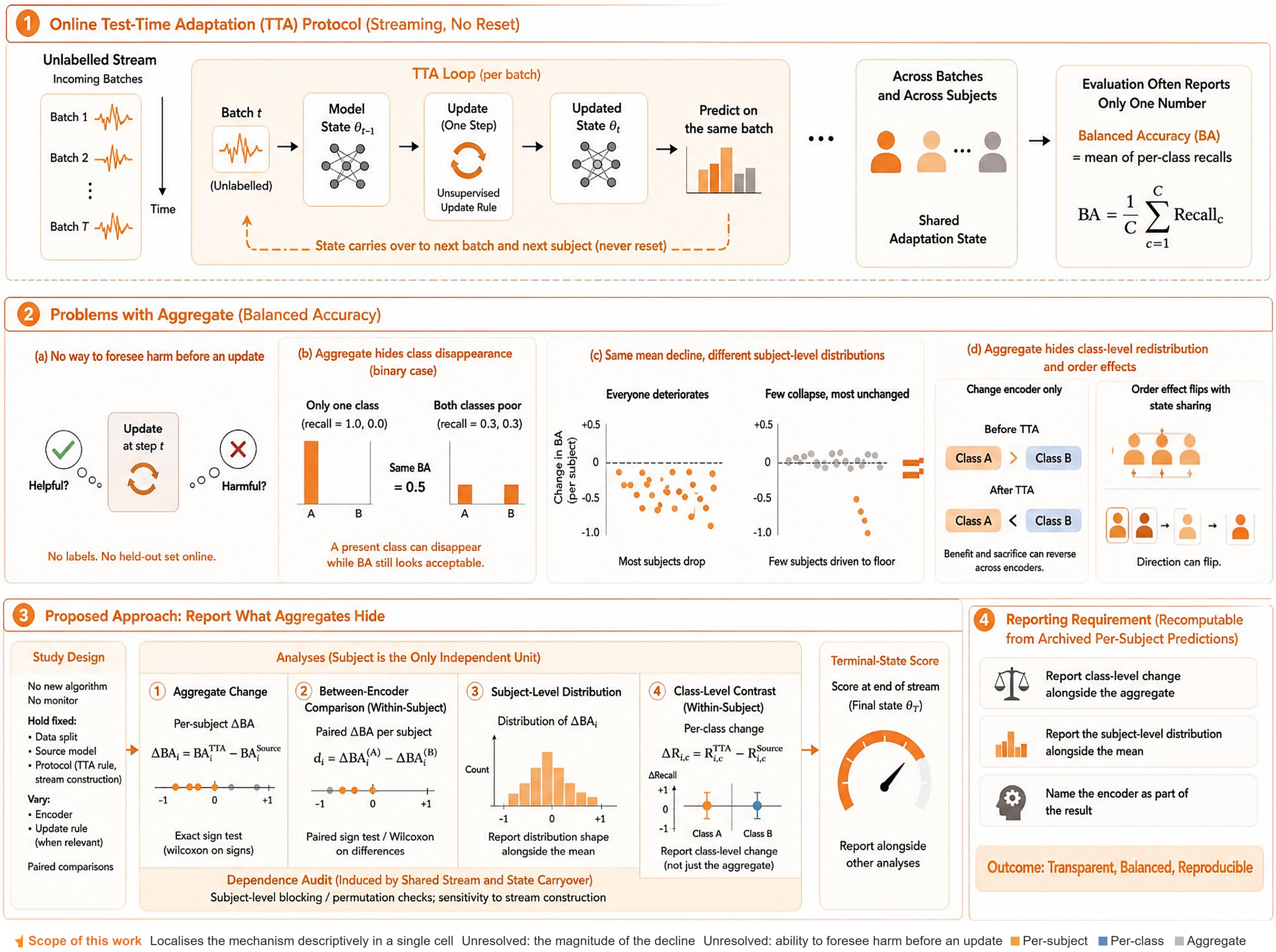

**Figure 1. What a single aggregate score cannot identify under streaming test-time adaptation.** (A) Online TTA applies one unsupervised update to each unlabelled batch and scores that same batch with the updated model; the state carries across batches and subjects without a reset, and evaluation often reports one number, balanced accuracy (BA), the unweighted mean of per-class recalls. (B) Four limits of that aggregate: no labels or held-out set are available to foresee a harmful update; a decoder reporting one class and one reporting both classes at 50% receive the same BA; the same mean change is compatible with a broad decline and with a few subjects driven to the floor; and class-level redistribution and the sign of a stream-order effect can reverse with the encoder or with state sharing. (C) The readouts reported here, recomputable from archived per-subject predictions, keep the subject as the only independent unit: the within-subject aggregate change, the between-encoder contrast, the subject-level distribution of changes and the within-subject class-level contrast, accompanied by a terminal-state score and a dependence audit. The analyses group into performance/subject, class-structure and decision/stream evidence families; the requirement is to report class-level change alongside the aggregate, the subject-level distribution alongside the mean, and the encoder as part of the result. Colour encodes per-subject, per-class and aggregate quantities, and the footer marks what remains unresolved. Design specification (§3.2–§3.4): five datasets enter subject-wise cross-validation and each seed defines a new split and refitted source model; four encoders give three datasets × four backbones, minus the memory-limited MedFormer × Sleep-EDF cell, which is omitted rather than scored as zero, with BCI-IV-2a a boundary extension and CHB-MIT a monitoring/low-skill extension. Under `full_ttt`, all weights are updated with masked-feature reconstruction (mask ratio 0.3; span 16); `h0` is held at zero throughout; the streaming score concatenates post-update predictions, and the terminal-state score re-evaluates the final model on the full test set for four cells in S11. Paired orders share the recorded test-subject lists, the source-only evaluation and one cached source model; the adapted weights of the paired-order runs were not hashed. All graphics are schematic, not raw data.

The second is operational: balanced accuracy does not separate that failure from a uniform loss. It is the mean of per-class recalls, so a binary model that emits one class scores exactly 50 — the same as a model scoring 50% on both classes. The class contrast does separate them, though it has costs: its direction is encoder-dependent, the aggregate and the contrast name the same encoder in only 1 of 11 comparisons, and it needs labels. A label-free surrogate, the change in predicted class share, recovers its choice in 8 of 10 unbalanced five-class pairs against 1 of 10 for the aggregate (§4.3.6–4.3.7).

Both results are bounded by the protocol: the sign of a stream-interleaving effect depends on whether adaptation state is shared across subjects, so an order effect measured under one construction does not transfer to the other (§4.5, S7–S8), which is a boundary on the measurement, not a third finding. Both rest on the subject-level distribution, and §5 sets out what that distribution licenses.

Neither result is a new phenomenon; they are new as an identifiability limit and as an empirical dependence. A mean can hide individual cases [1], collapse to a single predicted class is a documented failure mode of entropy-based adaptation [25], and class-level loss under adaptation is documented [23, 24]. What the literature does not provide is a per-subject count of classes present in the stream but never predicted, nor the within-subject class contrast compared across encoders under a fixed protocol; both counts can be checked against the archived predictions (§4.1, §4.3.3–4.3.5, S11). Subject-level heterogeneity is not claimed as new, and this paper proposes no new adaptation algorithm. The boundary analyses of S14–S17 ask where in the model the change sits and are reported as boundaries, not as a fourth contribution. Figure 1 frames the evaluation problem and the readouts reported here; tables and figures are numbered continuously, with Tables 1–6 and Figures 1–3 here and Tables 7–19 and Figures 4–9 in the supplement.

# 2 Related work

## 2.1 Evaluating test-time adaptation

Test-time adaptation results depend on the model, data distribution and experimental setup. TTAB [6] compared ten methods under one protocol and found that their ranking changed with hyperparameters selected without labels, source-model quality and shift type. A gain therefore reflects the configuration as well as the method. Computational budget matters too: Alfarra et al. [7] found that time limits can change the ranking.

Reported scores also depend on the unit over which they are averaged. Wang et al. [1] adapted a segmentation model to each case in turn and found the cohort mean ΔDice indistinguishable from zero while 58.7% of cases were individually made worse. Their remedy is a label-free per-case statistic, prediction fragmentation — the disagreement geometry between the source prediction and the adapted one — which ranks that harm well enough to route around most of it. That establishes that a cohort mean can conceal per-case harm, but it does so in episodic segmentation, where each case is an independent input and adaptation restarts from the source weights. Whether the same concealment holds for a decoder that keeps updating along one continuous stream, and whether what is concealed is structured or diffuse, is a different question taken up in §4.1 and §4.2.

Class-level effects of adaptation are also established in vision. Under a shifted label distribution, entropy minimisation drives the normalisation shift parameter toward the over-represented class, so that classes absent from the target stream lose accuracy while the observed ones hold; that failure has been named partial forgetting, and the aggregate in that line of work includes the unobserved classes so that it registers them [23]. Confusion structure has been mapped as well: under label distribution shift the adapted classifier's errors concentrate in identifiable class pairs and head classes lose the most accuracy [24]. Neither reports the case measured here: in the first the failing class is absent from the adaptation stream, and in the second the class-level loss accompanies an aggregate loss. Here the class is present in the stream, covers at least a tenth of the subject's samples, and still receives no prediction at all, while in the binary case balanced accuracy gives that decoder the same value as one that reports both classes badly (§4.1).

Collapse is a separate line of work: Han et al. [3] report abrupt collapse when adaptation tasks arrive in a particular sequence. The setting used here instead permutes samples within a single test set, and S19 asks whether reordering alone can trigger collapse. Wang et al. [8] align the adapted distribution with the source distribution, making the outcome depend on the model's predicted distribution and on how the test data are organised.

## 2.2 Test sample order and online adaptation

In online adaptation, sample order may provide useful temporal evidence or confound evaluation. OATTA [9] learns a transition matrix over consecutive predictions and applies a likelihood ratio to decide whether that evidence warrants action. Jhawar and Wang [10] frame the issue as one of identifiability: if two deployments look the same through the observable channel but produce different outcomes, no rule based on that channel can distinguish them.

In EEG decoding, Kozyrskiy et al. [5] benchmarked label-revealed online updates — predict, then train on the revealed label — under a time-ordered prequential protocol [11], across four datasets and two pipeline families, and found gains in 13 of 14 model–dataset pairs on the two largest streams, up to about 18% relative. Two differences separate their study from this one. Their updates consume the true label, so the measured quantity is what a decoder gains when the label arrives, whereas the arms used here are label-free; their endpoint is also a dataset-level score, which does not say which subjects or classes moved.

This paper does not claim to be the first to study order sensitivity: [10] already compares deployment structures for fixed sample sets, with a different decision endpoint. The primary question is the direction of the within-subject class contrast and whether it survives a change of encoder. The paired stream-organisation contrasts are supporting evidence: they hold the test-subject set and recorded source evaluation fixed while changing adjacency, batch composition and boundaries together, so they do not isolate a single temporal-order mechanism.

### 2.3 Test-time adaptation on neural signals

NeuroAdapt-Bench [4] evaluated adaptation methods on EEG foundation models across shifts in clinical setting, recording device and modality, crossing four foundation models with three adaptation methods on five datasets while holding the split and the adaptation hyperparameters fixed. Gains did not transfer; gradient-based methods degraded often, and optimisation-free methods stayed near where they started. Of 180 comparisons, 145 were significant (128 degradations and 17 improvements), and all of it is reported at the dataset level, with no per-subject and no per-class breakdown. Per-stage information has also been reported for sleep staging [2], but as a confusion matrix for one pipeline after adaptation, without the frozen-baseline counterpart, so it is a single snapshot rather than a contrast across encoders.

On sleep staging, Arasu and Jahangiri [2] report consistent gains over a frozen baseline on Sleep-EDF Expanded, using entropy minimisation with batch-normalisation statistic refresh and two safety rails. Their result is not directly comparable with the present one, since the objective, the trainable parameter set, the normalisation mode and the evaluation protocol all differ. Their arm refreshes the batch-normalisation running statistics, whereas the models used here stay in evaluation mode throughout, although the affine parameters that scale and shift the normalised activations are updated with the rest of the weights. These differences may contribute to the opposite outcomes, but the comparison that would settle it — the same arm under both normalisation regimes on the same split — has not been run. Read together, the two studies indicate that the outcome depends on the update rule and the evaluation configuration, so they do not contradict each other.

Table 15 places the studies cited here on three axes that decide what a result can support: what the experiments vary, the unit over which the score is averaged, and what is then analysed. The table is not a record of carelessness: most of those designs do not vary the encoder at all, so the question this study asks does not arise in them, and the two that do vary it report at the dataset level.

Read against that table, the contribution lies in the endpoint. This paper does not ask whether adaptation can degrade, whether per-class scores can be computed, or whether stream organisation can matter; the analysis asks which classes improve or deteriorate within each subject, and whether the sign of the within-subject class contrast survives a change of encoder. PTB-XL supports between-subject, class-stratified comparisons only; Siena and Sleep-EDF support within-subject class comparisons, and §4.3.5 tests the contrast on both.

### 2.4 Monitoring a continuously updated model

Schirmer et al. [12] apply sequential testing and confidence sequences to risk monitoring for continuously updated models, asking how to decide without live labels whether adaptation should be stopped. The method requires calibration data, assumes batches are internally i.i.d., and trades detection delay against false alarms.

This paper takes up the monitoring question but tests a different mechanism. For a monitor to help, three conditions must hold: there must be a gain worth protecting, the monitored quantity must change before performance falls, and outperforming continuous adaptation must be equivalent to outperforming no adaptation. S20 tests each of the three conditions.

## 3 Methods

### 3.1 Design and scope

In online test-time adaptation for neural-signal models, acquisition shapes the incoming stream. This study asks how adaptation-related changes are distributed across subjects and classes, whether paired stream-organisation contrasts alter the measured outcome, and whether the prespecified unlabeled monitoring and protection strategies meet their criteria. For subject-level inference the independent unit is the subject: within each seed, cross-validation places each subject in one held-out fold and yields one out-of-fold difference. The four seeds repartition the same subjects and retrain source models, so each seed is tested separately and cross-seed agreement is used only as a stability check; fold-level records remain the aggregation unit for the aggregate table, not the inferential sample size.

Without labels, no rule in these experiments can determine whether an update will be harmful, and none of the three strategies meets its criterion. The results do show what such a decision would need to account for: how damage is distributed across subjects and classes, how stream organisation affects it, and when the two label-free indicators can be interpreted.

### 3.2 Datasets and splits

PTB-XL, Siena and Sleep-EDF form the 11-cell core analysis; the memory-limited MedFormer × Sleep-EDF cell is omitted. BCI-IV-2a and CHB-MIT are extensions used for boundary and monitoring context and are not pooled into the core subject-level claims (Table 1).

Four of the five datasets record EEG: Siena [13] (13 subjects, two balanced classes, five folds), Sleep-EDF [14] (20 subjects, five sleep stages, five folds), BCI-IV-2a [15] (9 subjects, four balanced classes, five folds) and CHB-MIT [16] (24 subjects, two classes, eight folds). PTB-XL [17] records ECG (198 subjects, two classes in a subject ratio of 51:147, eight folds) and enters as a cross-modal comparison rather than as neural evidence.

**Table 1. Datasets and their role.** PTB-XL is ECG; Siena, BCI-IV-2a, Sleep-EDF and CHB-MIT are EEG. The core cross-cell analysis uses PTB-XL, Siena and Sleep-EDF; the two EEG datasets with near-chance source skill (BCI-IV-2a, CHB-MIT) are extensions. Samples per subject is the number of test samples the subject contributes in this study, given as median [IQR] across subjects; it is not the size of the original recording, and subjects differ in how many windows survive preprocessing. Window gives channels × samples per input. Per-subject counts are not stored for the two boundary datasets and their entries are marked —. Sampling rate is fixed by the preprocessed feature files rather than recorded per run, so window length is given in samples. Folds is the subject-wise partition used for the dataset and Seeds the number of seeds run for it: four for the three core datasets, one for each boundary dataset.

| Dataset | Signal | Subjects | Classes | Samples per subject | Window (channels × samples) | Chance BA | Folds | Seeds | Role |
|---|---|---|---|---|---|---|---|---|---|
| PTB-XL | ECG | 198 | 2 (one per subject) | 229 [139–519] | 15 × 300 | 50.0% | 8 | 4 | order comparison; class stratification |
| Siena | EEG | 13 | 2 | 214 [148–256] | 29 × 800 | 50.0% | 5 | 4 | order comparison; within-subject classes |
| BCI-IV-2a | EEG | 9 | 4 | — | 22 × 600 | 25.0% | 5 | 1 | three-arm order; boundary observation |
| Sleep-EDF | EEG | 20 | 5 | 488 [447–500] | 1 × 3000 | 20.0% | 5 | 4 | order comparison; within-subject classes; monitoring |
| CHB-MIT | EEG | 24 | 2 | — | 23 × 1024 | 50.0% | 8 | 1 | low-skill boundary case |

The available subject-level measures depend on how labels are assigned. Each PTB-XL subject has one true class, so balanced accuracy cannot be calculated for that subject; recall for its class is reported instead. Siena and Sleep-EDF include several classes per subject, which allows subject-level balanced accuracy to be calculated and compare class-specific recall within a subject.

Each seed uses a different cross-validation split and a source model retrained on that split, so seeds are not technical replicates; all stream orders within a seed share the split and source model, and the same fold number refers to different subject partitions across seeds (S12).

## 3.3 Encoders and arms

The evaluation covers four encoders from three architecture families: a Mamba sequence model [18], MedFormer [19], EEGNet [20] and a plain convolutional encoder. These are treated as four implementations; because MedFormer and the CNN differ in several respects, the comparison does not isolate a single architectural factor.

All four share one self-supervised objective. Let `f` be the encoder's feature map, `x` an input batch and `M(x)` the same batch with a temporally contiguous span set to zero. The target is the model's own feature map on the unmasked batch, computed before that batch's update under `no_grad` and detached, so the target follows the model as it adapts instead of staying at the source model's features; one update minimises

$$\mathbf{L(x) = || f(M(x)) - sg[\ f(x)\ ] ||^2} \tag{1}$$

where `sg[·]` denotes the stop-gradient operator. The mask `M` zeroes 30% of the time steps in spans of 16 samples, with span starts drawn uniformly and the same mask applied to every channel of a sample. This study reconstructs features, not inputs, because the source model is already trained: an input-space objective admits the identity solution, whereas a feature-space objective against a detached target does not; a constant feature map makes the loss exactly zero, so nothing in the objective opposes a loss of feature diversity. That state is stronger than the concentration of predictions this paper measures, and §5 keeps the two apart. Each incoming batch receives one Adam step. Table 2 lists the arms, all of which use the same evaluation pipeline. For the sequence model the initial state `h0` is held at its zero value throughout: no arm in this grid updates it, so the comparison is between weight-space rules only.

**Table 2. Adaptation arms.**

| Arm | Updated parameters | Objective | Budget | Learning rate |
|---|---|---|---|---|
| `source_only` | none | — | — | — |
| `full_ttt` | all backbone parameters | masked-feature reconstruction | full stream | 1e-4 |
| `tent_ttt` | normalisation affine only | entropy minimisation | full stream | 1e-3 |
| `shot_ttt` | normalisation affine only | information maximisation + diversity | full stream | 1e-3 |
| `adapter_ttt` | bottleneck adapters | masked-feature reconstruction | full stream | 1e-3 |

The inferential analysis uses `full_ttt`, but a single adaptation rule cannot tell whether the deterioration belongs to test-time adaptation in general or to one implementation, so the same runs executed three further rules on identical splits, seeds and cached source models; §4.4 reports them. They were chosen for their distinct mechanisms, not as a sample of baselines, and they vary three things: the objective, the regularisation term and the parameter scope. The comparison is between algorithms, not objectives alone, because the normalisation-affine rules and the adapter arm share a learning rate of 1e-3 while the full-parameter arm uses 1e-4.

The three rules differ in objective and in which parameters may move. `tent_ttt` updates only the scale and shift parameters $\gamma$, $\beta$ of the encoder's normalisation layers, by one Adam step on the mean conditional entropy. `shot_ttt` keeps that parameter set and subtracts the entropy of the batch mean, which penalises the degenerate solution in which every sample receives the same label. `adapter_ttt` keeps the reconstruction objective and inserts a zero-initialised bottleneck adapter at the encoder input, so that it starts as the identity and every change is attributable to adaptation; it is the parameter-scope control and applies the same objective as `full_ttt` to about 4,096 parameters instead of the whole encoder. S5 gives both objectives in full and records the negative control in which the diversity weight is set to zero.

`shot_ttt` and `tent_ttt` are SHOT-style [21] and TENT-style [22] rather than bit-exact reproductions: `shot_ttt` omits the pseudo-label refinement and clustering steps of the original and fixes the diversity weight at 1, and `tent_ttt` applies TENT's objective to the normalisation affine parameters of each encoder rather than to one published architecture. Results for these two arms are therefore statements about the objectives, not replications of the published implementations.

`full_ttt` follows the self-supervised test-time training of Wang et al. [26]: the objective is masked-feature reconstruction and all backbone parameters are updated, while the classification head receives no gradient because the objective is defined on the pooled feature before the head. Each update is made per incoming batch, so the number of steps follows stream length instead of a fixed count shared across datasets.

Two experiment blocks were run. The order-comparison block includes four encoders across PTB-XL, Siena and Sleep-EDF, with MedFormer omitted from Sleep-EDF because of limited GPU memory, leaving eleven cells; the seed block repeats these runs with four seeds. The blocks were run independently, so matching configurations may use different weights and their results are not pooled.

### 3.4 Stream construction and evaluation

Within each record, the same test samples are used and only their order is changed across four orders. `orig` is the order supplied with the dataset; `p1` randomly shuffles all samples. `b1` shuffles samples only within blocks that share both subject and class. `b2` shuffles within class blocks: samples from different classes stay in separate blocks, but subjects can mix.

Three permutations are generated for each shuffled order (`p1`, `b1` and `b2`) and their scores are averaged within the record. Each changes which samples are neighbours, which share a batch, and where batch boundaries fall, so the effect of stream organisation is assessed as a whole.

For each incoming batch, the adaptation update was performed first, followed by prediction on that same batch using the updated model. This protocol evaluates post-update predictions on the batch used for the corresponding unlabelled adaptation step. It is therefore batch-level transductive test-time adaptation, not strictly prequential prediction on an unseen incoming batch. Because the model carries its state forward, each prediction also depends on the updates made on all earlier batches.

One implementation detail determines how the subject-level results must be read. For each (dataset, encoder, seed, fold) the runner takes a single copy of the source model and passes the held-out samples of **all** test subjects to it as one continuous stream, in the order the arrangement defines. Subject labels are used only to group predictions after the run; the model, its optimiser state and its normalisation statistics are never reset between subjects. A subject's adaptation effect therefore describes the state of the stream at the point where that subject's samples arrive, not a model that adapted on that subject alone. A system installed once and left running across successive users would behave in the same way, so the choice is defensible. Even so, it does mean that subject-level observations are not independent draws from a per-subject process, and that the sign tests in §4.2 are conditional on the arrangement that produced them. §4.2 measures that dependence; S7 re-runs the two subject-level cells with the stream cut at subject boundaries.

The same run can be evaluated in two ways, and the two need not agree. For each batch, the **streaming score** pools the prediction made after that batch's update, so it mixes states from a barely adapted model to one that has absorbed the whole stream. The second is a terminal-state evaluation on the adaptation stream, abbreviated here to **terminal-state score**: the model is evaluated once on the full test set after the stream has ended, so it describes the state a run leaves behind, not an average over its states. Because that stream is the one used for adaptation, the score is not an estimate of performance on data the model has not seen, and not a forecast of deployment behaviour. Every performance value in §4.1–4.8 and S7–S10 is a streaming score; S11 adds the terminal-state score for four cells, recomputed from the saved final state of each run on the same stream used for adaptation.

### 3.5 Metrics and aggregation

Balanced accuracy (BA) is the unweighted mean of per-class recall, with a chance level of 1/K. For a subject with a single true class, individual BA is undefined and the recall of that class is reported instead.

The **adaptation effect** is adapted performance minus source performance on the same test samples in the same arrangement. The paired **stream-organisation contrast** δ (called the order effect in the tables) is the adapted result under shuffled arrangements minus that under the original; it says nothing about whether either arrangement is better than not adapting. The **transition quantities** are the share of samples whose predicted class changes, the share whose correctness flips, and the net change in the fraction correct; with more than two classes a label change need not flip correctness, and the gap is itself informative (S18).

For aggregate BA, paired differences are computed within each (dataset, seed, fold) record and records receive equal weight; they are not independent subjects and are not used as the inferential sample size (S6).

Subject-level results use subject-equal weighting as primary, with sample-weighted values added where the two differ by more than 1 point; the distinction matters because PTB-XL subjects contribute 39–965 samples each. Direction is tested separately within each seed using an exact, two-sided sign test of worsening versus improvement, excluding subjects with exactly zero change and reporting their count separately, with Holm correction across the 11 core cells within each seed; seeds are not pooled as independent observations. The four Siena within-subject class contrasts average each subject's class-recall difference across seeds and then use a sign test over 13 subjects, corrected across the four encoders.

Pairwise differences between encoders are tested by an exact sign-flip permutation, whose null is that the within-subject difference $D_i$ is symmetric about zero — what exchangeability of the two encoders within a subject implies. Every sign assignment is enumerated, so no assumption is made about the shape of the distribution. Exactness needs a second assumption that the permutation does not test: that the $D_i$ of different subjects are independent. The fold design does not supply it, because one seed assigns the same folds to all four encoders, which removes any fold-level term the two encoders share but leaves the difference of their fold-level terms, common to every subject of a run, in D. §4.3.3 therefore audits the dependence on D itself, and reports beside the permutation two tests that need no subject independence: an exact sign flip at the level of the run, and a cluster-robust Wald test with the run as cluster. Where the encoders are not exchangeable within a subject, for instance because a subject's fold differs between them, the symmetry null fails outright; §4.3.3 records the check that no subject falls in different folds across encoders. For four-split subject means in §4.2, unique subject IDs are resampled 50,000 times (percentile CI; S6); the interval is descriptive and conditional on the four evaluated splits, not a further test.

The sign test treats the subjects of a fold as independent, and they are not, because the stream is shared (§3.4). §4.2 measures that dependence at both levels where it can act: the intraclass correlation of a single (seed, subject) change within its run, and the correlation between subject means that shared runs would produce. It gives two readings, a direction count with each fold run counted once, which is not reported as a test, and the sign test at the effective sample size implied by the most adverse reading, which is a sensitivity check; the subject-level test remains the one reported here.

The relation between aggregate change and the class contrast is tested with two nested linear models on subject-centred data, so only within-subject variation enters. For class pair (a, b) and encoder e, $C_{i,e}$ is subject i's within-subject contrast and $A_{i,e}$ that subject's change in balanced accuracy. M1 regresses C on A with subject fixed effects, M2 adds encoder dummies, and $\Delta R^2$ is the increase in explained variance, tested by a partial F with $k-1$ and $n(k-1)-k$ degrees of freedom (n subjects, k encoders). Subject centring absorbs n means, so the residual degrees of freedom are not $n \cdot k$ minus the parameter count. Under the repeated-measures design that F is exact under sphericity of the residual covariance across encoders, not under independence, and the residual correlations sit at the value subject centring imposes mechanically (§4.3.6). Two checks accompany it, with different nulls. A Monte Carlo permutation of encoder labels within each subject, 10,000 draws, preserves each subject's joint distribution of the two quantities; full enumeration is infeasible, each subject contributing 6 or 24 label orders depending on the cell. The attainable two-sided p-value has a floor of 1/10001. Its null — joint exchangeability across encoders — is stronger than the question asked, whether encoder identity adds explanatory power once aggregate change is controlled, so rejecting it does not establish the incremental claim and the partial F remains the reported test. A cluster-robust Wald test on the encoder coefficients needs neither sphericity nor exchangeability, only independence across clusters; its cluster is the subject, because the model is fitted to subject means, where a single run cannot be assigned to an observation. That covers a subject's own four splits but not runs shared between subjects, which the refit of §4.3.6 measures and which is not negligible; neither covers the source models, trained on the same cohort, which had the held-out subject in their training or validation split for 83 to 85% of that refit's training rows. The refit is a leave-one-subject-out prediction, not the jackknife of §4.2: both models are re-estimated on the remaining subjects and applied to the held-out subject's centred deviations, in-sample and out-of-sample residuals are divided by one total sum of squares, and encoder labels are permuted within subject for the null. One model pair is fitted per class pair; the eleven p-values form one Holm family, and the medians in §4.3.6 are descriptive across those eleven comparisons, not the explained variance of a pooled model.

Four Holm families are defined by the quantity tested, not by the cells they cover. The first is the subject-level direction test of §4.2, within each seed across the 11 core cells. The second is the Siena class contrast of §4.3, across the four encoders. The third is the subject-level change by adaptation rule, 11 cells × 4 rules = 44 tests, of which Table 18 displays the eight belonging to the two narrative cells. The fourth compares aggregate change against the class contrast of §4.3.6, across the eleven class-pair comparisons. A test belongs to exactly one family; Table 19 sets out the full plan, giving the unit, the method and the correction family for every analysis reported here (S22). Table 7 compares two stream constructions and carries no test, and the transition counts of S18 and the monitoring quantities of S20 are reported without correction because they are not used to declare an effect.

A subject with a change below −20 pp is classified as substantially affected for this operational analysis. This threshold is not a clinically validated risk boundary, and its sensitivity is given in §4.2 at −10 and −30 pp; the resulting proportions are not compared across datasets.

Reproducibility, archive scope and the deviations from the protocol fixed before each run are reported in Supplementary S12.

# 4 Results

Three subsections carry the claims of §1: §4.1–4.2 fix the aggregate and the subject-level distribution against which the class contrast is formed, §4.3 establishes that the direction of that contrast is encoder-dependent, and §4.3.6 quantifies how much of it an aggregate change already accounts for; the evidence for the main claim is complete by §4.3. One condition applies to every subject-level test below: all held-out subjects of a fold are adapted as one continuous stream (§3.4), so the subject is the unit but not an independent draw; §4.2 quantifies that dependence and what survives it. The subsections after §4.3 are boundary conditions rather than further claims: §4.4 asks whether the same deterioration appears under other adaptation rules, and §4.5 whether the sign of an order effect depends on whether adaptation state is shared. S14–S17 report a dated-protocol intervention on the timing of input-stage parameter displacement that is not part of the two claims; none of these protocols was registered in a public registry.

## 4.1 Aggregate performance conceals individual deterioration

Table 3 gives the source model, the adapted model under the original order and after permutation, all as four-seed means.

**Table 3. Balanced accuracy under three conditions (four-seed means, pp).**

| Dataset | Encoder | Source | Original | Shuffled | Original − source | Shuffled − source | δ (SD over seeds) |
|---|---|---|---|---|---|---|---|
| PTB-XL | CNN | 80.7 | 49.1 | 52.2 | −31.6 | −28.5 | +3.12 (0.19) |
| PTB-XL | EEGNet | 73.7 | 67.6 | 64.1 | −6.2 | −9.7 | −3.47 (1.27) |
| PTB-XL | MedFormer | 78.9 | 70.6 | 62.6 | −8.3 | −16.3 | −7.92 (8.23) |
| PTB-XL | Mamba | 73.6 | 54.0 | 62.5 | −19.6 | −11.1 | +8.48 (1.51) |
| Siena | CNN | 66.8 | 50.0 | 50.3 | −16.7 | −16.4 | +0.29 (3.08) |
| Siena | EEGNet | 60.4 | 60.3 | 61.0 | −0.1 | +0.6 | +0.71 (0.47) |
| Siena | Mamba | 57.2 | 57.9 | 55.5 | +0.6 | −1.7 | −2.38 (0.13) |
| Siena | MedFormer | 78.0 | 48.4 | 50.4 | −29.6 | −27.6 | +2.07 (1.65) |
| Sleep-EDF | CNN | 43.0 | 19.4 | 20.0 | −23.6 | −23.0 | +0.61 (0.84) |
| Sleep-EDF | EEGNet | 54.4 | 29.2 | 27.7 | −25.2 | −26.7 | −1.54 (3.26) |
| Sleep-EDF | Mamba | 49.3 | 31.6 | 31.1 | −17.6 | −18.1 | −0.49 (0.38) |

Under the original order, ten of the eleven core cells are below source; Siena Mamba is the only exception (+0.6 pp). After permutation the count is unchanged but the exception moves to Siena EEGNet (+0.6 pp), and Siena Mamba falls to −1.7 pp. In separate subject-level sign tests for each seed, 7, 9, 8 and 8 cells remain significant after Holm correction across the 11-cell family, and seven pass in all four seeds; counts of subjects worsened, unchanged and improved are in `v2_subject_level.json`. These tests address whether more subjects worsened than improved, not the magnitude of the mean change, and they are conditional on the shared stream. S7 cuts the stream at subject boundaries: the full-parameter result is unchanged, with 20 of 20 Sleep-EDF subjects falling under all four constructions, but for the adapter arm neither the level nor the identity of the subjects who gain and lose survives the cut. Those counts therefore describe the evaluated run, not a per-subject tendency (§6).

Negative mean changes have been reported before: a benchmark of 180 comparisons found 128 degradations [4]. Here the size of the change cannot be compared across rows, because balanced accuracy weights classes equally and the same point change has different implications at different starting levels. The claims below are therefore made within cells rather than across datasets.

**What the adapted model emits.** The mean does not say which classes survive. The analysis restricts attention to subject-cells whose ground truth contains at least two classes each covering a tenth or more of that subject's samples: 112 of the 3616 subject-cells, with PTB-XL contributing none because a patient's beats in one fold carry a single label. Over the pooled post-update predictions of the stream, the adapted model gave one of those classes no prediction at all, in all four seeds, in 44 of 112 cases (39.3%), against 9 of 112 (8.0%) for the source model. Both counts use the same 112 cells, the same samples and the same four seeds; only the prediction source differs, so the comparison is paired within subject, not between cohorts. The count is not specific to the full-parameter rule: under the same definition it is 47 of 112 under entropy minimisation of the normalisation affine parameters and 26 under information maximisation with a diversity term, against 7 for the adapter arm, which leaves the backbone frozen. The arms differ in learning rate as well as in objective (§4.4), so these counts describe the rules as implemented rather than isolating a cause. The eligibility rule and the deletion criterion were fixed after an earlier count was found to be contaminated — a patient's beats in one fold carry a single true label, so a model that emits one class is correct there — and they are a corrected definition rather than a prespecified one. The 10% threshold is not load-bearing either: no class in these cohorts falls between 5% and 15%, since every Siena subject is exactly 48/48 and the smallest Sleep-EDF class covers 20%, so 5%, 10% and 15% select the identical 112 cells and the adapted count moves only between 45 and 43, the source count between 9 and 6. The distribution is uneven: 20 of 20 on Sleep-EDF with CNN, 13 of 20 with EEGNet, 1 of 20 with Mamba; 6 of 13 on Siena with MedFormer, 4 of 13 with CNN, none with EEGNet or Mamba (`v2_class_collapse.json`). On Sleep-EDF it is the short class that disappears. REM recall falls from 30.9% to 0.00% under CNN, from 79.7% to 10.5% under EEGNet and from 56.2% to 10.9% under Mamba; N1 falls to 0.00%, 1.1% and 4.7%; and the share of predictions assigned to N3 rises from 13.4% to 45.0%, from 23.2% to 62.9% and from 20.7% to 59.5%. The adapted model has not become uniformly worse; it has become a different classifier, one that reports deep sleep and little else.

The 44-of-112 count is measured under the shared-stream construction. S7 recounts the same criterion on the archived isolated-stream predictions for the two cells with cut-stream runs, 33 eligible subjects: 24 of 33 under the shared stream and 25 of 33 under the cut stream, with 9 of 33 from the source model in both. Those two cells account for 24 of the 44 deletions.

## 4.2 Subject-level distribution of adaptation harm

Figure 2 shows subject-level changes for PTB-XL × CNN, Siena × Mamba and Sleep-EDF × CNN across four splits; Table 4 summarizes their distributions. PTB-XL combines a large unchanged group with a long negative tail, whereas most Sleep-EDF subjects decline by a more similar amount. Their mean changes are both negative but are not identical; Siena × Mamba is the near-null comparison. These are contrasting examples, not an exhaustive taxonomy.

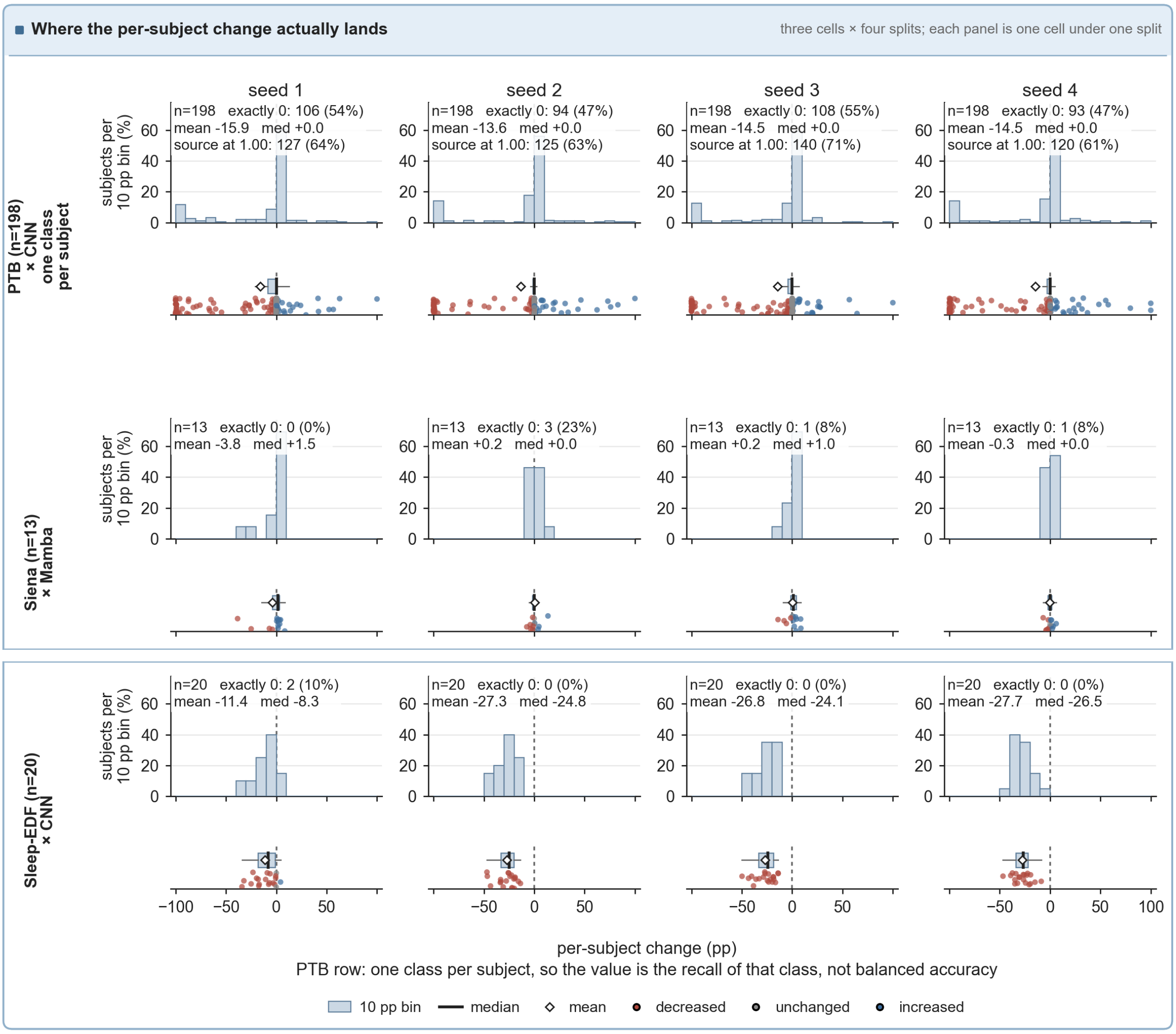

**Figure 2. Negative mean changes conceal different subject distributions.** Each panel shows one cell and one random split (seed): PTB × CNN (n = 198), Siena × Mamba (n = 13), Sleep-EDF × CNN (n = 20). Upper part: percentage of subjects per 10 pp bin of the per-subject change (adapted − source, same test samples, pp), with bin edges identical in all twelve panels; lower part: one dot per subject, coloured by sign, with a box plot (Q1–Q3, median, whiskers at 1.5 × IQR) and the mean as a white diamond. Splits are shown separately rather than pooled, because a new split re-partitions the same subjects and pooling would inflate n fourfold. Numbers above each panel give n, the count at exactly zero, the mean and the median. The three rows are not the same quantity: in Siena and Sleep-EDF each subject contributes both classes and the value is a change in balanced accuracy, while in PTB every subject contributes a single true class, so balanced accuracy is undefined and the value is the change in the recall of that class; the point mass at zero is a ceiling, not indifference.

**Table 4. Shape of the per-subject change (four splits; PTB-XL values are changes in the recall of the subject's own class).**

| | PTB-XL × CNN | Sleep-EDF × CNN | Siena × Mamba |
|---|---|---|---|
| Mean (pp) | −15.9 / −13.6 / −14.5 / −14.5 | −11.4 / −27.3 / −26.8 / −27.7 | −3.8 / +0.2 / +0.2 / −0.3 |
| Median (pp) | 0.00 in all four | −8.3 / −24.8 / −24.1 / −26.5 | +1.5 / 0.00 / +1.0 / 0.00 |
| Share unchanged | 47.0%–54.5% | 0–10% | 0–23% |
| Share worse | 33.3%–37.4% | 85%–100% | 31%–46% |
| Share ≤ −50 pp | 16.2%–18.7% | 0% in all four | 0% in all four |
| Interquartile range (pp) | 1.50–8.34 | 12.23–17.05 | 2.14–7.03 |
| SD across subjects (pp) | 36.91–40.51 | 9.01–11.29 | 3.39–13.23 |

After averaging each subject's change over the four seed-specific splits, the mean was −14.61 pp for PTB-XL × CNN (95% subject-cluster bootstrap CI, −19.76 to −9.58), −23.30 pp for Sleep-EDF × CNN (CI, −26.33 to −20.24), −0.90 pp for Siena × Mamba (CI, −3.51 to +1.51), and −0.19 pp for Siena × EEGNet (CI, −1.04 to +0.40). Each interval resamples subjects after averaging their four split-specific values and is conditional on the four fitted split/model instances; it is an uncertainty summary, not the decision rule for the subject-level tests. The means differ in magnitude, but the mean alone cannot recover the distribution shape: on Sleep-EDF the median tracks the negative mean, whereas on PTB-XL the median is 0 while a tail falls by more than 50 points.

The zero mass on PTB-XL has a concrete source: every subject contributes one class, and the 147 subjects in class 1 start at a recall between 93.4% and 96.1%, leaving almost no room to rise; of the 106 subjects with exactly zero change in one split, 100 were already at recall 1.00. The zero is a ceiling, not an absence of effect, which is why the dataset is excluded wherever a change in balanced accuracy is required at the subject level.

Siena × Mamba is a near-zero average with changes in both directions: 31%–46% of subjects worsen and the median reaches +1.5 pp in one split. With EEGNet the spread is much narrower and the mean stays within 1 pp of zero. These are different near-null patterns, and a non-significant test is not evidence that the effect is exactly zero.

Across the four Siena encoders, the median subject change separates the two groups: −23.7 pp (CNN) and −31.9 pp (MedFormer) against +0.2 pp (EEGNet) and +0.6 pp (Mamba), with 11 of 13 and 13 of 13 subjects negative for the first two and 3 of 13 and 6 of 13 for the latter two.

Removing any one subject leaves both cells where they were: the 20 jackknife fits for Sleep-EDF × CNN stay between −24.0 and −22.6 pp with every fit still significant after correction, and the 13 fits for Siena × CNN between −20.3 and −16.6 pp; no single removal flips either sign (S4).

Subjects within a fold are not independent observations, and the dependence was measured instead of being assumed away. Taking each fold run as a cluster and each (seed, subject) change as an observation, the one-way intraclass correlation of δ is at or below 0.02 in eight of the eleven cells. It is largest for Sleep-EDF × CNN, at 0.786, where adjacent subjects in stream order correlate at +0.78 and the 80 (seed, subject) changes carry about as much information as 24 (design effect 3.36); the next largest values are 0.194 and 0.064. Under the most adverse reading that decomposition admits, the adjusted p for that cell rises to 0.094 without changing the verdict, and the direction count does not depend on how the stream is partitioned (S24).

Figure 9 shows the same distributions as the share of subjects below each change threshold.

Averaged over four splits, the shares at or below −10 / −20 / −30 pp were 22.1% / 20.8% / 19.3% for PTB-XL, 85% / 60% / 30% for Sleep-EDF and 5.8% / 3.8% / 1.9% for Siena. This sensitivity analysis shows how threshold choice changes the operational count; it does not identify a clinically meaningful cut-off. These percentages should not be ranked across datasets, because PTB-XL uses each subject's class-specific recall change whereas Siena and Sleep-EDF use subject-level BA. Below −20 the PTB-XL curve is nearly flat, whereas the Sleep-EDF curve falls to 8.8% at −40 and zero at −50 (S4).

## 4.3 Class-selective deterioration differs by encoder

Figure 3 separates PTB-XL's between-subject class stratification from the within-subject class comparisons available in Siena and Sleep-EDF.

### *4.3.1 PTB-XL: class-stratified subject groups*

Each PTB-XL subject contributes samples from only one true class, so the comparison is between class-associated subject groups rather than class-specific changes within a subject, and the unit is the subject's own class recall (Table 16, S25). Under CNN, sample-level class-0 recall falls from 66.5% to 0.08% while class-1 recall changes little, so balanced accuracy falls by 31.6 points against a 14.6-point subject-equal mean; 147 of the 198 subjects belong to the less affected group. MedFormer reverses the ordering: its class-0 group mean is positive in all four splits while class-1 recall falls by 25–40 points, yet the per-subject median there is −0.2 pp and 15 of 51 class-0 subjects still fall by more than 20 pp, so the group-level result is described as no overall deterioration in that group rather than as a class-specific improvement. The label files do not preserve recoverable clinical semantics, so classes are identified by index only.

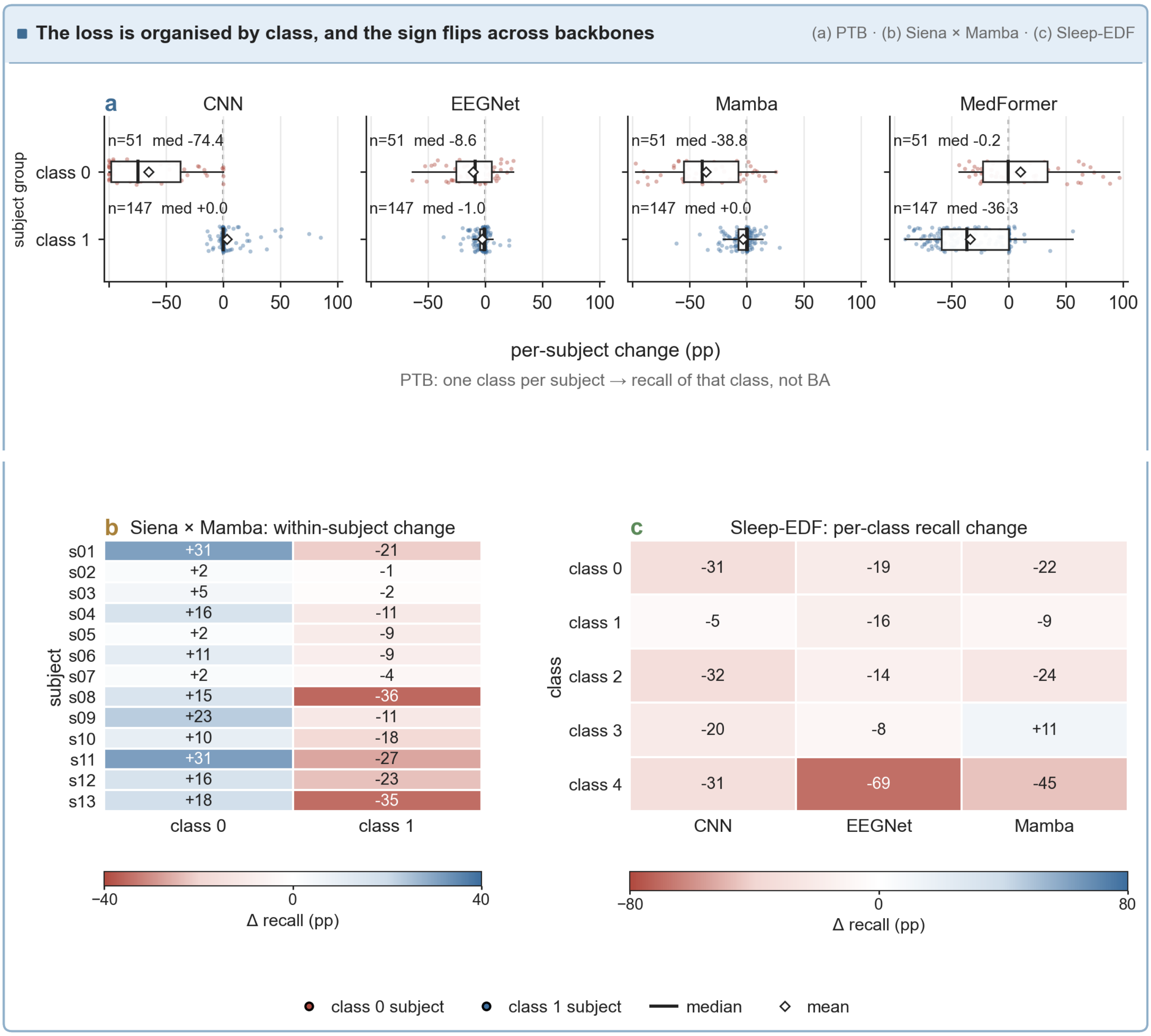

**Figure 3. Class-associated and within-subject changes across datasets.** (a) PTB: each subject contributes one true class, so class stratification forms two subject groups (n = 51 / 147); this is a between-group comparison. Values are changes in recall for each subject's true class; points show subjects across four backbones, averaged over four splits, with boxplots as in Figure 2. (b) Siena × Mamba: within-subject change in per-class recall, one row per subject (n = 13). (c) Sleep-EDF: mean within-subject change in per-class recall for five sleep stages and three backbones. Panels (b) and (c) use separate symmetric colour scales (±40 and ±80 pp); values are printed in each cell. Subject–class cells with fewer than 20 samples are missing, not zero.

### *4.3.2 Within-subject comparisons across classes*

Siena (13 subjects) and Sleep-EDF (20 subjects) permit class comparisons within each subject, which PTB-XL does not; the pairing controls between-subject differences in the class contrast, although the small cohorts limit generalization.

**Table 5. Siena: within-subject difference in per-class recall, class 0 minus class 1 (pp).** Each subject's class-recall changes are first averaged across four seeds. The counts give subjects whose mean difference is negative, exactly zero and positive. A subject with a mean difference of exactly zero carries no sign and is dropped from that test, so the number entering each test (n used) is 13 except for EEGNet, where one zero difference leaves 12; p-values are two-sided exact sign tests with Holm correction across the four encoders.

| Encoder | Mean difference | Median difference | Class 0 worse / equal / better | n used | p (Holm) |
|---|---|---|---|---|---|
| CNN | −54.58 | −40.00 | 12 / 0 / 1 | 13 | 0.0068 |
| EEGNet | −4.39 | −4.17 | 12 / 1 / 0 | 12 | 0.0015 |
| Mamba | +29.93 | +28.12 | 0 / 0 / 13 | 13 | 0.0010 |
| MedFormer | −29.68 | −1.37 | 7 / 0 / 6 | 13 | 1.00 |

After Holm correction, three of the four encoder-specific within-subject tests are significant, with effects in different directions. CNN lowers class 0 by 45.9 points and raises class 1 by 8.7; EEGNet moves both by less than 2.5 points; Mamba reverses the pattern, raising class 0 by 14.1 points and lowering class 1 by 15.9 in all thirteen subjects; and MedFormer lowers both classes, by 44.3 and 14.6 points, so its between-class difference is not significant. Within this small cohort and the evaluated protocol, the direction of the within-subject class contrast varies across the four implementations.

A cell counts as computable when the subject has at least 20 samples of that class; the smallest cell here has 104, so the threshold does not bind.

Sleep-EDF shows a different pattern: under CNN all five classes fall, by 5.3 to 32.3 points, and under EEGNet all five fall, by 8.4 to 69.2. Under Mamba four fall and one rises: class 3 gains 11.2 points in nineteen of twenty subjects while class 4 loses 45.3. Those are changes; Table 6 gives the levels they are computed from, so that the recall a stage retains can be read next to the amount it lost. The changes do not point to a single directional shift that explains the results, and §4.3.5 forms the contrast between stages and asks whether the encoder changes it.

**Table 6. Sleep-EDF: per-stage recall before and after adaptation (%, sample level).** Values pool every test sample of the cohort across four seeds and five folds (n = 37,780 per encoder); Δ is the difference of the two adjacent columns and differs by at most 1 pp from the subject-level means of Figure 3(c), which weight subjects equally. MedFormer is absent from this cohort (§3.3). Bold marks a stage that receives no prediction at all.

| Stage | CNN source → adapted | Δ | EEGNet source → adapted | Δ | Mamba source → adapted | Δ |
|---|---|---|---|---|---|---|
| W | 66.6 → 35.4 | −31.2 | 64.6 → 45.8 | −18.8 | 43.1 → 21.5 | −21.6 |
| N1 | 5.4 → **0.00** | −5.4 | 17.0 → 1.1 | −15.9 | 13.8 → 4.7 | −9.2 |
| N2 | 52.3 → 20.1 | −32.3 | 21.1 → 6.8 | −14.3 | 46.2 → 22.5 | −23.6 |
| N3 | 61.2 → 41.0 | −20.2 | 88.9 → 80.0 | −9.0 | 86.7 → 98.5 | +11.7 |
| REM | 30.9 → **0.00** | −30.9 | 79.7 → 10.5 | −69.2 | 56.2 → 10.9 | −45.3 |

A large fall and a disappearance are different statements. Mamba's REM loses more than CNN's, 45.3 against 30.9 points, and still receives 10.9% of that stage's samples, whereas CNN's receives none; N1 ends at 0.00% under CNN and at 1.1% under EEGNet. The size of a change therefore does not decide whether a stage is still emitted, and a table of changes alone would report the two cases alike. The two cases separate when subject-cells are counted: in 44 of the 112 subject-cells whose ground truth contains at least two classes each covering a tenth of that subject's samples, a major class received no prediction at all over the same stream-wide predictions, in all four seeds, against 9 for the source model. The distribution is uneven: 20 of 20 on Sleep-EDF with CNN, 13 of 20 with EEGNet and 1 of 20 with Mamba, and 6 of 13 and 4 of 13 on Siena with MedFormer and CNN. The count is lower than the endpoint behaviour of S17 would suggest, because a run that ends on one label may still have emitted the class earlier in the stream. §4.1 gives the same count for the other rules and states the eligibility rule, which is a corrected rather than a prespecified definition.

In a two-class problem, predicted class counts are linked, so a fall in one class's recall and a rise in the other's are coupled; the changes should not be counted as two independent harms; Table 5 gives the direction and size of the shift.

The class contrast carries its own dependence, and it needed its own audit, because it is not the quantity §4.2 tested: balanced accuracy is the mean of the two recalls while the contrast is their difference, and two such combinations need not share a dependence structure. The observation-level correlation with fold as cluster runs from 0.198 to 0.748 across the four encoders, and the bound it implies sets the effective sample size at 6.3 to 11.1, which moves the four Holm-adjusted contrasts so that two survive rather than three. A cluster-robust Wald test on the same contrast gives three, so the CNN contrast is the one whose verdict depends on which of the two readings is used. This paper keeps the reversal, which rests on the ordering of the four means and on Mamba's 13 of 13 subjects, and reports the CNN contrast as a direction with an effect size, not as one that survives every reading (S26).

### *4.3.3 Do the contrasts differ between encoders?*

Table 5 shows contrasts that differ in sign and size, but four tests against zero do not establish that the encoders differ from one another: one contrast can be significant while another is not, with no test of the difference between them. That difference was therefore tested directly. With $C_{i,e}$ the contrast of subject i under encoder e, the statistic is $D_{i,a,b} = C_{i,a} - C_{i,b}$, formed within subject after averaging the four seeds, so the subject remains the unit and the splits add no independent observations. One seed assigns the same folds to all four encoders, and it was verified that no subject falls in different folds across encoders. That removes any fold-level term the two encoders share, but it does not make D free of the run. Writing $C_{i,e} = \mu_e + u_{f(i),e} + \varepsilon_{i,e}$, the difference carries $[u_{f(i),a} - u_{f(i),b}]$, which is common to every subject of a run, so the design replaces a shared shift with a shared interaction. The term is not small: taking each (seed, subject) value of D as an observation and each (seed, fold) as a cluster, the one-way intraclass correlation is 0.451 to 0.805 across the six pairs, so most of the variance of a single value lies between runs, not within them.

The four contrasts differ (Friedman $\chi^2(3) = 21.18$, p = 0.0001, n = 13). Three of the six pairwise comparisons survive Holm correction under an exact sign-flip permutation: Mamba's contrast exceeds CNN's by 84.5 pp ($p_{Holm} = 0.0049$), EEGNet's by 34.3 pp (0.0015) and MedFormer's by 59.6 pp (0.0215). The other three do not: EEGNet over CNN by 50.2 pp (0.071), MedFormer over CNN by 24.9 pp (0.29) and EEGNet over MedFormer by 25.3 pp (0.29). The reversal is therefore established for Mamba against the other three encoders by this test, and it is carried by Mamba: its contrast is positive in all thirteen subjects, CNN's negative in twelve.

This study tested whether that dependence changes the verdict instead of arguing about it. Four readings are available, and they agree on the ordering while disagreeing on how many comparisons survive. The label permutation of §3.5, which assumes independent subjects, leaves three of six after Holm. An exact sign flip at the level of the run, which flips every observation of a run together and enumerates all 2^20 assignments over the 20 runs, leaves the same three ($p_{Holm}$ 0.0025, < 0.0001 and 0.0469). A cluster-robust Wald test with the run as cluster, which estimates the variance from the 20 clusters instead of assuming a form, inflates the standard error by 1.38 to 1.57 times and again leaves the same three (0.0013, < 0.0001 and 0.0447). Those three readings each handle one axis and assume the other away: the subject-level test removes the within-subject correlation by construction but treats subjects of the same run as independent, while the two run-level readings treat a subject's four seeds as independent observations in four different runs. The 20 runs are four seeds over the same five folds and the same thirteen subjects, so the axes are crossed rather than nested. A two-way cluster covariance, $V = V_{subject} + V_{run} - V_{subject \cap run}$, estimates both and is referred to t(12), the smaller cluster count less one; the difference need not be positive definite in general, but here it is positive definite in all six pairs, and the less favourable of the two single-axis variances is reported alongside. This paper does not claim that reference distribution is calibrated to the fold structure: the run axis has five fold levels, not twenty independent ones, and the variance under a fold-level cluster has not been recomputed. Under the two-way covariance two of the six survive, Mamba vs CNN ($p_{Holm}$ = 0.0088) and Mamba vs EEGNet (0.0009), and the single-axis reading leaves the same two (0.0032, 0.0003). Mamba vs MedFormer, which the other three readings retain at 0.0215 to 0.0469, does not survive (0.0801; single-axis 0.0550). Crediting the run axis with five fold levels rather than twenty changes only the reference distribution: one pair survives, Mamba vs EEGNet ($p_{Holm}$ = 0.034), while Mamba vs CNN moves to 0.081 and Mamba vs MedFormer to 0.22. The reversal is therefore established against EEGNet under every reading and every reference examined, against CNN under the exact readings and the nominal references but not under a fold-level reference, and against MedFormer under three of the four readings. The direction and the encoder carrying it are unchanged; the number of encoders it is separated from is not. Where the dependence reaches the subject means the verdict holds — subjects sharing a run are more alike in D for one pair, EEGNet vs Mamba ($\Delta r$ = +0.61, p = 0.013), which also has the smallest Holm p under all four readings (< 0.0001); for the other five the shared-run contrast is not distinguishable from zero ($\Delta r$ +0.09 to +0.37, p = 0.10 to 0.68). A detected dependence in one pair and a failure to detect in five is not independence, which is why the run-level and two-way tests are reported rather than the assumption.

Two qualifications apply to that reading: the CNN–EEGNet comparison changes verdict with the test, a Wilcoxon signed-rank test giving $p_{Holm}$ = 0.024 against 0.071 for the exact permutation; the permutation is reported because it is exact at n = 13, and the difference is not resolved at this cohort size. MedFormer's mean (−29.7 pp) and median (−1.4 pp) also disagree, with 7 subjects below zero and 6 above, so it lies between the two groups without a stable direction.

#### *4.3.4 What the source predictions do not explain*

One possible explanation is the source model's predicted class distribution. If it over-predicts class 0 before adaptation, an update that moves the marginal toward the true distribution could produce the observed recall changes. Siena has a 50/50 class distribution, so this account can be checked across four encoders and four seeds, and it fails: CNN and EEGNet do move toward the true marginal, but Mamba starts only 7.2 points too high and its class-0 share still increases by 15.7 points, the only move away from it. Four encoders are too few for a correlation test, so this paper does not run one. The source prediction distribution does not explain the direction of the class-conditional changes; the factors that might do so were not measured (§6).

#### *4.3.5 The same contrast on a second cohort*

§4.3.2 describes Sleep-EDF as five class-wise changes that do not point in one direction, but those are levels. With five stages and an uneven prior, the contrast $\Delta R_a - \Delta R_b$ was never formed on that cohort, and a description of levels cannot say whether the encoder changes the relation between two stages. To form it, this study computed $C_{i,e} = \Delta R_a - \Delta R_b$ per subject for each of the ten unordered stage pairs, averaged the four seeds within subject, and ran one Friedman test across the three encoders, Holm-correcting across the ten pairs. Only pairs that survived that correction were followed up, with three pairwise exact sign-flip tests each, Holm-corrected within the pair. The procedure was fixed before this cohort was analysed, and the follow-ups are exploratory: they are not independent confirmations, and they do not establish a ranking of encoders; they lend descriptive support only to the presence of an encoder-dependent contrast in this cohort. Because the contrast and its expected pattern were already known from Siena, this is a prespecified extension to a second cohort rather than a prospective registration. MedFormer is absent because its four seeds are missing on this cohort, so this test has three related samples rather than four.

Five of the ten pairs show an encoder effect after Holm correction: W vs REM ($\chi^2(2)$ = 20.1), N1 vs N2 ($\chi^2(2)$ = 13.3), N1 vs N3 ($\chi^2(2)$ = 13.1), N2 vs REM ($\chi^2(2)$ = 20.1), N3 vs REM ($\chi^2(2)$ = 15.6). The other five do not, including W vs N1 ($p_{Holm}$ = 0.285), W vs N2 ($p_{Holm}$ = 0.285), W vs N3 ($p_{Holm}$ = 0.075), N1 vs REM ($p_{Holm}$ = 0.075), N2 vs N3 ($p_{Holm}$ = 0.088). Of the 15 follow-up comparisons, 13 survive Holm correction within their pair. The largest is CNN minus EEGNet in N2 vs REM, a difference of 56.3 pp (p < 0.001).

That is the prespecified test, and it is run on twenty subject means: the four seeds are averaged within subject before the Friedman test, so each unit is one subject and the seeds add no independent observations. The dependence audit of §4.3.3 needs both of its axes named here. The sensitivity reading is a subject-fixed-effects regression on the 240 observation-level values — twenty subjects × four seeds × three encoders — with the encoder effect tested jointly and the covariance clustered by run, that is, by (seed, fold), twenty clusters. That handles one axis only: a subject's four observations fall in four different clusters and are treated as independent, and a subject's mean has no single run to belong to. The two-way covariance of §4.3.3 estimates both, V $= V_{subject} + V_{run} - V_{subject∩run}$, with 19 degrees of freedom for the smaller of the two cluster counts; it is not positive definite in any of the ten pairs, so it yields no valid joint-cluster inference here and this paper does not report one. The gate does not move at the run-clustered reference: two of the ten pairs survive Holm, W vs REM ($p_{Holm} = 0.009$) and N2 vs REM (< 0.001), with N2 vs N3 the first to fall at 0.057. That reference is itself tied to the twenty clusters; crediting the run axis with five fold levels leaves no pair (W vs REM 0.244, N2 vs REM 0.078). The largest of the three p-values — subject-clustered, run-clustered and two-way — is a numerical sensitivity summary, not a calibrated test: 0.041 and 0.001 for the same two, 0.121 for N2 vs N3. The observation-level intraclass correlation is lower than on Siena, 0.104 to 0.237, and the run-clustered standard error is 0.83 to 2.58 times the uncorrected one. The follow-up family those two pairs admit is six comparisons rather than fifteen; all six survive Holm within their pair, and four of the six survive the run-clustered reading, the two that do not being CNN vs Mamba in both pairs. The encoder effect is therefore established for five stage pairs under the prespecified test and for two under the run-clustered reading, the only one of the three that supports a valid one-way inference here; the two-way estimate adds no second confirmation.

The aggregate and the class contrast say different things here, so they have to be kept apart. At the level of the aggregate this cohort is close to uniform: all twenty subjects worsened under each of the three encoders, and the three subject-mean changes span 7.7 pp. At the level of the class contrast the same cells are far apart: contrasts differ by up to 56.3 pp and change sign between encoders. A single aggregate therefore reports the same conclusion for three models whose class-level redistribution is different. That is the point of the measurement, not an incidental detail: for the pairs that showed an encoder effect, a model with aggregate change alone accounted for 0.00 to 0.17 of the contrast variance (§4.3.6).

The encoder-dependence of the direction reproduces across the two cohorts. Within each cohort the contrast changes sign across encoders: on Siena it is negative under CNN and positive under Mamba, and in both Sleep-EDF pairs that survive the run-clustered gate CNN sits at zero (−0.3 and −1.4 pp) while the other two are positive. CNN is the lowest of the encoders in both cohorts, but the upper end is not the same — Mamba on Siena, EEGNet on Sleep-EDF — so the ordering does not carry over intact. CNN's position is not a property of the architecture: the two contrasts are different quantities, the comparison rests on two cohorts and four implementations that differ at once, and no manipulation isolates an architectural cause (§4.3.6). What does not reproduce is the claim that the departing encoder varies with the stage pair: N1 vs N3 puts CNN at the top and Mamba at the bottom, but that pair does not survive the run-clustered gate, so the variation is an observation from the uncorrected reading and not part of the cross-cohort result. This is a conceptual reproduction on a second cohort, not an independent replication of a measured effect.

Two limits bound what this result can support: the cohort has 20 subjects and ten pairs, so the correction is applied to ten tests and power is limited: not surviving is not evidence of no difference, and N2 vs N3, at 0.057 under the run-clustered reading, would be the first to revisit with a larger cohort. The contrast is also between two stages of an ordinal, unevenly distributed label set, whereas Siena's is between two balanced classes; the same sign does not mean the same quantity, which is why this paper compares the pattern and not the effect sizes. Four checks on the 21 contrasts of the prespecified family, which contains the six that survive the run-clustered gate, say the pattern is not carried by a few subjects (S23). The four contrasts whose interval includes zero — CNN vs MedFormer and EEGNet vs MedFormer on Siena, CNN vs Mamba in N1 vs N2, EEGNet vs Mamba in N3 vs REM — are reported as a direction with an effect size rather than as a confirmed difference, the same boundary the Holm results draw in §4.3.3.

#### *4.3.6 Aggregate change and the class contrast carry different information*

The preceding subsections establish that the contrast differs by encoder. They do not establish how much of it an aggregate change already accounts for, which is what an evaluation must answer before a class-level report can be said to add anything. This was tested in four steps of unequal weight: the nested-model comparison is descriptive of these eleven pairs, and the held-out and run-restricted refits are exploratory.

**What a per-class summary does give.** A cohort-level per-class summary reports the average level of each class, and the difference between two class means is the mean of the within-subject differences, up to cells missing for some subjects (largest discrepancy 1.92 pp). It also reflects the direction seen in most subjects: of 34 class-pair by encoder cells, the mean contrast had the opposite sign to the majority in 3, each within 1.8 pp of zero. A claim that per-class summaries cannot represent class-wise change would therefore be false.

**What the aggregate change does not do.** The test is the nested-model comparison of §3.5. With aggregate change alone the median $R^2$ was 0.025 (range 0.00 to 0.17); adding encoder identity raised it to 0.41, a median $\Delta R^2$ of 0.34, both in-sample. The partial F reached $p < 0.05$ in 9 of 11 comparisons and the same 9 survive Holm correction. Two checks that assume no residual covariance structure agree with it. Residual correlations between encoders within a subject have median −0.47, close to the −0.50 that subject centring imposes mechanically, so there is no sign of dependence beyond that constraint. A Monte Carlo permutation of encoder labels within each subject, 10,000 draws, gives the same 9 comparisons and the same Holm outcome; its null is joint exchangeability across encoders, which is stronger than the incremental question asked here (§3.5). A cluster-robust Wald test, the least favourable of the three readings, retains 7 of 11 on 13 and 20 clusters; the two that differ are W vs N1 and N1 vs REM, both already non-significant in §4.3.5, and the two pairs that carry the cross-cohort argument survive under all three. The analysis was specified after the contrasts of §4.3.2–4.3.5 were seen, so it is exploratory in the sense of Table 19.

Two limits apply to that comparison. The eleven comparisons are not independent: the ten Sleep-EDF pairs share subjects and encoders, so they are ten questions about one cohort, not ten replications. ΔBA also enters the model as a measured covariate with its own sampling error: classical attenuation would bias its coefficient toward zero and so inflate the share attributed to encoder identity, and that bias is not estimated here.

**Whether the increment carries to a held-out subject.** Refitting both models with one subject left out, the median increment falls from 0.34 to 0.30, positive in 10 of 11 comparisons, and 9 of 11 exceed the 97.5th percentile of a within-subject label permutation and survive Holm correction. The refit is not source-model-disjoint: 83 to 85% of its training rows come from models that had the held-out subject in training or validation, and the restricted refit that removes them drops the median to 0.02 (S26).

**The decision consequence.** Selecting an encoder by mean ΔBA gives the same answer for every class pair in a cohort (EEGNet on Siena, Mamba on Sleep-EDF), because ΔBA does not depend on the pair at all. Selecting by mean contrast gives different answers for different pairs, and on Sleep-EDF each of CNN, EEGNet and Mamba is preferred for at least one pair. With the preference reversed the winner still varies with the pair, and the same three appear. The two rules selected the same encoder in 1 of 11 comparisons, but that count has to be read against the constancy just described: the aggregate rule makes one fixed choice per cohort, so the eleven comparisons contain two such choices, and the count reports how often the pair-varying rule landed on the fixed one. This is a statement that the two criteria measure different objectives, not that one of them is wrong: no independent gold-standard encoder exists here, so the disagreement cannot be scored as an error by either rule.

**What does not follow.** The design does not show that encoder identity caused the redistribution: the four implementations differ in architecture, parameter count and optimisation, and none was manipulated, so this is a statistical association under a fixed protocol rather than a mechanism. Nor was a within-subject relation found between the size of the aggregate loss and the size of the contrast shift: after removing subject and encoder means the median within-subject correlation was +0.05, and 3 of 11 intervals excluded zero, the count expected under no association. That null is consistent with the absence of a simple monotone account and supports that narrow reading; it does not establish that the two quantities are unrelated, because a non-linear or subject-specific relation would not appear in this statistic. What the four steps support is narrower than a mechanism and sufficient for the evaluation claim: aggregate performance change and within-subject class contrasts report different aspects of an adaptation outcome, and neither substitutes for the other.

#### *4.3.7 A label-free criterion for the same choice*

The contrast needs labels, and an evaluation that has to choose an encoder online does not have them. One label-free quantity tracks it: the change in the predicted class share, $m_c$ being the fraction of samples assigned to class c, and its contrast $\Delta m_a - \Delta m_b$.

In a balanced two-class problem the two are the same quantity. With n samples per class and N = 2n, writing a for the count of class 0 predicted as 0 and d for the count of class 1 predicted as 1, the row sums fix the off-diagonal cells, so $\Delta R_0 - \Delta R_1 = (\Delta a - \Delta d)/n$ and $\Delta m_0 - \Delta m_1 = 2(\Delta a - \Delta d)/N = (\Delta a - \Delta d)/n$, an algebraic equality rather than an empirical one. On Siena, where every subject contributes exactly equal class counts (range 0 across all subjects and seeds), the two statistics agree to machine precision for all four encoders, and that agreement is a check on the implementation rather than evidence about the cohort.

With five unbalanced classes the identity fails, and the question becomes empirical. On Sleep-EDF the subject-level correlation between the two statistics is 0.70 to 0.93 across the ten stage pairs (median 0.87), and the encoder the label-free criterion selects is the encoder the contrast selects in 8 of the ten pairs, against 1 of 10 for mean ΔBA. The eight is not a point estimate: the four partitions, each a different split with a refitted source model, give 7, 9, 7 and 6 of ten, and resampling subjects gives a 95% interval of 6 to 10, against 0 to 3 for the aggregate. The direction separates the two criteria, while the exact count does not. The criterion therefore recovers, without labels, the decision the aggregate misses; it is a surrogate for the contrast rather than a validated selector, since the comparison that establishes it is itself computed from labels.

The limit lies in the ordering, not the winner: Kendall's τ between the two rankings of the three encoders averages +0.53, so the criterion identifies which encoder is best more reliably than it orders all three, and with three encoders and ten pairs the middle positions are not resolved. Two further limits follow from the construction: the statistic is computed on the same stream that is being adapted, so it is available once that stream has been scored, and it compares predicted shares, which depend on the class prior of the stream. It is a criterion for comparing encoders on one stream, not an estimate of the contrast itself. S20 reports that the absolute majority-class share does not track collapse, because an imbalanced prior moves it on its own. That result concerns a level read against a threshold, whereas here the shares enter as a difference between two encoders scored on the same stream, so the prior-determined level cancels and only the prior dependence of the change remains.

### 4.4 Dependence on the adaptation rule and what a cell mean hides

Is the deterioration reported above a property of `full_ttt`, or does it appear under other rules as well? The runs that produced Table 3 executed three further rules on the same folds, seeds and cached source models — entropy minimisation and information maximisation with a diversity term, both updating only the normalisation affine parameters, plus the bottleneck-adapter arm that shares the masked-reconstruction objective with `full_ttt`. Each cell is therefore a paired comparison in which only the objective, the parameter set and the learning rate differ.

The three additional rules all fall below source in a majority of cells, by very different amounts. Entropy minimisation is below source in ten of eleven cells, as `full_ttt` is, with a cell-level mean change of −14.1 pp against −16.2 pp; on Sleep-EDF × EEGNet it is worse (−33.9 against −25.2 pp), and on that dataset all 20 subjects worsen. Information maximisation is below source in seven cells and above in four, all four exceptions on PTB-XL, where the score sits against a ceiling (§4.2, §6), so an improvement there is not evidence that adaptation helps multi-class decoding; on the multi-class cohorts it is below source in all seven Siena and Sleep-EDF cells. The adapter arm is below source in six cells and above in five, with a cell-level mean of −0.5 pp.

Neither factor orders the four rules on its own. The two normalisation-affine rules share a parameter set and differ only in objective, yet they are 9.5 pp apart at the cell level, and one of them matches the full-parameter arm; the two masked-reconstruction rules share an objective but differ in parameter set and learning rate, so their 15.7 pp gap cannot be assigned to either. Only the `tent_ttt`–`shot_ttt` pair isolates a single factor, the diversity term. What this supports is narrower than "test-time adaptation fails", and this paper claims only that narrower statement: deterioration of the size reported in §4.1 is not a property of masked-feature reconstruction alone, since a rule with a different objective, a different learning rate and a different updated parameter set degrades on the same cells. It does not support a statement about test-time adaptation in general — four rules on three datasets is a sample of implementations, and the PTB-XL exceptions show that the sign is not universal.

A cell mean also hides which of two opposite regimes produced it (Table 17). Under the adapter arm, all 13 Siena subjects stay within 1 pp of their source score, the largest change being 0.70 pp, so the rule barely moves the model and its near-zero mean reports inactivity rather than safety. In Sleep-EDF × CNN the same arm moves 9 of 20 subjects by 5 pp or more, and the mean change of −2.4 pp is the residue of a mean absolute change of 5.0 pp, a net-to-gross ratio of −0.48. That cancellation is not a property of the rule: S8 reports the same arm at −4.7 pp with 19 of 20 subjects falling, once the stream is cut at subject boundaries and interleaved. Table 18 resolves the two cells at subject level; a cell mean cannot separate these cases, and neither supports reading the adapter arm as the milder rule.

The three rules that degrade most take the uniform form instead. Under `full_ttt`, entropy minimisation and information maximisation, every one of the 20 Sleep-EDF subjects falls, by medians of 24.2, 23.0 and 21.5 pp, and no subject stays within 1 pp; on Siena the counts are 11, 12 and 11 of 13. Averaged over the four splits, 12 of the 20 Sleep-EDF subjects fall under all four rules and none is left unharmed by all four, and on Siena the counts are 2 and 1 of 13 — the difference following from the adapter arm being inert there rather than from any subject being protected by it. The rule therefore changes the shape of the subject-level distribution, not only its level. The two cells are separate cohorts and their counts are not pooled; as in the comparison above, the rules differ in learning rate and in which parameters they update, so this is a comparison between algorithms and not between objectives alone.

Three further boundary analyses are reported in the supplementary material: whether the harm falls on the subjects who start low (S9), when in the stream the change happens (S10), and whether it persists in the final adapted model (S11).

## 4.5 Whether adaptation state is shared decides the sign of an order effect

The subject-level results above were measured on one stream construction: all held-out subjects of a fold adapted as one continuous stream. Two properties vary independently — whether the subjects' samples arrive in blocks or interleaved, and whether the adaptation state is shared across subjects or reset at each subject boundary — giving four constructions, each run with four seeds. The full per-subject comparison and the variance decomposition are in S7–S8; what follows is the part that sets the scope of both claims.

For the full-parameter rule the answer does not depend on any of them. The cell mean is −23.3 pp with a shared contiguous stream, −22.9 shared and interleaved, −23.0 isolated and contiguous, and −22.8 isolated and interleaved, and 20 of 20 subjects fall in all four (Table 9). The subject share of the variance is 0.56 and the construction share 0.00, so individual differences in how much a subject loses are reproducible while the organisation of the stream moves the cell mean by 0.5 pp. The first result is therefore not an artefact of the shared stream.

For the adapter arm, construction decides the answer. Its four means are −2.4, +1.1, −2.3 and −4.7 pp, and the interleaving shift is +3.5 pp when the stream is shared and −2.4 pp when each subject is adapted alone. Because the level sits near zero, that reversal moves the count of subjects who fall from 12 of 20 to 4 of 20 in one direction, and from 11 to 19 in the other. The sign of the interleaving effect therefore depends on whether cross-subject state is shared, which rules out batch composition as a sufficient explanation: a purely compositional effect would carry one sign in both constructions. The variance decomposition points the same way, a construction share of 0.56 against a subject share of 0.15 and an interaction share of 0.01, so the shift is close to a common offset rather than a reordering.

This is a boundary on what an order result can claim. An order effect measured under one construction does not transfer to the other, so a study reporting one of them cannot say whether it measured the stream or the protocol. The adapter's near-zero mean (§4.4) holds only for contiguous streams; cut and interleaved it falls to −4.7 pp with 19 of 20 subjects worsening. What survives all four constructions is that arm's inertness on Siena, where the largest change is 0.65 pp.

## 4.6 Boundary analyses in the supplement

Four analyses follow the two results above. They are printed in the supplement because they ask where in the model the change sits instead of how it is distributed, and they are neither a fourth contribution nor one chain.

S14 times the write-back of input-stage parameter displacement in one cell (Sleep-EDF with Mamba, 20 subjects, five folds), with a 2 × 2 design that allows or suppresses that write-back in an early and a late window while gradients, Adam moments and every non-input parameter keep evolving. The late-stream carry contrast is −11.50 pp (95% CI −12.65 to −10.39; exact p = 0.0020; direction agreed in all five folds), and its precision is set by five folds rather than by the ten subjects the family counts. It is the same 20 subjects as the earlier descriptive estimate of −14.17 pp, so this is not replication in an independent cohort: the two are not pooled and the second does not retroactively confirm the first, though the agreement in direction provides descriptive support for the sign of the earlier estimate. Exchanging the parameter state at the window boundary reproduces −11.38 pp of it, while exchanging the remaining optimiser state moves the same quantity by −0.10 pp; that decomposition is descriptive, because its admissibility criterion was resolved only after the run.

S15 measures the class-conditional geometry of the representation that intervention acts on, in the same cell. The Fisher ratio D = $\mathrm{tr}(S_B)/\mathrm{tr}(S_W)$ falls from 1.077 to 0.720 (Δ −0.357, 20 of 20 subjects), driven by the between-class term falling about five times faster than the within-class term. N3 accounts for 74% of that fall, but both of its own terms halve, so this is the largest class contributing the largest absolute change, not one class losing separability. The fixed head's macro-recall falls from 0.4367 to 0.2889 and its predictions concentrate on N3 in 20 of 20 subjects.

S16 replaces the readout with one fitted on each fold's training subjects, separately for the two conditions. The gap between conditions falls from 0.1478 to 0.0117 (paired reduction +0.1361, 20 of 20 subjects, exact p = $1.9 \times 10^{-6}$), and prediction concentration falls from 0.606 to 0.287. The refit uses labels, so it is not a label-free procedure; the recovery of the condition gap lends qualified support to a readout mismatch as its main locus, but it does not establish that the class information is fully retained.

S17 aligns the feature geometry with affine maps estimated from unlabelled statistics. Anisotropic second-order alignment adds nothing over a mean shift with a single scale (−0.0010, 95% CI −0.0042 to +0.0022, p = 0.546), the spectrum-matched random orientation scores below doing nothing, and a post-hoc mean shift alone matches the full map. For a linear head a translation is a class-bias correction, so this experiment cannot separate a displaced feature distribution from a mismatched bias.

Beyond those four analyses, S27 takes the S16 readout one step further: it crosses the two feature conditions with the two refitted readouts, so that their separate contributions and their interaction can be read apart in one cell. With the encoder and the feature archive held fixed, the two readouts exchanged rank between feature conditions; the interaction was +0.2146 in 20 of 20 subjects (Holm-adjusted p = $5.72 \times 10^{-6}$ under joint sign exchangeability, from a separately frozen subject-level test). A second cohort of 28 subjects, none of whom entered readout fitting, was then scored under frozen development-set readouts: the interaction was positive in 28 of 28 (mean +0.1628), and the rank reversal reappeared in 25 of 28 against a prespecified descriptive criterion of 0.80. Those 28 subjects share five donor states and the same fitted readouts, which is where their remaining dependence sits. Their readouts were fitted on all 20 development subjects, where the first cohort's were fitted on 16 training subjects per fold. The quantity here varies the readout and the feature condition at one encoder; §4.3 measures how the within-subject class contrast varies across encoders. S27 gives the fold-level enumeration resolution of the subject-level test.

# 5 Discussion

The result placed first here concerns the standard summary, not the size of the performance change: it cannot say which class stopped being reported. In the binary case a decoder that emits one class and one that reports both classes at 50% receive the same balanced accuracy, and the first of those states occurred in 44 of 112 subject-cells against 9 for the source model (§4.1). On Sleep-EDF the aggregate was close to uniform across three encoders and every subject worsened, yet the class contrasts within the same cells differed by up to 56.3 pp and changed sign; on Siena the within-subject contrast reversed sign between Mamba and the other three encoders. A mean change, and even the direction of every subject's change, can be shared while the redistribution across classes is not. The two quantities are not in a refinement relation: they answer different questions, and only one of them is the question a clinic asks when a stage stops being reported. Adaptation did not produce one performance change; its effects differ across classes and, in direction, across encoders (§4.3), across subjects (§4.2), across stream organisations (S19, S7–S8), and with the time of scoring (S10–S11). The subject-level distribution is the substrate for the main claim rather than a finding in its own right: it fixes the individual as the unit in which the class contrast must be formed, and it is not uniform. On Sleep-EDF a mean of −23.3 pp describes nearly every subject; on PTB-XL a comparable −14.6 pp describes almost nobody, the median being zero with a long lower tail. The same statistic therefore licenses opposite statements about individual reliability, and a collapsed stage is not a small loss in a scalar but a category that stops being reported. The two results do not carry the same evidence: the first is measured, while the second is partly identified — one process is measured at the output level and dated against the loss without being shown to drive it, the other is located but not explained. The analyses that follow delimit them, and none is offered as a further finding.

The results show two processes, which answer different questions. The first is a loss of discriminative output diversity: predictions concentrate on fewer labels and at the extreme on one. Its clearest signature is the end of the run, where three of the four re-scored cells return almost one label and the terminal score sits at chance (S11, Table 11) — an output-level degeneracy, not a collapse of the feature representation and not a stable fixed point. The same process appears earlier: subjects arriving after a run has concentrated receive identical predictions in the shared and the isolated construction (S7), and in three cells almost the whole drop is complete inside the first fifth of the stream (S10). It is observed at its end and in fifths of the stream, not in the features of the main runs, which retained no feature stream; the one cell instrumented later (S15) shows a contraction of class geometry, not output diversity following from feature diversity. Nor does this paper claim that a single-label predictor is a stable fixed point: the objective assigns zero loss to a constant feature map, so nothing in it opposes concentration, but a zero-loss state, an observed terminal state and a stable attractor are three claims and only the second is measured. The second process redistributes class-specific performance. With dataset, split, samples and update rule fixed, the direction of the within-subject class contrast follows the encoder. On Siena the four contrasts differ (Friedman $p = 0.0001$), three of six pairwise differences survive the prespecified test and two a two-way cluster covariance at its nominal reference, one at a fold-level reference, and Mamba's contrast is positive in all thirteen subjects (§4.3). Two processes would produce that — a change in representational separability, or a class-specific update magnitude. The first has since been measured in one cell (S15); the second has not been measured anywhere, and a separability measured on the source model would describe the geometry the update started from, not the one it produced. What the result does establish is that a class-level statement about adaptation is incomplete without naming the encoder, and the observed contrast directions support that reading; it does not establish a clinical consequence, since the stage labels carry task semantics but per-class recall is not a patient outcome and no outcome was measured here.

The two processes do not carry equal weight here. Concentration accounts for the terminal state: once a run returns one label, balanced accuracy is 1/K, and the near-identical terminal scores of three cells follow from that identity, not from residual skill. It does not account for the size of the change, which S11 finds falling first at least as often as the predictions concentrate, nor for direction, since a single-label predictor sends every class but one toward zero recall. Redistribution accounts for direction, but not for distance. Neither accounts for the size of the change, which this study measures without explaining it. Five conditions determine where each process is visible. One is the distance between the source score and chance: every cell within 5 points of chance has a non-negative stream-organisation contrast (S19), which bounds the measurement instead of showing that a low starting score protects a subject (S9). Another is which parameters may move: the adapter arm is below source in 6 of 11 cells against 10 of 11 (§4.4), though the arms also differ in learning rate. A third is the objective, where `tent_ttt` and `shot_ttt` share a parameter set and a learning rate and their 9.5 pp gap is the one contrast that isolates the diversity term. Stream organisation is a fourth: cutting at subject boundaries leaves the full-parameter result intact but moves the adapter's level and the identity of who gains and loses (§4.5, S7–S8). Position in the stream is the fifth (S10). None of these conditions was manipulated independently, so the list locates where each process binds without measuring its strength.

The practical form of this is a reporting requirement, weaker than it may sound: this paper proposes no monitor, and nothing here shows that a class-level indicator would have flagged the change earlier than the aggregate did. The results show that a criterion stated on balanced accuracy alone is insensitive to a failure mode that occurred in these runs. In three of the four re-scored cells the final model returned one label for almost every sample (S11), and on PTB-XL × CNN class-0 recall fell from 66.5% to 0.08% while class-1 recall barely moved, so the sample-level change was 31.6 points against a 14.6-point subject-equal mean (§4.3.1). Per-class change therefore belongs next to the aggregate, the distribution next to the mean, and the end-of-streaming score next to the average, because the last two disagree in both directions across the four re-scored cells; the dependence of the subject-level change on the run that carried it belongs beside them. It does not follow that a system should be rejected whenever a class-level change is present: which class matters is a clinical judgement this study does not supply. Two further limits bound what can be claimed here. Of the two measurements that would turn the second process from a located effect into an identified one, the class-conditional representations now exist for one cell (S15); per-class update magnitudes measured during adaptation do not exist for any cell. And the cohort sets the precision of every subject-level statement: 13 and 20 subjects. Until both change, the defensible claim is conditional rather than mechanistic — unlabelled adaptation produces heterogeneous and sometimes opposing changes, at least two processes are distinguishable in the results, and the conditions under which each binds are mapped here but not explained. The boundary analyses of S14–S17 are read the same way: they name a location in one cell, they do not compose a mechanism, and the one family of maps tested as a candidate cause returned a null result (S17).

# 6 Limitations

The 95% intervals of §4.2 are percentile subject-cluster bootstrap intervals for four-seed subject-average changes: conditional on the four split/model instances, they exclude split and source-training variability and do not establish transport to a new cohort. The sign tests treat the subject as the unit; subjects reappear across seeds and across cells within a dataset, and folds, seeds and permutations are never counted as additional subjects. The same limit applies to the encoder differences of §4.3.3 and the stage-pair contrasts of §4.3.5, where run-level dependence is measured beside the test rather than removed, and 13 and 20 subjects cannot exclude it. Those axes are crossed rather than nested, so the two-way reading separates Mamba from two encoders rather than three, which is the count carried forward.

The subject-level tests are conditional on the stream construction: held-out subjects of a fold are adapted as one continuous stream (§3.4), so an observation is not a draw from a process that adapted on that subject alone. Under the full-parameter rule the result survives cutting the stream at subject boundaries; for the adapter arm on Sleep-EDF neither the level nor the identity of the subjects who gain and lose survives, so those counts describe the evaluated run. For Sleep-EDF × CNN the most adverse reading puts the Holm-adjusted p at 0.094, which 20 subjects cannot exclude, while the other five cells that survive correction stay at or below 0.002 under the same bound. On the class contrast the audit does detect shared-run structure, in two of four encoders, and under that reading two rather than three contrasts survive, though the reversal does not rest on the one that does not. The nested-model comparison treats eleven class-pair comparisons as separate units although the ten Sleep-EDF pairs share subjects and encoders, so its medians are descriptive and its inference is reported under three readings rather than the most favourable one: the within-subject permutation test and the partial F agree on 9 of 11 after Holm, and the cluster-robust Wald test retains 7. Restricting the refit to training subjects sharing no run with the held-out one lowers the median increment from 0.30 to 0.02 against 0.25 for a size-matched random draw, so the increment belongs to subjects adapted together; N3 vs REM is the exception.

PTB-XL supports no subject-level balanced-accuracy outcome, because each subject contributes one class; recall of that class is reported. Its class-stratified results are between-subject group contrasts, confounded by subject-group characteristics, and the majority-class group's source recall is already near 1.0, so a zero change is not evidence of no adaptation effect. That ceiling-linked profile does not generalise to the multi-class-per-subject datasets.

This study cannot identify the cause of the Siena class reversal: the source prediction distribution does not explain it, and representational separability and per-class update size were not measured, so these data do not favour one mechanism over the alternatives.

The follow-ups of S15–S17 are confined to one cell (Sleep-EDF with Mamba), one representation — the 64-dimensional readout input — and one randomisation for S17. The S16 readout is fitted on labelled training subjects of the target condition, so it is not label-free; its gain is consistent with a recoverable readout mismatch and lends support to that narrower reading, but it does not establish that class information is fully retained: `N3` recall falls after refitting. The S17 alignment covers affine maps and second-order statistics only, and that the recoverable part is a translation is also why it cannot separate a displaced feature distribution from a mismatched class bias, the two being one operation for a linear head. The S14 decomposition is descriptive: its admissibility criterion was resolved only after the run, and ten late-stream subjects set its precision. The dependence audit is itself limited — the intraclass correlation rests on five clusters — and no reading here removes the overlap between folds, whose training sets share 12 of 16 subjects.

The subject-level profiles are examples, not a taxonomy. Siena (13 subjects) and Sleep-EDF (20) give small-cohort within-subject evidence that needs independent confirmation: the 95% subject-cluster half-width is 3.8–7.4 pp for the two Siena cells with a negative mean and 3.0–3.7 pp for the Sleep-EDF cells, so a smaller change cannot be separated from zero here. Additional seeds do not help, because a seed redraws the split without adding subjects, and halving these margins would take about four times the subjects. For one component-level quantity in one cell, S27 reports a second cohort: 28 subjects who entered no readout fit had a positive readout–feature interaction in 28 of 28, all scored with five development-set readouts shared by the whole cohort. The four encoders span three architecture families and show variation among the evaluated implementations rather than predicting a new architecture.

The study evaluates measured outcomes under a fixed protocol, not a deployed system: no adapted model was run online and no system-level outcome was measured, so the contribution is to what an evaluation report must contain rather than to what a decoder can achieve. This paper evaluates two label-free indicators and three prespecified strategies but proposes no new detector; the negative results are specific to these indicators, implementations, criteria and cohorts. They provide descriptive support for the narrower claim that these particular indicators and strategies did not detect the degradation, but do not establish that label-free monitoring is generally infeasible. This paper also does not evaluate the framework of [12] on its benchmark data or test its formal assumptions.

# 7 Conclusion

Aggregate balanced accuracy cannot say which class stopped being reported: in the binary case a decoder that emits one class and one that reports both classes at 50% receive the same value. Nor does it say how the deterioration is distributed: a broadly distributed decline on Sleep-EDF and a zero-median, long-tailed pattern on PTB-XL produced the same kind of negative average. On Siena the direction of class-conditional change followed the encoder, reversing between Mamba and the others, and the encoder-to-encoder differences were tested directly instead of being inferred from separate tests against zero. Stream organisation shifted outcomes without a common direction; the two label-free monitors tested in S20 could not separate a collapsed model from a healthy one, while the class-share criterion of §4.3.7 tracked the contrast in 8 of 10 pairs without labels — a surrogate for the contrast, not a substitute for the monitor. For stream-based TTA under this protocol, the requirement is to report the subject-level distribution and class-conditional behaviour alongside the aggregate, and to treat the encoder as part of the result. That relation is within-cohort under this fold structure and does not carry to unseen subjects; what travels is the need to report the encoder, not the size of its contribution (§4.3.6). The held-out-subject check of S27 measures a different quantity: the interaction of feature condition with a refitted readout, with the encoder held fixed at Mamba in one cell. Both recommendations are checkable from archived per-subject predictions, which is what makes them evaluation criteria rather than caveats. Two follow-up analyses in one cell narrow the interpretation without closing it: replacing the readout recovers most of the gap between conditions, while aligning the feature geometry beyond a translation recovers nothing — and a translation is what a linear readout absorbs as a change of bias. The recommendation applies to post-update, batch-level transductive evaluation, where a batch is scored with the model already updated on it; a strictly prequential protocol scores the incoming batch before any update and is a different target.

## Data and code availability

All five datasets are publicly available through the sources cited in §3.2: Siena [13], Sleep-EDF [14], BCI-IV-2a [15], CHB-MIT [16] and PTB-XL [17]; no new human data were collected. A reproducibility package accompanies the submission and contains the per-fold prediction archives with subject identifiers, the subject-level and class-level statistics, and the audit and figure scripts, so that the reported sign tests, Holm corrections and intervals can be recomputed without retraining. The package includes SHA-256 checksums for all archived payload files and a manifest of the files it manages; both metadata files are excluded from the checksum list itself. Model weights are not included, and the adapted weights of the order runs remain unrecoverable; both omissions are stated in S12 rather than filled with defaults. Because the study is a secondary analysis of previously published, de-identified public datasets, no new ethical approval or informed consent was required; users of these datasets remain bound by their original terms of use.

## References

[1] Lili Wang, Jing Li, Xiaowen Sun, Xiangyu Hu, Zhuangzhuang Gu, Jian Liu, Srihari Nelakuditi, Yan Tong. Should This Case Be Adapted? Prediction Fragmentation Controls Test-Time Adaptation. arXiv:2609.20700 (preprint, September 2026).

[2] Hritik Arasu, Faisal R. Jahangiri. StableSleep: Source-Free Test-Time Adaptation for Sleep Staging with Lightweight Safety Rails. arXiv:2509.02982 (preprint, September 2025).

[3] Jisu Han, Jaemin Na, Wonjun Hwang. Ranked Entropy Minimization for Continual Test-Time Adaptation. ICML 2025; arXiv:2505.16441.

[4] Gabriel Jason Lee, Jathurshan Pradeepkumar, Jimeng Sun. Test-Time Adaptation for EEG Foundation Models: A Systematic Study under Real-World Distribution Shifts. Proceedings of the 11th Machine Learning for Healthcare Conference, PMLR 340:948–984, 2026; arXiv:2604.16926.

[5] Bogdan Kozyrskiy, Artem Grachev, Abraham I. Camelo Guerrero. Benchmarking Label-Revealed Online Updates for EEG BCI Decoding. arXiv:2610.07420 (preprint, October 2026).

[6] Hao Zhao, Yuejiang Liu, Alexandre Alahi, Tao Lin. On Pitfalls of Test-Time Adaptation. ICML 2023; arXiv:2306.03536.

[7] Motasem Alfarra, Hani Itani, Alejandro Pardo, Shyma Yaser Alhuwaider, Merey Ramazanova, Juan Camilo Perez, Zhipeng Cai, Matthias Müller, Bernard Ghanem. Evaluation of Test-Time Adaptation Under Computational Time Constraints. Proceedings of the 41st International Conference on Machine Learning, PMLR 235:976–991, 2024; arXiv:2304.04795.

[8] Ziqiang Wang, Zhixiang Chi, Yanan Wu, Li Gu, Zhi Liu, Konstantinos N. Plataniotis, Yang Wang. Distribution Alignment for Fully Test-Time Adaptation with Dynamic Online Data Streams. ECCV 2024; arXiv:2407.12128.

[9] Young Kyung Kim, Oded Schlesinger, Qiangqiang Wu, J. Matías Di Martino, Guillermo Sapiro. Order-Aware Test-Time Adaptation: Leveraging Temporal Dynamics for Robust Streaming Inference. arXiv:2601.21012 (preprint, January 2026).

[10] Kartik Jhawar, Lipo Wang. When is Test-Time Adaptation Identifiable From Unlabeled Evidence? arXiv:2609.11235 (preprint, September 2026).

[11] A. Philip Dawid. Statistical theory: The prequential approach (with discussion). Journal of the Royal Statistical Society, Series A 147(2):278–292, 1984.

[12] Mona Schirmer, Metod Jazbec, Christian A. Naesseth, Eric Nalisnick. Monitoring Risks in Test-Time Adaptation. Advances in Neural Information Processing Systems 38 (NeurIPS 2025); arXiv:2507.08721.

[13] Paolo Detti. Siena Scalp EEG Database (version 1.0.0). PhysioNet, 2020. https://doi.org/10.13026/5d4a-j060. Original publication: Paolo Detti, Giampaolo Vatti, Garazi Zabalo Manrique de Lara. EEG synchronization analysis for seizure prediction: a study on data of noninvasive recordings. Processes 8(7):846, 2020.

[14] Bob Kemp, Aeilko H. Zwinderman, Bert Tuk, Hilbert A. C. Kamphuisen, Josef J. L. Oberye. Analysis of a sleep-dependent neuronal feedback loop: the slow-wave microcontinuity of the EEG. IEEE Transactions on Biomedical Engineering 47(9):1185–1194, 2000.

[15] Clemens Brunner, Robert Leeb, Gernot Müller-Putz, Alois Schlögl, Gert Pfurtscheller. BCI Competition 2008 – Graz data set A. Graz University of Technology technical report, 2008.

[16] John Guttag. CHB-MIT Scalp EEG Database (version 1.0.0). PhysioNet, 2010. https://doi.org/10.13026/C2K01R. Original study: Ali Shoeb. Application of Machine Learning to Epileptic Seizure Onset Detection and Treatment. PhD thesis, Massachusetts Institute of Technology, 2009.

[17] Patrick Wagner, Nils Strodthoff, Ralf-Dieter Bousseljot, Dieter Kreiseler, Fatima I. Lunze, Wojciech Samek, Tobias Schaeffter. PTB-XL, a large publicly available electrocardiography dataset. Scientific Data 7:154, 2020. https://doi.org/10.1038/s41597-020-0495-6.

[18] Albert Gu, Tri Dao. Mamba: Linear-Time Sequence Modeling with Selective State Spaces. arXiv:2312.00752, 2023.

[19] Yihe Wang, Nan Huang, Taida Li, Yujun Yan, Xiang Zhang. Medformer: A Multi-Granularity Patching Transformer for Medical Time-Series Classification. NeurIPS 2024; arXiv:2405.19363.

[20] Vernon J. Lawhern, Amelia J. Solon, Nicholas R. Waytowich, Stephen M. Gordon, Chou P. Hung, Brent J. Lance. EEGNet: a compact convolutional neural network for EEG-based brain–computer interfaces. Journal of Neural Engineering 15(5):056013, 2018; arXiv:1611.08024.

[21] Jian Liang, Dapeng Hu, Jiashi Feng. Do We Really Need to Access the Source Data? Source Hypothesis Transfer for Unsupervised Domain Adaptation. ICML 2020; arXiv:2002.08546.

[22] Dequan Wang, Evan Shelhamer, Shaoteng Liu, Bruno A. Olshausen, Trevor Darrell. Tent: Fully Test-Time Adaptation by Entropy Minimization. ICLR 2021; arXiv:2006.10726.

[23] Thilina Weerasinghe, Ruwan Tennakoon, WeiQin Chuah, Alireza Bab-Hadiashar. Logit-Adjusted Test-Time Adaptation under Partial Class Imbalance. WACV 2026.

[24] Minguk Jang, Hye Won Chung. Label Distribution Shift-Aware Prediction Refinement for Test-Time Adaptation. arXiv:2411.15204 (preprint, 2024; revised 2025).

[25] Shuaicheng Niu, Jiaxiang Wu, Yifan Zhang, Zhiquan Wen, Yaofo Chen, Peilin Zhao, Mingkui Tan. Towards Stable Test-Time Adaptation in Dynamic Wild World. ICLR 2023 (Notable-Top-5%); arXiv:2302.12400.

[26] Suli Wang, Yangshen Deng, Zhenghua Bao, Xinyu Zhan, Yiqun Duan. NeuroTTT: Bridging Pretrain–Downstream Task Misalignment in EEG Foundation Models via Test-Time Training. arXiv:2509.26301 (preprint, 2025).

# Supplementary Material

Section numbers S1–S27 are the ones cited in the main text.

## S1 What the archive supports

"Reproducible" is used loosely, and three levels of it are not interchangeable.

- *Recomputable from the package*: the per-fold predictions, subject identifiers, sign tests, Holm corrections and subject-cluster intervals. These can be re-derived without retraining and are the basis of every inferential statement in the paper.
- *Reproducible as a command, not documented per run*: `lr_full` = 1e-4, `steps` = 1, `batch` = 64 and the source-training defaults. The launch script, the code defaults and the source-cache key agree on all four, but no per-run snapshot was written, so a reader can re-issue the command and cannot prove that a historical run used those values.
- *Unrecoverable*: the adapted weights of the order runs. They were not saved and not hashed, so the adaptation must be re-run. The four cells of S11 are the exception: the final state of each arm was saved and its state digest recorded, so those runs can be re-scored without retraining. Those 320 checkpoints total 386 MB and are not shipped; the package carries their digests and the terminal-state predictions instead.

Permutation identity is recorded as a hash rather than as an array: the original order carries `perm_hash = identity`, and the three permuted orders carry distinct hashes with fixed order seeds. Two runs can therefore be checked for using the same permutation, although rebuilding an array requires the same generator and call sequence. Block-preserving orders exist for 67 of the 268 groups, so that comparison is available only where they were run.

The package's `README.md` is its manifest: it records source-file hashes, the run fields, test-subject lists, the adaptation seed and the evaluation protocol. With the source model cached, one fold takes a median of 14 s for all five arms; the first pass over a cell also trains the source model and reaches about 13 minutes.

## S2 How the final state was captured and checked

The checkpoint of a run holds the model parameters, the normalisation statistics and, where an arm has one, its adapter. Its SHA-256 digest is computed over keys, dtypes, shapes and contents rather than over the bytes of the saved file, so a re-serialisation does not change it. The optimiser state is not stored, so a digest over the parameters, the buffers and the optimiser state is taken immediately before and immediately after the scoring pass and compared; that comparison establishes that the pass did not keep adapting. Predictions are recomputed in a single pass without gradients and without any update, and the labels are stored alongside them, so every terminal-state score in Table 11 is re-derivable from the archive rather than transcribed from a log. The recomputation calls each arm's own forward path, because the adapter arm applies its adapter on one path only and a default forward would have scored the source model instead. The `source_only` rows are the internal control: their two scores are equal by construction, and the equality is reproduced in all 20 runs of each cell.

## S3 Dating output concentration against the loss

The comparison is made on ten equal blocks per stream. The concentration rule asks when the most frequent predicted label first reaches 0.8; the drop rule asks when balanced accuracy first falls 10 pp below the source score in the same block. Both events are observed in 229 of the 268 (seed, fold) units: concentration first in 97, the same block in 58, the score first in 74.

That split is not a property of the mechanism, for three reasons. In the two binary tasks the source model already places 0.85 to 0.88 of its predictions on one label, so the absolute threshold is met in the first block before adaptation begins, and 94% and 97% of blocks in PTB-XL × EEGNet and PTB-XL × Mamba satisfy it on the source predictions alone. In the six cells where the source model leaves headroom, concentration leads in 13 of 108 units, while 48 are simultaneous and 47 fall first. A paired rule that asks for the same margin over the source model's own concentration gives 77 simultaneous units of 167, 67 falling first and 23 concentrating first. The comparison is also asymmetric by construction, because the drop rule is a margin over the source score while the concentration threshold is absolute, so the sign of their difference mixes timing with distance to threshold.

## S4 Subject jackknife and threshold sensitivity

Removing any one subject leaves both subject-level cells where they were. Across the 20 jackknife fits for Sleep-EDF × CNN the cell mean stayed between −24.0 and −22.6 pp, and every fit remained significant after correction (largest Holm-adjusted p = $1.4 \times 10^{-4}$). Across the 13 fits for Siena × CNN it stayed between −20.3 and −16.6 pp; that cell is not significant at the full sample either (Holm-adjusted p = 0.449), so for it the jackknife result is stability of a description rather than of a rejected null. The direction is stable in both cells: every fit leaves 19 of 20 Sleep-EDF subjects negative and 10 or 11 of 13 Siena subjects negative, and no single removal flips the cell's sign. These fits are a sensitivity analysis, not a second family of tests.

Below −20, the PTB-XL threshold curve is nearly flat (17.9% at −40 and 14.5% at −80), whereas the Sleep-EDF curve falls to 8.8% at −40 and zero at −50. The shares are not ranked across datasets, because PTB-XL uses each subject's class-specific recall change while Siena and Sleep-EDF use subject-level BA.

## S5 The four arms in full

Write $p_k(x)$ for the softmax output of class `k` on sample `x`, and $q_k$ for its batch mean, $q_k = (1/|B|)\ \Sigma_x\ p_k(x)$. `tent_ttt` updates only the scale and shift parameters `γ, β` of the encoder's normalisation layers, by one Adam step on the mean conditional entropy

$$\mathbf{L_{tent} = -(1/|B|)\ \Sigma_x\ \Sigma_k\ p_k(x)\ \log\ p_k(x)} \quad \text{(S1)}$$

`shot_ttt` keeps that parameter set and replaces the objective with the conditional entropy minus the entropy of the batch mean,

$$\mathbf{L_{shot} = L_{tent} - \lambda\ \Sigma_k\ q_k\ \log\ q_k}, \text{ with } \lambda = 1 \quad \text{(S2)}$$

so that the second term rewards a batch whose average prediction is close to uniform and thereby penalises the degenerate solution in which every sample receives the same label. With $\lambda = 0$ the objective is identical to $L_{tent}$ under the same optimiser, parameter set and learning rate. Both rules call the encoder in evaluation mode and leave the normalisation statistics untouched, so only γ and β move.

`adapter_ttt` keeps the reconstruction objective and inserts a bottleneck adapter at the encoder input, `x → x + U V x`, where `V` is an `r × C` matrix and `U` is a `C × r` matrix. `r` is chosen so that the adapter holds about 4,096 parameters, and `U` is zero-initialised so that the adapter starts as the identity and every change is attributable to adaptation.

## S6 Aggregation and resampling detail

The three shuffled permutations are averaged within a record before records receive equal weight, so a record with three permutations does not count three times. This record-level aggregation describes the evaluated runs and is not used as an inferential sample size.

The subject-cluster intervals resample unique subject IDs 50,000 times with RNG seed 20261104, taking the percentile interval of the resampled means. Each subject enters once, after that subject's four seed-specific differences have been averaged.

Effect sizes are reported as paired changes in percentage points, with means, medians and subject counts.

## S7 Does the subject-level result depend on the shared stream?

If the subject-level counts were an artefact of the shared stream (§3.4), cutting it at subject boundaries should remove them. The two subject-level cells were re-run with the stream cut, so that every subject starts from the source weights with its own optimiser and its own normalisation statistics, and the two constructions were compared subject by subject. Both use the same source weights, splits and samples; only the stream boundary differs, and each subject's change is averaged over the four splits before the comparison. The source-only predictions are identical in the two constructions in all 40 (seed, fold) records, which shows that the samples and their labels are aligned.

Table 7 reports the comparison, and under `full_ttt` the subject-level result is unchanged: the Sleep-EDF count is 20 of 20 in both constructions and the Siena count 11 of 13, no subject changes sign, the rank correlations are 0.96 and 0.94, and the cell means differ by 0.3 and 0.4 pp. Seven of the 20 Sleep-EDF subjects and five of the 13 Siena subjects receive identical predictions in the two constructions, because the full-parameter update has already collapsed the model to a constant prediction before their samples arrive. The finding that this rule harms essentially every subject therefore does not depend on cross-subject state sharing.

Under the adapter arm the aggregate survives this cut but the individual ordering does not. In Sleep-EDF the count of subjects who fall moves only from 12 to 11 and the cell mean from −2.4 to −2.3 pp, so cutting the stream at subject boundaries alone does not disturb the reading of §4.4 — a near-zero mean built from gains and losses rather than from inactivity. But nine of the 20 subjects change sign, and the rank correlation between the two constructions is 0.30, which is not distinguishable from zero at this sample size ($p = 0.20$). A null of that size does not establish independence; it supports only the narrower reading that no reliable individual ordering was observed under this arm, so which subjects gain and which lose is not recoverable from a single run. In Siena the same arm is inert in both constructions: the largest change is 0.65 pp, six of 13 subjects are unchanged, and the count of subjects who fall moves from 2 to 3. The stability of the Sleep-EDF aggregate is itself limited to contiguous streams: S8 repeats the cut with the samples interleaved, and there the same arm falls to −4.7 pp with 19 of 20 subjects worsening.

Three statements must be kept apart in what follows. At the aggregate level, the adapter's mean effect is close to zero in both cells and stays close to zero when the stream is cut at subject boundaries, though not when it is cut and interleaved (S8). At the level of the individual distribution, the same cell mean is compatible with very different per-subject outcomes, and in Sleep-EDF the two constructions disagree about which subjects those are. At the level of stability, the widespread deterioration under `full_ttt` reproduces under both constructions, whereas the adapter's attribution of gain or loss does not.

The consequence of this check is narrowly bounded: the claim that the full-parameter rule harms nearly every subject survives the change of construction, and so does the claim that the adapter is inert on Siena. Two things must be withdrawn, and both concern the adapter arm in Sleep-EDF: its level and the identity of the subjects who gain and lose. Neither describes a stable per-subject tendency, so the counts for that arm in Table 18 are properties of the evaluated run and of the stream construction that produced it. Every other subject-level result here is reported under the shared-stream construction and inherits that conditioning. Two limits apply to this check: it covers the two cells analysed at subject level, not the other nine, and it is a sensitivity test of the evaluation protocol, not a performance result.

Class deletion was recounted on the archived isolated-stream predictions using the criterion of §4.1: an eligible subject has at least two true classes covering a tenth of its samples each, and a deletion is counted when one of those classes receives no prediction in all four seeds. The two cells with isolated runs contain 33 eligible subjects. Cutting the stream at subject boundaries gives 25 of 33 deletions in the contiguous arrangement, against 24 of 33 under the shared stream; the source model accounts for 9 of 33 in both. On Sleep-EDF with CNN the count is 20 of 20 in all eight construction-by-arrangement combinations. On Siena with CNN it is 4 of 13 in the shared contiguous stream and 5 of 13 in the isolated one; across the three interleaved orders the shared counts are 2/1/1 and the isolated counts 6/7/7. The isolated runs start from the same source weights and use the same samples, with a fresh model copy, optimiser and normalisation statistics for each subject. These two cells account for 24 of the 44 deletions reported in §4.1; isolated runs are archived for these two cells.

**Table 7. Subject-level change under two stream constructions (pp).** Shared is the archived construction, in which all held-out subjects of a fold form one continuous stream; per-subject cuts the stream at subject boundaries so that each subject starts from the source weights with a fresh optimiser and fresh normalisation statistics. Both columns use the same source weights, splits and samples, and each subject's change is averaged over the four splits. Falls counts subjects whose mean change is negative. Sign changes counts subjects whose change crosses zero between the two constructions. Identical counts subjects whose two values differ by less than 0.01 pp. Rank correlation is Spearman's ρ between the two per-subject orderings; with 20 and 13 subjects, ρ must exceed about 0.45 and 0.56 respectively to differ from zero at $p < .05$.

| Cell | Rule | Falls, shared → per-subject | Mean δ (pp), shared → per-subject | Mean abs δ (pp), shared → per-subject | Sign changes | Identical | Rank correlation |
|---|---|---|---|---|---|---|---|
| Sleep-EDF × CNN | Full TTT | 20 → 20 of 20 | −23.3 → −23.0 | 23.3 → 23.0 | 0 | 7 | +0.96 |
| Sleep-EDF × CNN | Adapter TTT | 12 → 11 of 20 | −2.4 → −2.3 | 5.0 → 4.2 | 9 | 0 | +0.30 |
| Siena × CNN | Full TTT | 11 → 11 of 13 | −18.6 → −18.9 | 19.1 → 19.5 | 0 | 5 | +0.94 |
| Siena × CNN | Adapter TTT | 2 → 3 of 13 | +0.2 → +0.1 | 0.3 → 0.2 | 0 | 6 | +0.80 |

## S8 Is the individual effect stable across stream constructions?

A single contrast between two constructions cannot separate a stable individual tendency from a property of the protocol. The archive supports a stronger design: every cell was run under four constructions of the same samples — the natural order and the three permutations of the fully shuffled order of §3.4 — with four seeds each, so every subject contributes 16 paired replicates of the same quantity. Writing Δ for the subject's balanced-accuracy change relative to source, this paper decomposes it as μ + u(subject) + v(construction) + (uv) + ε(seed). A large subject variance means individual differences are reproducible, a large construction variance means the effect moves with the organisation of the stream, and a large interaction variance means which subjects gain depends on it. Table 8 reports the share of variance carried by each term.

For `full_ttt` the verdict is unambiguous. The subject share is 0.56 on Sleep-EDF and 0.70 on Siena, the construction and interaction shares are at or below 0.00, and the median subject keeps the same sign in 1.00 and 0.94 of his or her replicates; the cell mean moves by 0.5 and 1.3 pp across constructions, against effects of −23 and −18 pp. Individual differences in how much a subject is harmed are therefore reproducible, and the widespread harm does not depend on how the stream is organised.

For the adapter arm on Siena the decomposition separates two things the cell mean hides. All four construction means lie between +0.09 and +0.33 pp and the median within-subject SD is 1.0 pp; the ordering is not noise, with a construction share of 0.00 and a rank correlation of +0.81, but it orders differences of about a point and the median subject keeps the same sign in only 0.47 of his or her replicates. A reproducible ordering of changes that small supports no statement about who benefits: the arm is inert here under every construction tested.

The adapter arm on Sleep-EDF is the case that matters. Its four construction means are −2.37, +1.18, +1.34 and +0.78 pp: the natural order is the most negative, and the three permutations sit about 3.6 pp higher. The construction share is 0.56 against a subject share of 0.15 and an interaction share of 0.01, so the shift is close to a common offset rather than a reordering, with a mean rank correlation of +0.66 across construction pairs. Because the level sits near zero, that offset alone moves the count of subjects who fall from 12 of 20 to 4 of 20.

Table 9 then separates the two mechanisms that the order manipulation confounds. Interleaving the subjects' samples raises the adapter's cell mean by +3.5 pp when the stream is shared, and moves it by −2.4 pp when each subject is adapted alone. The sign of the interleaving effect therefore depends on whether cross-subject state is shared, which rules out batch composition as a sufficient explanation: a purely compositional effect would carry the same sign in both constructions. The same manipulation leaves `full_ttt` almost untouched (+0.4 pp shared, +0.2 pp isolated) and the Siena adapter arm unchanged.

The consequence reaches back to S7, where cutting the stream at subject boundaries alone leaves that arm's aggregate where it was, but cutting it and interleaving the samples at the same time does not. The level of the adapter arm is therefore a property of the stream construction, not of the rule, and the near-zero mean of §4.4 holds only for contiguous streams. The inertness on Siena survives all four constructions.

**Table 8. Stability of the subject-level change across four stream constructions (pp).** Each subject contributes four constructions (the natural order and three fixed permutations) times four seeds, sixteen paired replicates of the same quantity. The change is decomposed as mean + subject + construction + subject-by-construction + residual; the four middle columns are the share of total variance attributable to each term, and the last three summarise the same replicates from the subject's side. Within-subject SD is the median over subjects of the standard deviation across that subject's sixteen replicates. Sign consistency is the median share of a subject's replicates that carry the sign of that subject's mean. ρ across constructions is the mean Spearman correlation between the six pairs of construction-specific per-subject means. Construction and interaction shares at or near zero mean the effect does not move with the organisation of the stream; a large construction share with a near-zero interaction share means the stream shifts all subjects together.

| Cell | Rule | Mean δ (pp) | Subject | Construction | Interaction | Residual | Within-subject SD (pp) | Sign consistency | ρ across constructions |
|---|---|---|---|---|---|---|---|---|---|
| Sleep-EDF × CNN | Full TTT | −23.0 | 0.56 | 0.00 | 0.00 | 0.44 | 9.3 | 1.00 | +0.98 |
| Sleep-EDF × CNN | TENT | −22.9 | 0.54 | 0.00 | 0.00 | 0.46 | 9.1 | 1.00 | +1.00 |
| Sleep-EDF × CNN | SHOT | −20.6 | 0.50 | 0.00 | 0.00 | 0.50 | 11.4 | 0.91 | +0.97 |
| Sleep-EDF × CNN | Adapter TTT | +0.2 | 0.15 | 0.56 | 0.01 | 0.27 | 4.8 | 0.62 | +0.66 |
| Siena × CNN | Full TTT | −17.7 | 0.70 | 0.00 | 0.00 | 0.30 | 11.3 | 0.94 | +0.93 |
| Siena × CNN | TENT | −14.9 | 0.70 | 0.01 | 0.00 | 0.28 | 10.0 | 1.00 | +0.93 |
| Siena × CNN | SHOT | −9.3 | 0.47 | 0.00 | 0.00 | 0.53 | 10.5 | 0.75 | +0.80 |
| Siena × CNN | Adapter TTT | +0.3 | 0.35 | 0.00 | 0.00 | 0.65 | 1.0 | 0.47 | +0.81 |

**Table 9. Cell mean under four constructions (pp).** Contiguous means the subjects' samples arrive in blocks, as in the natural order; interleaved means they are mixed, as in the three permutations. Shared and isolated are the two constructions of S7. Interleaving shift is the interleaved mean minus the contiguous mean within the same construction. The last column gives the number of subjects whose mean change is negative, in the same four conditions. The sign of the interleaving shift differs between the shared and isolated rows for the adapter arm on Sleep-EDF. Because batch composition is present in both constructions, a purely compositional effect would carry one sign; the reversal therefore shows that the interleaving effect depends on the state-sharing condition. The contrast is descriptive: it does not isolate which property of sharing produces the reversal.

| Cell | Rule | Shared, contiguous | Shared, interleaved | Isolated, contiguous | Isolated, interleaved | Interleaving shift, shared | Interleaving shift, isolated | Subjects who fall, shared-contiguous / shared-interleaved / isolated-contiguous / isolated-interleaved |
|---|---|---|---|---|---|---|---|---|
| Sleep-EDF × CNN | Full TTT | −23.3 | −22.9 | −23.0 | −22.8 | +0.4 | +0.2 | 20 / 20 / 20 / 20 |
| Sleep-EDF × CNN | Adapter TTT | −2.4 | +1.1 | −2.3 | −4.7 | +3.5 | −2.4 | 12 / 4 / 11 / 19 |
| Siena × CNN | Full TTT | −18.6 | −17.5 | −18.9 | −17.2 | +1.1 | +1.7 | 11 / 13 / 11 / 13 |
| Siena × CNN | Adapter TTT | +0.2 | +0.3 | +0.1 | +0.2 | +0.1 | +0.1 | 2 / 4 / 3 / 2 |

## S9 Does the harm fall on the subjects who start low?

A deterioration concentrated on low-scoring subjects would be a different finding from one concentrated on high scorers, and the cohort average alone cannot separate them. For each cell, Table 10 gives the rank correlation between a subject's source BA and that subject's own change under `full_ttt`.

The raw correlation is negative in 9 of 11 cells and reaches −0.98, which read alone would say that the better a subject starts, the more that subject loses. That reading does not survive a control: a change cannot exceed the room a subject has, since δ is bounded above by 100 − source BA, so if adaptation pulls every subject to the same level the correlation is −1 whatever the update did to individuals. The informative quantity is therefore the correlation between the source score and the score after adaptation.

That correlation is positive in 9 of 11 cells, between +0.01 and +0.98: the ordering of subjects survives, and the negative raw correlation is a headroom effect. Sleep-EDF × CNN is the clearest case: its raw correlation is −0.98 and its adapted score is 19.7 ± 1.2 pp, the chance level for five stages, so every subject reaches the same floor and the correlation measures the floor rather than a difference in vulnerability. One cell behaves otherwise: on Siena × CNN the correlation with the adapted score is −0.51 and the adapted mean is 49.2 ± 6.7 pp, the only cell in which the raw correlation can be read as a reversal of order.

**Table 10. Source-model quality and the subject-level change under `full_ttt`.** Source BA is the mean over subjects in the cell. ρ(source, δ) is Spearman's correlation between a subject's source BA and that subject's own four-seed mean change; ρ(source, adapted) is the same correlation computed against the adapted score rather than against the change. Adapted BA gives the mean and SD over subjects. A negative ρ(source, δ) together with a near-zero ρ(source, adapted) means that every subject converged to a common level: δ is then that level minus the source score, and the raw correlation is negative by construction. Only a negative ρ(source, adapted) shows that subjects who started higher ended lower.

| Cell | Source BA | ρ(source, δ) | p | ρ(source, adapted) | Adapted BA | n |
|---|---|---|---|---|---|---|
| PTB-XL × CNN | 87.8 | +0.04 | 0.627 | +0.53 | 73.2 ± 43.3 | 198 |
| PTB-XL × EEGNet | 83.6 | −0.20 | 0.005 | +0.78 | 78.9 ± 27.4 | 198 |
| PTB-XL × Mamba | 84.2 | −0.01 | 0.938 | +0.62 | 72.6 ± 36.9 | 198 |
| PTB-XL × MedFormer | 87.3 | −0.46 | <0.001 | +0.01 | 65.0 ± 26.2 | 198 |
| Siena × CNN | 67.8 | −0.89 | <0.001 | −0.51 | 49.2 ± 6.7 | 13 |
| Siena × EEGNet | 63.6 | −0.20 | 0.505 | +0.98 | 63.5 ± 15.6 | 13 |
| Siena × Mamba | 57.3 | +0.04 | 0.901 | +0.90 | 56.4 ± 8.6 | 13 |
| Siena × MedFormer | 78.2 | −0.85 | <0.001 | +0.39 | 48.7 ± 4.1 | 13 |
| Sleep-EDF × CNN | 43.0 | −0.98 | <0.001 | −0.07 | 19.7 ± 1.2 | 20 |
| Sleep-EDF × EEGNet | 54.5 | −0.47 | 0.035 | +0.19 | 29.2 ± 7.3 | 20 |
| Sleep-EDF × Mamba | 49.2 | −0.16 | 0.506 | +0.34 | 31.6 ± 7.8 | 20 |

Two statements follow from this, both of them negative. Source quality does not identify which subjects are harmed, so a risk score for individuals cannot be built from it; and a raw correlation between source score and change should not be reported without the control, because the headroom bound alone produces it. The adapter arm behaves differently again: its correlation with the adapted score is positive in every cell, between +0.56 and +1.00, so that rule shifts subjects while preserving their order — the same picture S8 reaches from the construction side.

## S10 When in the stream the change happens

The analyses above summarise a stream by one number per subject, and that number does not say whether the change was over before the stream had run a fifth of its length or was still accumulating at the end; Figure 4 separates the two cases.

Each stream is cut into fifths by arrival position and each fifth is scored on its own. Position is a proxy for time only under the interleaved arrangements, where every fifth covers all subjects; under the natural order the first fifth can contain as few as four subjects, so it is not used here.

Two shapes appear, and the size of the total drop does not predict which one. In 3 cells the loss is essentially over before the first fifth ends — 0.96 to 0.99 of it falls inside that fifth — and the score then stays at chance. In 4 cells it accumulates, 0.09 to 0.25 inside the first fifth, and the score is still falling at the end. Sleep-EDF makes the contrast sharpest: under CNN the arm loses 23.0 pp with 0.99 of it inside the first fifth, under Mamba 27.6 pp with 0.11 of it there — two encoders on one dataset, 4.6 points apart in magnitude and a factor of 9 apart in when the loss happens.

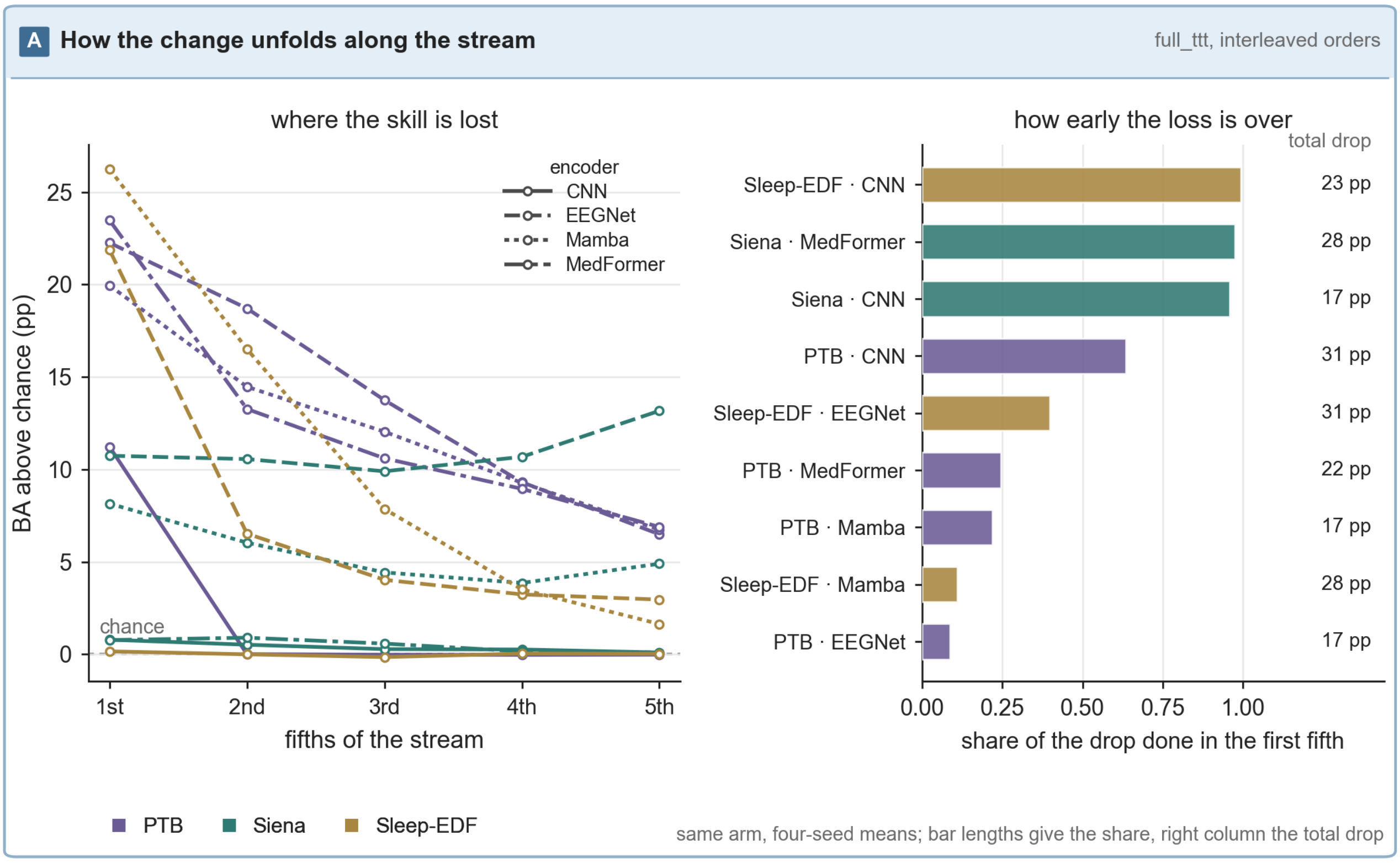

**Figure 4. How the change unfolds along the stream under `full_ttt`.** (a) Balanced accuracy above chance in each fifth of the stream, one line per dataset × encoder cell; horizontal position is arrival order. Only the three interleaved arrangements are used, because under the natural order the first fifth can contain as few as four subjects, which would confound a temporal reading with subject composition. (b) Share of the total drop already complete within the first fifth; bar length gives the share, the right-hand column the total drop in pp. Cells whose score did not fall are absent from (b) rather than plotted at zero, because a share is undefined when there is no drop to divide. Curves are four-seed means. The last fifth is not a terminal-state score: its earlier batches are scored by less adapted models and it covers only the last fifth of the samples (S11).

A within-subject check agrees where the effect is large. Cutting each subject's own samples into early, middle and late thirds and pairing within subject, Siena × CNN moves from +2.5 pp early to −44.9 pp late, with 11 of 13 subjects worse late than early ($p = 0.0225$); Siena × Mamba moves the other way, from −10.0 to +14.7 pp, with 0 of 13 worse late ($p = 0.000244$).

Three limits apply to these trajectories. The last fifth is not a terminal-state score: it is scored batch by batch as each batch arrives, and S11 re-scores the final weights on the whole stream. A recorded test set has no natural length, and the fifths are a division of it. These trajectories also describe only these runs; they do not show that adaptation must proceed in this order.

## S11 Does streaming deterioration persist in the final adapted model?

The streaming score pools post-update predictions (§3.4), so it averages over states: early batches are scored by a barely adapted model and late ones by a model that has absorbed the whole stream, and a run can end worse than that average or better. The terminal-state score asks the question directly, as a retrospective evaluation of the final adapted model on the same unlabeled stream.

All 400 run–arm combinations of the four cells were re-scored from their saved final state. The checkpoint holds the model parameters, the normalisation statistics and, where an arm has one, its adapter. A SHA-256 digest of that state is recorded with the run. Predictions were recomputed from that state in one pass, without gradients and without any update, so every terminal-state score can be re-derived rather than taken from a log. Three checks hold for all of them, and the digest over parameters, buffers and optimiser state is identical immediately before and after the pass, so the pass did not keep adapting; the recomputation runs through each arm's own forward path, so the adapter arm is not silently scored as its source model. `source_only`, whose two scores are equal by construction, returns that equality in all 20 runs of every cell (S2).

For the three shallow rules the two evaluations agree: across the four cells the terminal-state score differs from the streaming score by −1.9 to +3.7 pp, and the largest adapter discrepancy, +2.8 pp on Sleep-EDF × CNN, is smaller than that arm's mean absolute subject-level change in the same cell (5.0 pp, S7). Where the streaming score says a rule barely moved the model, the final state confirms it.

Under `full_ttt` the terminal-state score is 20.0 on Sleep-EDF × CNN, 21.3 on Sleep-EDF × Mamba, 50.0 on Siena × CNN and 50.1 on Siena × MedFormer, against chance levels of 20.0, 20.0, 50.0 and 50.0. In three of the four cells the final model is at chance within 0.1 pp; in Sleep-EDF × Mamba it is 1.3 pp above it. The source model's margin over chance in the same cells is 16.8 to 29.3 pp, and no cell retains it. The deterioration reported in §4.4 is therefore present in the weights each run ends with; it is not an artefact of when the prediction is taken.

The two evaluations agree in three cells and separate in Sleep-EDF × Mamba, and the predicted-class distribution accounts for both. Under `full_ttt` the terminal-state predictions collapse onto one label in three of the four cells (Table 11). A model that returns one label scores 1/K under either evaluation, so in those cells the terminal-state mean is pinned to chance and the two scores can differ only through what the stream still carries before the collapse; the largest such difference is +1.7 pp. On Sleep-EDF × Mamba the largest class holds 91.4% at the terminal state and the model is still moving when the stream ends: the whole-stream score of 31.6 sits 10.4 pp above the terminal-state score of 21.3, so the average overstates what the final weights can do. The last fifth of that stream reads 21.6, within 0.4 pp of the terminal-state score, which is the relation S10 anticipated between the last fifth and a re-evaluation of the final state.

A balanced accuracy at chance does not mean a model that discriminates at chance. When every sample receives the same label, balanced accuracy equals 1/K exactly, so a terminal-state score at chance is the signature of a collapsed predictor rather than a measurement of residual skill, and it locates the collapse at the end of the run rather than in a transient. The streaming score therefore errs in both directions across the four cells — harsher on Siena × MedFormer, 48.4 against 50.1, and gentler on Sleep-EDF × Mamba.

The direction of the change is the same under both evaluations in 76 of the 80 `full_ttt` runs (19, 20, 17 and 20 of 20 by cell, in the order of Table 11), so the sign of the effect does not depend on the evaluation time. The terminal-state score adds the magnitude at the end and the confirmation that the harm persists there. The 20 replicates behind each cell are four seeds and five folds over the same subjects, so the cell means are descriptive and no test is reported; and the re-scoring covers four of the eleven cells, the other seven being reported under the streaming score only.

Concentration is not only an end state, and the archived predictions can be re-read to date it. Cutting each stream into ten equal blocks and asking when the most frequent label first reaches 0.8, and when balanced accuracy first falls 10 pp below the source score in the same block, gives both events in 229 of the 268 (seed, fold) units: concentration is first in 97, in the same block in 58, and the score falls first in 74. That split is not a property of the mechanism, and the comparison is asymmetric by construction, because the drop rule is a margin over the source score while the concentration threshold is absolute. This paper therefore does not claim that concentration precedes the loss of accuracy: at a resolution of a tenth of the stream the two are usually not separable, and where they separate the score more often falls first. S3 reports the counts separately for the two binary tasks and for the six cells where the source model leaves headroom.

**Table 11. Streaming and terminal-state scores for the same runs (%).** Streaming is the archived score, pooling each batch's post-update prediction over the whole stream. Terminal state is the score of the final adapted state, recomputed from the saved checkpoint on the same stream and re-derivable from the stored predictions. Each value is a mean over 4 seeds × 5 folds = 20 runs per cell and arm, which are replicates of the same subjects rather than independent units. Same direction counts runs in which the terminal-state change relative to source carries the sign of the stream change. Most frequent terminal-state class is the share of the run's terminal-state predictions that fall on its single most frequent label; when that share is 100% the balanced accuracy equals 1/K exactly. The `source_only` rows are an internal control: their two scores are equal by construction, and the equality is reproduced in every run.

| Cell | Rule | Chance | Streaming | Terminal state | Terminal state − streaming (pp) | Same direction | Most frequent terminal-state class |
|---|---|---|---|---|---|---|---|
| Sleep-EDF × CNN | Source only | 20.0 | 43.0 | 43.0 | 0.0 | 20 / 20 | 70.8% |
| Sleep-EDF × CNN | TENT | 20.0 | 20.2 | 20.0 | −0.2 | 20 / 20 | 100.0% |
| Sleep-EDF × CNN | SHOT | 20.0 | 20.8 | 20.6 | −0.2 | 18 / 20 | 97.6% |
| Sleep-EDF × CNN | Full TTT | 20.0 | 19.4 | 20.0 | +0.6 | 19 / 20 | 100.0% |
| Sleep-EDF × CNN | Adapter TTT | 20.0 | 40.6 | 43.4 | +2.8 | 13 / 20 | 68.4% |
| Sleep-EDF × Mamba | Source only | 20.0 | 49.3 | 49.3 | 0.0 | 20 / 20 | 30.4% |
| Sleep-EDF × Mamba | TENT | 20.0 | 49.0 | 48.4 | −0.6 | 14 / 20 | 30.9% |
| Sleep-EDF × Mamba | SHOT | 20.0 | 49.0 | 48.7 | −0.3 | 13 / 20 | 27.8% |
| Sleep-EDF × Mamba | Full TTT | 20.0 | 31.6 | 21.3 | −10.4 | 20 / 20 | 91.4% |
| Sleep-EDF × Mamba | Adapter TTT | 20.0 | 48.5 | 48.5 | −0.01 | 18 / 20 | 30.0% |
| Siena × CNN | Source only | 50.0 | 66.8 | 66.8 | 0.0 | 20 / 20 | 76.3% |
| Siena × CNN | TENT | 50.0 | 50.4 | 50.7 | +0.3 | 20 / 20 | 99.0% |
| Siena × CNN | SHOT | 50.0 | 56.9 | 55.0 | −1.9 | 16 / 20 | 94.5% |
| Siena × CNN | Full TTT | 50.0 | 50.0 | 50.0 | −0.01 | 17 / 20 | 100.0% |
| Siena × CNN | Adapter TTT | 50.0 | 67.0 | 67.6 | +0.6 | 17 / 20 | 75.5% |
| Siena × MedFormer | Source only | 50.0 | 78.0 | 78.0 | 0.0 | 20 / 20 | 62.3% |

| Cell | Rule | Chance | Streaming | Terminal state | Terminal state − streaming (pp) | Same direction | Most frequent terminal-state class |
|---|---|---|---|---|---|---|---|
| Siena × MedFormer | TENT | 50.0 | 54.0 | 52.7 | −1.3 | 19 / 20 | 97.1% |
| Siena × MedFormer | SHOT | 50.0 | 63.0 | 66.7 | +3.7 | 16 / 20 | 78.2% |
| Siena × MedFormer | Full TTT | 50.0 | 48.4 | 50.1 | +1.7 | 20 / 20 | 99.9% |
| Siena × MedFormer | Adapter TTT | 50.0 | 77.9 | 77.7 | −0.2 | 16 / 20 | 62.4% |

## S12 Reproducibility, archive scope and deviations

For each record the manifest stores source-file hashes, the run fields, test-subject lists, adaptation seed and evaluation protocol; within a seed the test-subject lists match across orders and the per-fold source-only BA agrees to machine precision.

The source model is shared across orders by construction. Under `--deterministic` the runner caches the trained source model per dataset, encoder, seed and fold, and every order loads the same file, so the four arrangements start from identical weights. Across the 268 (dataset, encoder, seed, fold) groups, source-only BA is identical to the last digit (maximum spread 0.0). This study does not have a hash of the adapted weights for the order runs, which differ across orders by design; the four cells re-scored in S11 are the exception, and each of their final states was saved with a recorded SHA-256 state digest.

Adaptation settings are fixed by the launch script and the code defaults rather than by a per-run snapshot: `full_ttt` uses learning rate 1e-4, one step per batch, batch size 64 and the Adam optimiser, with source-training epochs and learning rate at their defaults, which the source-cache key records as default. This paper therefore does not claim a per-run configuration record. This paper claims that the four orders were launched by the same script with the same arguments and the same cached source weights. S1 gives the timing of a fold and the field list of the manifest.

What the archive supports falls into three levels — recomputable from the package, reproducible as a command but not documented per run, and unrecoverable — and S1 states them, because "reproducible" is used loosely and the three are not interchangeable. The operative limit is the third: the adapted weights of the order runs were not saved, so that adaptation must be re-run, and the four cells re-scored in S11 are the exception. Statistical fields can therefore be recomputed from the archive, but the model executions that produced them cannot be restored from it, and the two claims should not be conflated.

Permutation identity is recorded as a hash rather than as an array, so two runs can be checked for using the same permutation, although rebuilding an array requires the same generator and call sequence (S1). Block-preserving orders exist for 67 of the 268 groups, so that comparison is available only where they were run.

Some early permuted-order runs on Siena, Sleep-EDF and BCI-IV-2a were launched without resetting the adaptation random stream, so its state carried over from the preceding job. Those runs were repeated with the reset enabled, and only the repeated results are used. BCI-IV-2a remains an exception: method and training-file fingerprints differ between the original and permuted sides, so it is retained only as a boundary observation and does not support the primary order comparison.

## S13 Per-comparison estimates under the run restriction

Table 12 lists the eleven comparisons one by one, because the medians in §4.3.6 hide how differently the restriction acts on them and because a reader cannot otherwise tell which comparisons carry the claim and which do not.

**Table 12. Per-comparison estimates under the run restriction.** Leave-one-subject-out is the out-of-sample increment of §4.3.6 with every remaining subject in the training set; run-disjoint restricts each training set to the subjects that share no run with the held-out one; size-matched draws the same number of training subjects at random from those remaining. The last column drops any one training subject from every fold that contains it and reports the smallest and largest resulting estimate; it is a stability check and not a confidence interval, because the restricted training set is a given set of six to eleven subjects rather than a draw from a population, so no sampling distribution for the restricted estimate exists in this design. Restricted p is a within-subject Monte Carlo permutation of the encoder labels over 10,000 draws and is uncorrected; its null is exchangeability of the contrast across encoders within a subject, which is stronger than the incremental question because it also breaks the pairing between contrast and aggregate change. Holm corrects it across the eleven comparisons, which are not independent, since ten of them ask about the same twenty subjects. The difference between the first and third columns is the sum of the two difference columns. The size-matched column is the only one that bears on whether the restriction itself removes the increment, and it carries no sampling theory: it is a reference distribution, not a test. Nor does that column separate the two things the restriction changes, because sharing no run with the held-out subject is also the condition under which every remaining training subject's source model was trained on it.

| Comparison | Leave-one-subject-out | Run-disjoint | Size-matched | Leave-one-subject-out − run-disjoint | Run-disjoint − matched | Restricted p | Holm p | Drop one training subject |
|---|---|---|---|---|---|---|---|---|
| Siena, class 0 vs 1 | +0.308 | +0.085 | +0.252 | +0.224 | −0.167 | 0.0492 | 0.3936 | +0.024 … +0.127 |
| Sleep-EDF, W vs N1 | +0.133 | −0.203 | +0.079 | +0.337 | −0.283 | 0.7600 | 1.0000 | −0.259 … −0.150 |
| Sleep-EDF, W vs N2 | −0.063 | −0.261 | −0.140 | +0.198 | −0.121 | 0.8660 | 1.0000 | −0.311 … −0.232 |

| Comparison | Leave-one-subject-out | Run-disjoint | Size-matched | Leave-one-subject-out − run-disjoint | Run-disjoint − matched | Restricted p | Holm p | Drop one training subject |
|---|---|---|---|---|---|---|---|---|
| Sleep-EDF, W vs N3 | +0.019 | −0.198 | −0.039 | +0.217 | −0.159 | 0.6790 | 1.0000 | −0.255 … −0.159 |
| Sleep-EDF, W vs REM | +0.401 | +0.086 | +0.377 | +0.314 | −0.290 | 0.0614 | 0.4298 | +0.045 … +0.153 |
| Sleep-EDF, N1 vs N2 | +0.331 | +0.123 | +0.296 | +0.208 | −0.173 | 0.0234 | 0.2106 | +0.068 … +0.194 |
| Sleep-EDF, N1 vs N3 | +0.301 | +0.018 | +0.284 | +0.283 | −0.266 | 0.1856 | 1.0000 | −0.020 … +0.085 |
| Sleep-EDF, N1 vs REM | +0.186 | −0.222 | +0.140 | +0.408 | −0.362 | 0.7936 | 1.0000 | −0.267 … −0.154 |
| Sleep-EDF, N2 vs N3 | +0.190 | −0.123 | +0.155 | +0.313 | −0.278 | 0.5524 | 1.0000 | −0.191 … +0.013 |
| Sleep-EDF, N2 vs REM | +0.522 | +0.227 | +0.490 | +0.295 | −0.264 | 0.0042 | 0.0420 | +0.178 … +0.337 |
| Sleep-EDF, N3 vs REM | +0.561 | +0.544 | +0.560 | +0.017 | −0.016 | 0.0001 | 0.0011 | +0.516 … +0.580 |

## S14 Timing the write-back of input-stage parameter displacement

The three results above characterise how adaptation-related change is distributed. A separate question is whether the state that carries a change along the stream can be dated. One prespecified boundary experiment was run to address the narrow version of it. In a five-class sleep-staging cell (Sleep-EDF with Mamba; 20 subjects, five folds of four, one continuous stream per fold), a 2 × 2 design allowed or suppressed the write-back of input-stage parameter displacement in an early and a late window. Suppressed is exact: gradients were still computed, Adam moments still evolved on every batch, and every non-input parameter still moved; only the displacement produced by `opt.step()` on the input-stage parameters was copied back to its pre-step value. This is not an update switch, and it does not test whether input-stage gradients exist or whether the optimiser runs. The window boundary was fixed at a batch edge with the boundary batch excluded from the segmented endpoint, leaving 10 late-stream subjects (two per fold) whose samples belong wholly to one window; chance balanced accuracy is 20%. The subject is the unit of inference, and folds are not counted as additional subjects.

The estimability gate was filed before any confirmatory test. The reference arm, in which write-back was suppressed in both windows, scored 37.28% on the late stream, +17.28 pp above chance and above the prespecified 10 pp margin. Allowing early write-back only scored 25.78%; allowing late write-back only scored 31.18%; the arm allowing both scored 23.19%, only +3.19 pp above chance with 0.818 of its predictions in one class, and was flagged as at the floor. The gate therefore marked the early-carry contrast estimable and the interaction contrast saturated at its end, fixing the single confirmatory family at one test (Holm m = 1).

The carry contrast — late-stream subject-level balanced accuracy with early write-back allowed minus the same quantity with both windows suppressed — was −11.50 pp (95% subject bootstrap CI, −12.65 to −10.39; two-sided exact sign-flip permutation $p = 0.0020$, Holm-adjusted $p = 0.0020$), with the same sign in all 5 folds. An earlier run of the same design under the original partition gave −14.17 pp (CI, −15.98 to −12.56), but its confirmatory verdict was suspended because a prespecified cell-level floor rule triggered, so that estimate is descriptive. Both runs use the same 20 subjects: the second is a new cross-validation partition and refitted source models, not an independent cohort. The two are not pooled, the two identical p-values are simply the smallest two-sided value attainable by full enumeration at $n = 10$, and the second does not retroactively confirm the first. Their intervals overlap over 0.10 pp; with the same subjects and different fitted models that difference has no independent sampling distribution and is recorded as a magnitude, not tested. The interaction contrast was +3.51 pp (descriptive CI, +2.14 to +5.11). Because one of its endpoints sat at chance, it was excluded from the confirmatory family: the value can be read as at least this poor, not as an absence of interaction and not as a tested effect.

A second cell, Sleep-EDF with EEGNet, was prespecified as a descriptive comparator and never assigned a confirmatory family. Its reference margin was +33.91 pp; both contrasts were estimable by the gate, at −27.07 pp (CI, −32.35 to −21.61) and +18.77 pp (CI, +13.43 to +24.52). No confirmatory p-value is reported for this cell. That the first cell was tested and the second was not is a design decision fixed in advance, not a difference in significance; with different source margins and floor positions, a claim that the architectures differ in mechanism is not supported.

Two qualifications belong with that estimate. Implementation checks passed on 9/10 folds (679/680 checks); on fold 4 of the Mamba cell, one construction check comparing non-input parameters between the fully suppressed arm and a frozen-input baseline exceeded its frozen tolerance at 1.04 float32 eps. The criterion was not relaxed, the fold was not removed and the batch was not rerun. Predictions on that fold were bit-identical (0 of 1852 differing) for both endpoint equivalences, and the frozen-input baseline enters no endpoint, so the contrast values are unaffected; nevertheless the implementation cannot be described as fully passing. Second, in the earlier run the arms allowing write-back in both windows and in the late window only produced bit-identical late-stream predictions in all ten folds. This supports only the statement that, when late write-back is allowed, additionally allowing it early leaves late predictions unchanged. It does not support the statement that early write-back has no effect: the arm that allowed it early but not late differs from the arm that allowed it late but not early, and these data cannot distinguish a trajectory pulled back to a common attractor from one that differs while remaining invisible at the prediction level.

This experiment lends support to only a narrow claim. Allowing input-stage parameter displacement to be written back early is associated with a lower late-stream score, and that association reappears under a new partition and new randomisation under a prespecified confirmatory test whose precision is limited by five folds rather than by the ten subjects the family counts. It does not identify which parameter, which normalisation process or which optimiser dynamic carries the association, and it does not establish the association as a mechanism; the prespecified replication instead lends support only to the narrower claim that the association recurs under a new partition and a new randomisation. The full set of contrasts is archived in `v2_expB_synthesis.json`.

One further contrast bears on the carrier question and is reported as descriptive. Exchanging the parameter state between trajectories at the window boundary, with everything else held fixed, reproduced most of the carry contrast (−11.38 pp on the late stream); exchanging the remaining optimiser state moved the same quantity by −0.10 pp (95% CI −0.24 to +0.02). The decomposition is not treated as confirmatory: its admissibility criterion — the scope of a bit-identity check — was resolved only after the run (whole stream 10 of 20 subjects, from the exchange point 20 of 20, and the whole-stream reading is unsatisfiable under any faithful implementation), and ten late-stream subjects set its precision. It is consistent with the parameter state carrying the association and lends descriptive support to that reading; it does not establish a carrier.

How much of that estimate rests on the fold structure can be bounded, and it is the audit that §4.2 and §4.3 apply. The ten late-stream subjects are two per fold, so the design offers five units if the fold is taken as the unit and ten if the subject is. With fold as cluster the one-way intraclass correlation of the paired difference is −0.52 (95% fold-cluster bootstrap, −0.91 to +0.38), and the same sign-flip test applied to the five fold means rather than to the ten subjects gives 5 of 5 folds in one direction at $p = 0.0625$. That value is what the design yields under complete within-fold correlation, and the interval excludes it. Taking the interval's upper bound as the dependence instead, the effective sample size is 7 and the smallest two-sided value attainable is 0.016, against a reported $p = 0.0020$: the reported value stays below that bound, but by less than an order of magnitude, which is why Table 19 marks this endpoint dependence-limited rather than design-defined alone. The estimate itself moves little under the same repartitioning — dropping any one fold leaves the contrast between −11.77 and −11.20 pp with the same sign, and resampling folds rather than subjects gives −12.27 to −10.82 pp against −12.65 to −10.39 pp. The exchange endpoint of the previous paragraph is the same quantity on the same ten subjects and carries the same audit (ICC −0.48, −0.95 to +0.38, the same bound).

## S15 Class-conditional geometry under an input-stage intervention

One cell was instrumented at the representation level: Sleep-EDF with Mamba, 20 subjects in five folds of four, the same cell as S14. For every test sample in both conditions, this paper retained the 64-dimensional vector the classifier head receives — the output of the final normalisation layer, averaged over time. Two adapted models exist per fold: one adapted with labels on that fold's 16 training subjects, and one adapted without labels on the held-out stream; below, *label-adapted* and *stream-adapted* refer to those two. Because the retained vector is exactly the head's input, every statement in this subsection concerns the representation the readout sees, not an earlier layer, and none of it describes the encoder's internal hierarchy.

The first measurement is a ratio of traces over the whole space. With $S_B$ the between-class and $S_W$ the within-class scatter matrix, $D = tr(S_B)/tr(S_W)$ was 1.077 under label adaptation and 0.720 under stream adaptation. The paired within-subject change was −0.357 (95% CI −0.423 to −0.294), negative in 20 of 20 subjects and in 5 of 5 folds, with a two-sided exact sign-flip $p = 1.9 \times 10^{-6}$ ($= 2/2^{20}$), the smallest value full enumeration attains at $n = 20$; the family held one endpoint, Holm $m = 1$. A manipulation check on a down-sampled stream gave a change 0.3% of the observed one, so the drop is not an artefact of the sampling rate. This establishes that the intervention is accompanied by a contraction of class geometry. It does not say which class moved, because a ratio of traces aggregates over all five. The same audit leaves this endpoint intact: with fold as cluster the intraclass correlation of the paired change is −0.09 (95% interval −0.33 to +0.20), the fold means are negative in 5 of 5 folds ($p = 0.0625$ on the same sign-flip test), and at the interval's upper bound the effective sample size is 12, where the smallest attainable two-sided value is $4.9 \times 10^{-4}$.

The decomposition answers that question: between-class scatter fell from 4550 to 2800 (mean change −1750, negative in 19 of 20 subjects) while within-class scatter fell from 4265 to 3922 (−343, 17 of 20), so the ratio falls because the between-class term falls about five times as fast. By class, `N3` accounts for 1289 of the 1750 total, or 74%. That number is easy to misread. `N3` is by an order of magnitude the most spread-out class to begin with — its unweighted centroid distance is 32.9 against 0.9 to 8.6 for the others — and both of its terms roughly halve (centroid distance 32.9 → 17.8, ratio 0.54; per-sample within-class variance 4.49 → 2.31, ratio 0.53), so its own ratio of the two is almost unchanged. What the decomposition shows is the largest class contributing the largest absolute change, not one class becoming disproportionately less separable; two other classes move the other way on within-class scatter (`N1` +13%, `REM` +16%). This paper reports the decomposition and does not attribute the total to a class.

The fixed head then fails unevenly across classes. Per-subject macro-recall fell from 0.4367 to 0.2889, and accuracy from 0.4345 to 0.2800 — two different metrics, reported separately and not interchanged. The concentration of predictions rose from 0.353 to 0.606, and the modal predicted class was `N3` in 20 of 20 subjects. Class by class, recall moved from 0.472 to 0.174 (`W`), 0.383 to 0.193 (`N2`), 0.259 to 0.047 (`REM`), 0.099 to 0.033 (`N1`) and 0.971 to 0.998 (`N3`). The only class that improves is the one

the predictions collapse onto; the others lose 7 to 30 points, and 68% of true `W` and 71% of true `N2` are sent to `N3`.

Per-class separability does not track per-class recall, which is itself informative. `N2`'s ratio of centroid distance to within-class variance improves by 15% while its recall halves; `N1`'s ratio falls by 35% while its recall falls by 7 points. Recall is therefore not a function of the class's own separability, which is what a boundary displaced relative to all centroids at once would produce. It is also what a stale output bias would produce, and this design cannot separate the two: a linear head applied to a shifted feature distribution and a linear head with a bias that no longer matches that distribution give the same predictions. S16 separates them by replacing the head; S17 tries to separate them by moving the features.

## S16 A refitted linear readout recovers most of the gap

The retained features allow the readout to be replaced without retraining the encoder. For each fold, this paper fitted a linear readout on that fold's 16 training subjects only, separately for the two conditions, and evaluated it once on the four held-out subjects; the test subjects enter no fit, no hyper-parameter choice and no scaling statistic. The ridge coefficient was selected by an inner leave-one-subject-out loop over the same grid and solver for both conditions; the selected value differed between conditions in four of five folds, so the two were not constrained to a common hyper-parameter. The metric is the per-subject macro-recall used above.

The fixed head's gap between conditions was 0.1478 (0.4367 against 0.2889). With the refitted readout the two conditions scored 0.5048 and 0.4931, a gap of 0.0117. The paired within-subject reduction in that gap was +0.1361 (95% CI +0.1208 to +0.1520), positive in 20 of 20 subjects, two-sided exact sign-flip $p = 1.9 \times 10^{-6}$, one endpoint in the family, Holm m = 1. The same audit leaves it intact as well: intraclass correlation 0.01 (−0.28 to +0.34), direction agreed in all five folds, effective sample size 9 at the interval's upper bound, where the smallest attainable two-sided value is $3.9 \times 10^{-3}$.

The refit does not merely raise a mean. Concentration fell from 0.606 to 0.287, the modal class was `N3` in 2 of 20 subjects instead of 20, and measured against the concentration the class prior itself produces — 0.213 for 500 samples per subject in five equal classes — the stream-adapted predictions moved from 2.85 times the prior to 1.35 times. Recall under stream adaptation moved from 0.174 to 0.453 (`W`), 0.193 to 0.462 (`N2`), 0.047 to 0.419 (`REM`), 0.033 to 0.250 (`N1`) and 0.998 to 0.883 (`N3`). The single-class dominance is therefore largely a property of the readout applied to these features rather than of the features alone. The refit is not uniformly better: under label adaptation `N3` recall falls from 0.971 to 0.873, so the recovery is in the macro-average and not in every class.

Three limits attach to that number: the readout is fitted on labelled training subjects of the target condition, so this is not a label-free procedure and it says nothing about what a deployment could do without labels. "Most" describes the effect size; the protocol fixed no numeric threshold for "close", and 0.0117 is not zero. And a 64-dimensional linear readout fitted on 16 subjects can recover a macro-average without recovering all the information: recall falls after the refit for `N3`, and what is measured is what a linear readout can extract, not what the representation contains.

The feature pipeline was verified in two parts. On the test side it reproduced the earlier run bit for bit. On the training side, sample identity — labels, subject assignment and counts — matched in all ten files, and a bit-level replay of the retained features succeeded for both conditions of fold 1. Folds 2 to 5 were not replayed; their training-side features rest on the pipeline being deterministic rather than on a direct comparison. The coverage is stated here, not a conclusion. S27 crosses the two readouts fitted here with the two feature conditions, separating the feature-condition and readout contributions to that gap.

## S17 An affine alignment of the feature geometry does not add to translation

If the readout fails because the feature distribution moved, moving the features back should help, and how much it helps says which part of the movement matters. This study estimated affine maps from unlabelled statistics only — the mean and covariance of the retained features, per fold, for each condition — and applied them to the stream-adapted features before the fixed head. Five arms were compared: the first applies no map, the second the full second-order map $A = S_{HL}^{-1} S_{LL}$ with $S_c = (\Sigma_c + \lambda I)^{1/2}$, and the third a mean shift with a single isotropic scale. A fourth is a control with the same eigenvalue spectrum as the full map but a random orientation drawn once from a fixed seed, and a fifth is a split-half control that maps one half of the stream-adapted features onto the other half. The ridge term was shared by the two conditions, $\lambda = 10^{-3}\,\mathrm{tr}(\Sigma_{HL})/d$ per fold, so that they could not differ in regularisation strength; with that term the identity $A^{T}(\Sigma_{HL} + \lambda I)A = \Sigma_{target} + \lambda I$ holds to $10^{-14}$, and the residual deviation from the unregularised target covariance is $4 \times 10^{-5}$ in relative Frobenius norm.

Macro-recall was 0.2889 with no map, 0.3983 with the full second-order map, 0.3993 with the mean shift and one scale, 0.2273 with the spectrum-matched random orientation, and 0.2816 with the split-half control. The prespecified contrast — full second-order against mean plus scale — was −0.0010 (95% CI −0.0042 to +0.0022), positive in 9 of 20 subjects, exact p = 0.546, one endpoint, Holm m = 1. Anisotropic second-order alignment therefore adds nothing over a mean shift and a single scale. Clustering does not weaken that reading: resampling folds rather than subjects widens the interval from −0.0042 to +0.0022 to −0.0041 to +0.0033, and the fold-level sign-flip test gives p = 0.625, so the audit leaves the absence of an increment where it was.

The random-orientation arm is what makes the comparison informative rather than merely null. It has the same spectrum and the same complexity as the full map, and it scores below doing nothing: a full-rank linear map applied to these features does not help by virtue of being full-rank, and can hurt.

A post-hoc arm that applies the mean shift alone, with the identity as its linear part, scored 0.3989 — within 0.001 of both the full map and the mean-plus-scale arm. The dose curve, interpolating the target covariance linearly from the stream-adapted to the label-adapted one, rises monotonically (0.2889, 0.3158, 0.3433, 0.3704, 0.3983), but the mean is interpolated on the same parameter, so that curve does not attribute the rise to the second-order term either. Only a translation is recovered. That arm was added after the results were seen, it is reported as a diagnostic, and it enters no family.

For a linear head this is not a neutral operation. Adding a constant vector `δ` to the features adds `Wδ` to every logit, which is exactly a class-bias correction. In this cell the norm of that term is 2.7 to 3.0, its `N3` component is −2.2 to −2.6, while the head's own bias has norm 0.14 to 0.21 and a typical absolute logit is 0.93 to 1.26; a pure translation flips 35% to 40% of predictions. The experiment therefore cannot separate "the feature distribution moved" from "the readout's bias no longer matches the feature distribution", and the second is not a geometric explanation. The ceiling of the whole alignment family is the fixed head's score on label-adapted features, 0.4367, not the 0.5048 a refitted head reaches.

This lends support to only a bounded claim: within affine maps and second-order statistics, in one cell and one randomisation, aligning the geometry beyond a mean shift does not help. It supports only the narrower claim that alignment beyond a mean shift brings no further gain, but it does not establish that geometry is irrelevant: the family is affine, the statistic is second-order, and a single seed was used. Nor does it establish that the refit of S16 works for the same reason — the two interventions move different objects, and only one of them can be written as a change of bias.

## S18 Prediction transitions and what aggregate scores miss

Accuracy and balanced accuracy summarize the net result of prediction changes. Table 13 reports the transitions themselves alongside these summaries for three cells.

**Table 13. Prediction transitions under the original order (% of test samples).**

| Cell | Label changed | Correctness flipped | Net change (subject-equal) | Net change (sample-weighted) |
|---|---|---|---|---|
| Siena × Mamba | 16.31 | 16.31 | −0.90 | +0.52 |
| PTB-XL × CNN | 16.19 | 16.19 | −14.61 | −8.04 |
| Sleep-EDF × CNN | 72.44 | 36.88 | −23.30 | −24.74 |

Figure 5 shows the same three quantities for all eleven cells. For the two binary datasets, the first two columns are equal by construction: any label change also changes correctness. On the five-class Sleep-EDF dataset they differ. With CNN, labels change for 72.4% of samples on average across the four splits (42%–85% per split), but only 36.9% change correctness; the remaining 35.6% (95% subject-cluster CI, 31.6–40.2) move between two incorrect classes. Neither accuracy nor balanced accuracy registers these relabelings, which affect about a third of the test set.

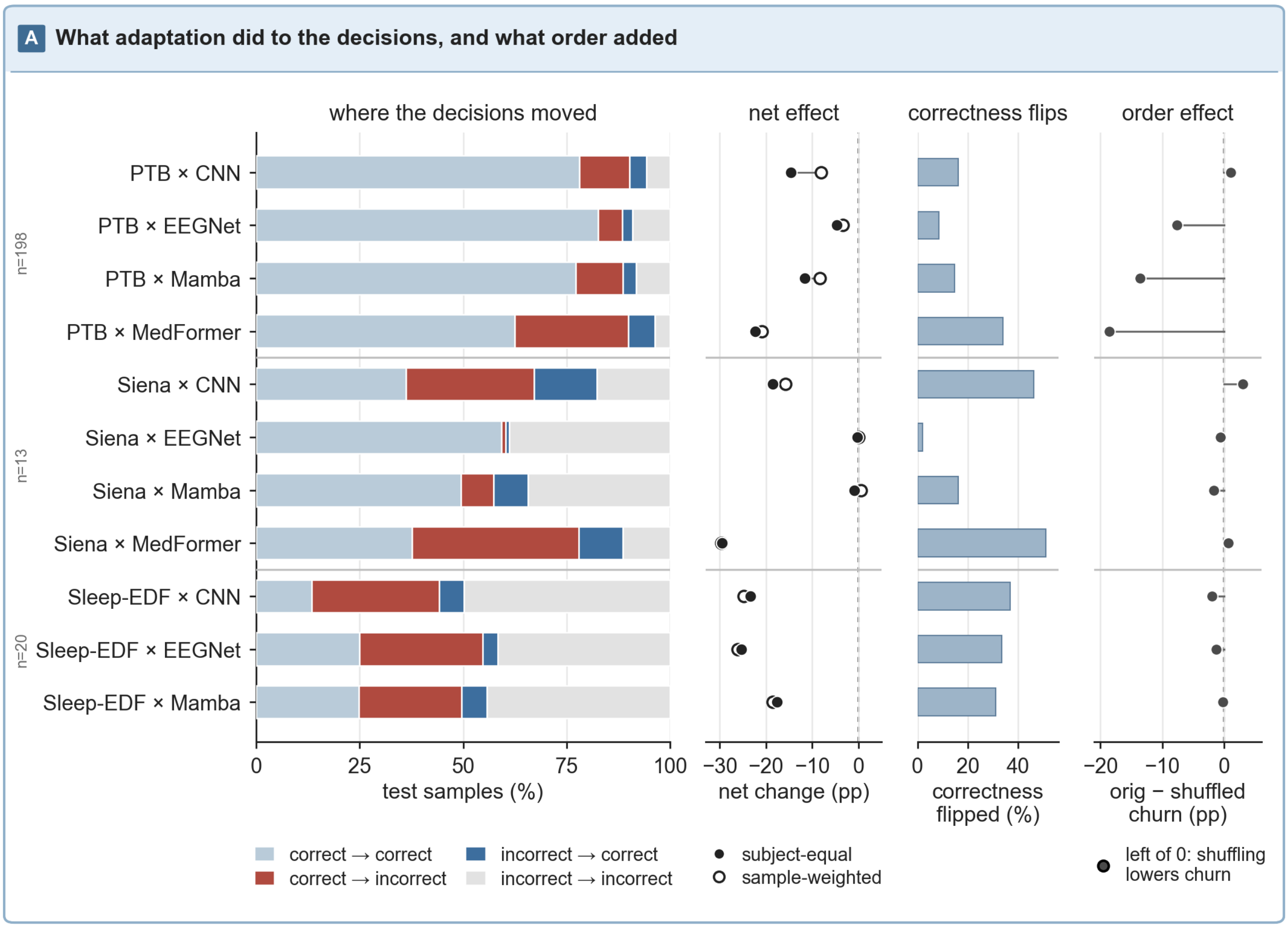

**Figure 5. Three quantities that are often conflated, shown side by side for all eleven cells.** Rows are datasets × backbones. (a) Sample-level composition of the 2 × 2 outcome table: whether a test sample was correct before and after adaptation, as a percentage of all test samples. (b) Net effect: per-subject balanced accuracy after minus before, in pp, reported under two weightings — subject-equal (filled) and sample-weighted (open); they coincide only when every subject contributes the same number of samples. (c) Correctness-flip rate: percentage of test samples whose prediction changed between correct and incorrect. On the two-class datasets this equals the rate at which the predicted label changed; on the five-class dataset labels change more often than correctness does. (d) Order effect on transitions: churn under the original order minus the mean churn under three shuffled orders, pp; negative values mean shuffling reduced churn. Panels (a) and (c) are sample-level; panel (b) is subject-level. No confidence intervals are drawn: the independent unit is the subject, and two of the three datasets have fewer than 21 of them.

Siena shows a different mismatch: about one sixth of samples switch correctness, while the net change is −0.9 points with subject-equal weighting and +0.5 with sample weighting. Neither summary describes the individual flips, and this paper does not calculate a ratio, since the denominator changes sign with the weighting.

Where the flips go is also uneven across classes. On PTB-XL with CNN, class 0 recall falls from 66.5% to 0.08%, and 99.9% of the samples the source model had classified correctly as class 0 are reassigned to class 1. With EEGNet the same quantity falls from 52.5% to 41.9%, and with Mamba from 50.4% to 10.8%. MedFormer goes the other way: class 1 falls from 95.5% to 68.4% while class 0 gains 11.1 points. These are sample-weighted recalls, so they differ from the subject means of Table 16 by 1 to 2 points.

## S19 Sensitivity to test-stream organisation

This paper treats stream organisation as a stress test of the subject-level picture, not as a second main result: the contrasts show that the measured outcome depends on how the stream is arranged, and do not isolate which property of the arrangement does the work. The paired stream-organisation contrast $\delta$ (mean adapted BA across three shuffled arrangements minus adapted BA under the original) is reported as a four-seed mean. On PTB-XL it is +3.12, −3.47, −7.92 and +8.48 pp for CNN, EEGNet, MedFormer and Mamba: two positive and two negative in the same dataset. On Siena the four values are +0.29, +0.71, +2.07 and −2.38 pp, and on Sleep-EDF +0.61, −1.54 and −0.49 pp. No direction summarizes the eleven cells, and the sign is constrained by the headroom the original arrangement leaves.

Figure 6 relates $\delta$ to balanced accuracy under the original arrangement. If the shuffled stream moves a collapsing arm back toward chance, $\delta$ is positive; an arm already at chance has little room to fall. All five cells within 5 points of chance have $\delta \geq 0$, while five of the six cells clearly above chance have $\delta < 0$. The association is descriptive, and when shuffled-side scores vary little across seeds $\delta$ is partly arithmetic.

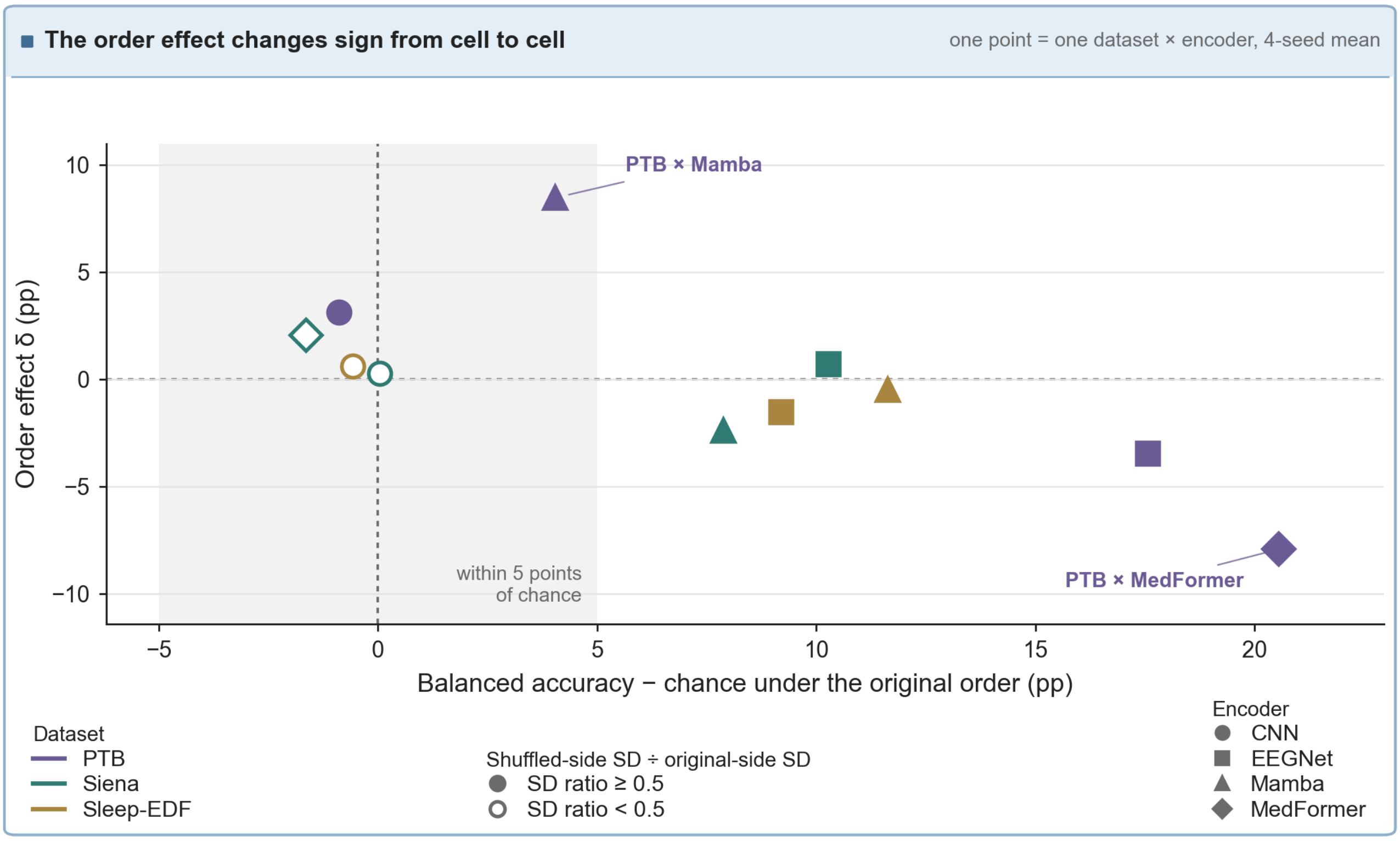

**Figure 6. The stream-organisation contrast depends on source headroom.** One point per core cell. Horizontal axis: balanced accuracy minus chance under the original arrangement. Vertical axis: δ, the mean over three shuffled arrangements minus the original adapted score, in pp; both values are four-split means. The average can hide a wide range: for PTB-XL × MedFormer, the original score ranges from 13.4 to 30.5 points above chance across splits, and δ from −19.1 to +0.2. The shaded region marks cells within 5 points of chance, where little skill remains to lose. Hollow markers identify cells with shuffled-side SD below half the original-side SD; there δ is partly arithmetic because the shuffled score varies little. The independent unit is the subject, not the split, so the plotted values are descriptive contrasts, not confidence intervals. The three stream arrangements also change adjacency, batch composition and boundaries together.

The block-preserving orders can be compared with free permutations across all eleven cells. In eight cells, shuffling within class-coherent blocks changes the result by no more than 0.51 points from the original; free permutation shifts six cells by 1.8 to 9.7 points. Siena with Mamba provides a clear example: the score is 58.4 under the original order, 55.9 after free permutation and 58.2 when the blocks are preserved. Preserving the blocks recovers 92% of the gap.

Three cells show the opposite pattern. For PTB-XL with MedFormer, block-preserving order lowers the score by 12.6 points against 8.2 under free permutation; the same direction appears for PTB-XL with EEGNet (4.8 points) and Siena with CNN (10.3 points), although the latter has the widest fold spread. The differences are described at the level of stream organisation as a whole (§3.4); Figure 7 shows the six arrangements for all eleven cells.

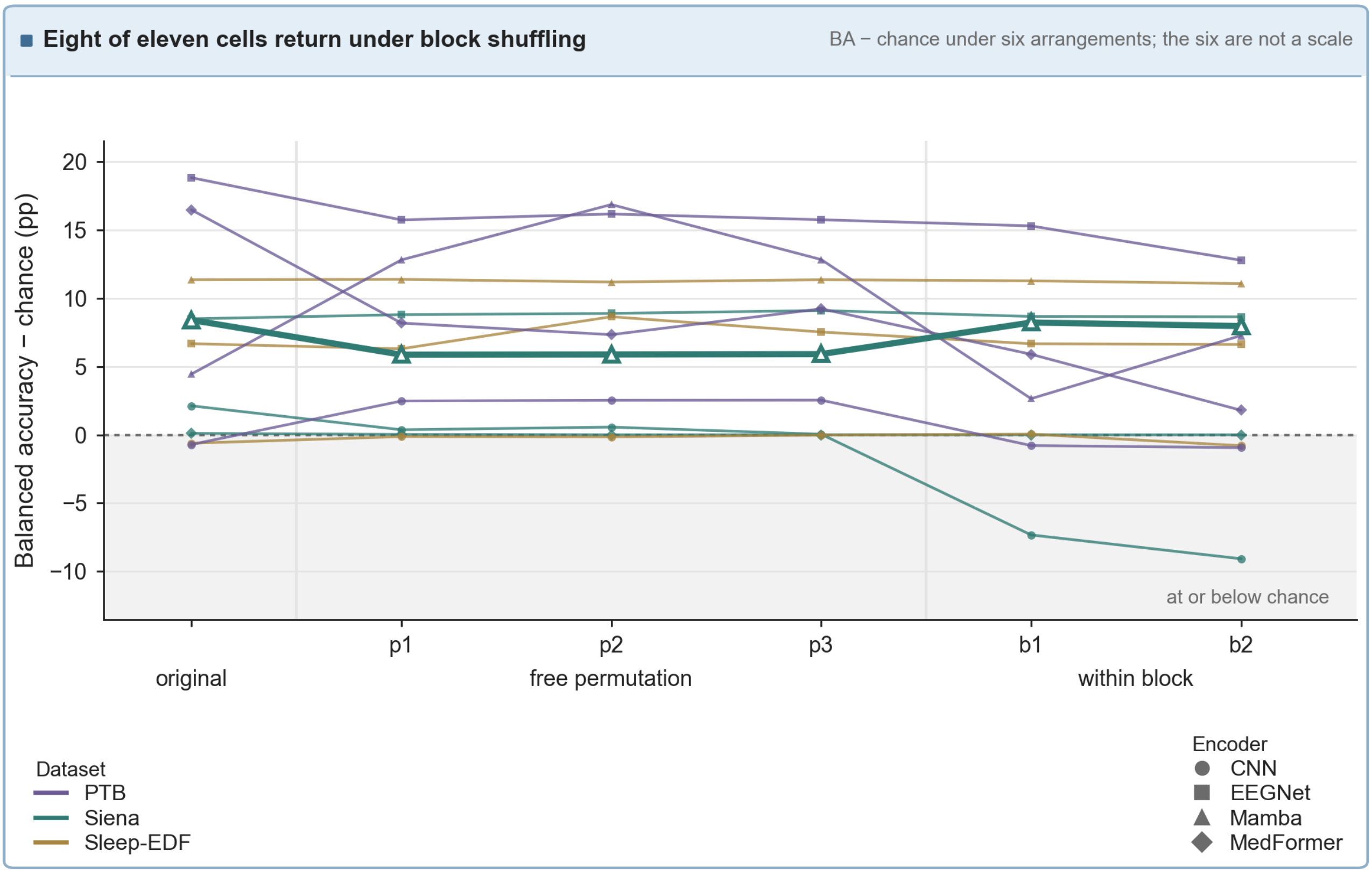

**Figure 7. Two block-preserving arrangements against three free permutations, eleven cells.** Vertical axis is balanced accuracy minus chance, so that datasets with different chance levels share an axis; each point is the mean over the folds of one cell under one arrangement, seed 1, with the same source weights throughout. Colour gives the dataset and marker the encoder; the thick line is the cell used as the example in the text. The three free permutations are draws from one family rather than three replicates — on PTB-XL with Mamba they read 12.8, 16.9 and 12.8 points above chance, so no single permutation is a reliable estimate for that cell — and on Siena the two block-preserving arrangements differ by at most 1.7 points in every cell, so they act as one control. A cell already at chance has little room to fall, so a small move there is not evidence of stability.

Under the original arrangement, batch unanimity is at least 0.92 in five of the sixteen cells, and four remain above 0.75 under all three tested arrangements. PTB-XL × Mamba is the exception: free permutation lowers unanimity from 0.92 to 0.25 while balanced accuracy rises from 54.4 to 62.8. This paper therefore does not conclude that reordering triggers collapse. Read together, these contrasts support one descriptive claim — the measured outcome is sensitive to how the stream is organised — and no causal claim about temporal order.

## S20 Label-free monitoring and prespecified protection strategies

Two label-free quantities are often treated as signs that a model has stopped responding to its input. Neither distinguishes the two subject-level patterns in §4.2: batch unanimity is 0.93 for PTB-XL and 1.00 for Sleep-EDF, despite their different per-subject distributions.

Figure 8 shows what the two readings separate and what they do not. Batch unanimity carries information in one direction: of the sixteen cells in the panel — the eleven above plus four on BCI-IV-2a and one on CHB-MIT — all five with unanimity above 0.90 have balanced accuracy within 4.4 points of chance, yet the best cell, PTB-XL with EEGNet, sits 18.8 points above chance at 0.73 unanimity. High unanimity is consistent with a collapsed model, but low unanimity does not guarantee healthy behaviour: a class-pure batch can produce unanimous predictions even when the model distinguishes classes well. Under the process named in §1 this asymmetry is expected rather than incidental: unanimity reads output concentration, so it can register concentration that has occurred and cannot register a loss of discrimination that has not.

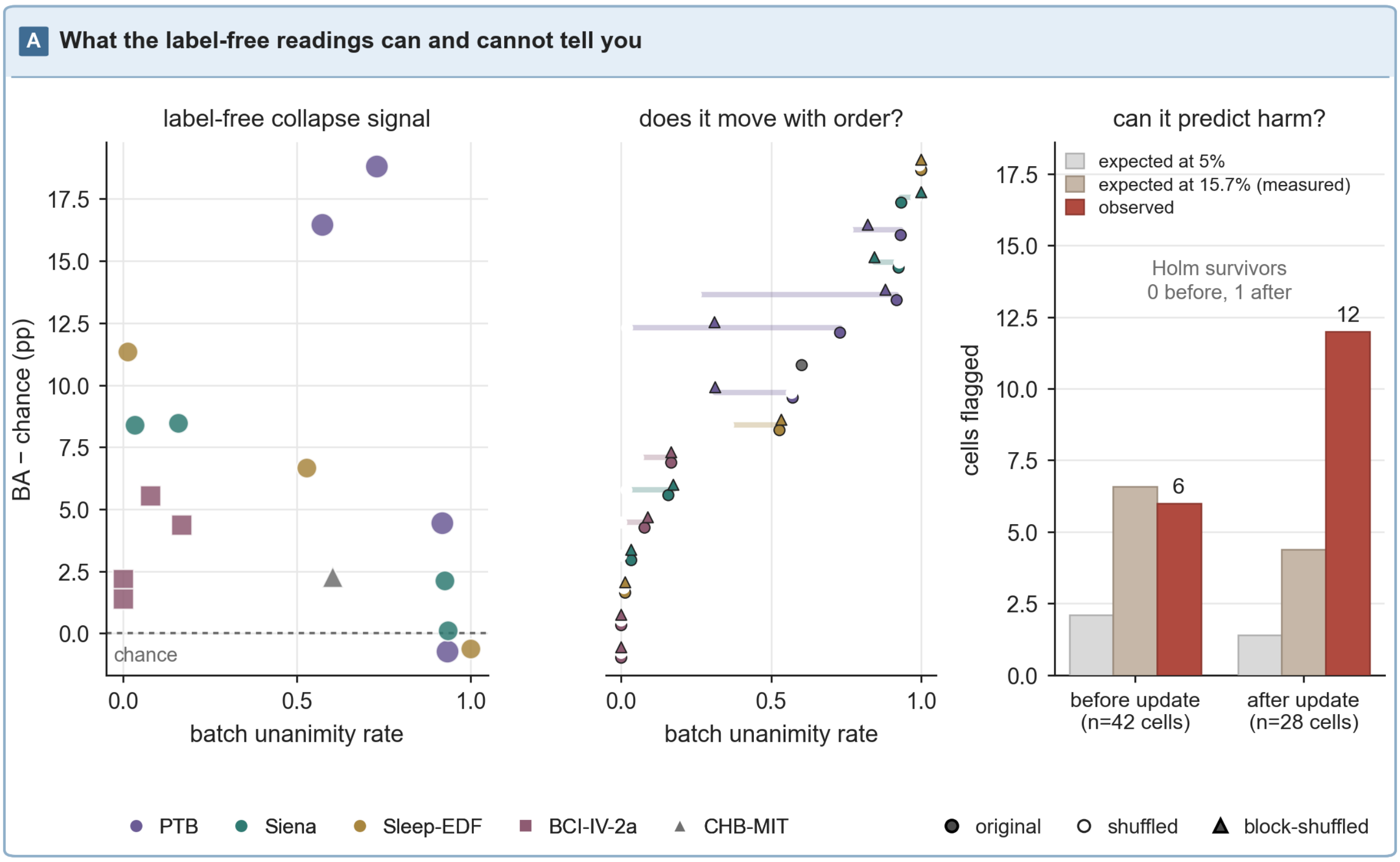

**Figure 8. What can be seen without labels, and what cannot.** (a) Batch unanimity rate (fraction of batches whose predictions are all one label) against balanced accuracy minus chance, under the original order; the size of each marker is the number of folds. A high unanimity rate means the model emits one label per batch. (b) The same rate under three arrangements — original, fully shuffled, block-shuffled — one row per cell, sorted by the original value; rows are not labelled individually, colour gives the dataset. (c) Number of cells flagged by a single label-free risk indicator, against the number expected under two null rates: the nominal 5% and the empirically measured false-positive rate of this pipeline. Holm-corrected survivors are shown in the panel.

The majority-class share has a different limitation: when the test set is imbalanced it can rise regardless of model behaviour, and on PTB-XL it rises after permutation even as balanced accuracy improves. This paper does not recommend either indicator as a detector; it specifies the conditions under which each can be interpreted.

Table 14 lists the criteria each strategy had to meet and how many it met; the reasons differ. The subject-identification strategy required residual signal above evaluation noise, stable individual estimates, a non-empty working range and a positive oracle upper bound, and satisfied none: at 15 and 22 steps residual variance fell below noise variance, so no ceiling could be estimated rather than a small one. Gradient-similarity gating failed three of four criteria, and because its rectified cosine can only shrink the step, it constrains update size as much as it monitors direction. Rollback had no gain to protect (0/6), and on CHB-MIT it exceeded continuous adaptation by 0.0202 while differing from source-only by +0.0000, restoring the baseline rather than preserving a gain. Both indicators measured here are computed from a batch's output distribution. Wang et al. [1] build a different label-free quantity, the disagreement geometry between the source and the adapted prediction, and it ranked per-case harm closely enough to route around most of it in episodic segmentation, where each case resets the model. Whether that quantity carries information in a stream that is never reset has not been tested here. The negative result therefore bounds output-distribution indicators in a continuous unlabelled stream; it does not bound label-free monitoring as a class.

**Table 14. Prespecified protection strategies and their criteria.** The criteria were fixed before the analysis; the table records which were met, not which were added afterwards.

| Strategy | Criteria fixed before the analysis | Met |
|---|---|---|
| Subject identification | residual signal above evaluation noise; stable individual estimates; a non-empty working range; a positive oracle upper bound | 0 of 4 |
| Gradient-similarity gating | four prespecified criteria for directional agreement between consecutive batches | 1 of 4 |
| Rollback | an adaptation gain that is significant in all three orders and transfers across them | 0 of 6 targets |

## S21 Tables and figures moved from the main text

These support the argument without being needed to follow it. Each is cited at the point in the main text where it is used.

**Table 15. What the cited studies vary, over what unit they report, and what they analyse.** *n/a* marks a design in which the encoder is not varied, so a class contrast across encoders is not a quantity the study could report; *not reported* marks a design that does vary the model but reports only at the dataset or model–dataset level. The row for this study is the quantity §4.3 measures. The table covers the studies cited in this section, not the literature: it is not a systematic review, and absence here is not evidence of absence.

| Study | What the experiments vary | Unit of the reported score | What is analysed | Class contrast across encoders |
|---|---|---|---|---|
| Zhao et al. [6], TTAB | Ten adaptation methods, their hyperparameters, source-model quality and shift type, under one protocol | Dataset-level error on benchmark test sets | Ranking of methods and its instability | n/a: encoders not varied |
| Alfarra et al. [7] | Computational time budget and method set | Dataset-level error | Ranking under time constraints | n/a: encoders not varied |
| Wang et al. [1] | Episodic reset per case and a fixed adaptation horizon | Case: ΔDice and harmful accepted area | Per-case harm and a per-case controller | n/a: unit is the case |
| Han et al. [3] | Task sequence in continual adaptation | Dataset-level, after each task | Abrupt collapse during adaptation | n/a: encoders not varied |
| Wang et al. [8] | Distribution alignment on dynamic online streams | Dataset-level | Drift of the predicted distribution | n/a: encoders not varied |
| Kim et al. [9], OATTA | Transitions between consecutive predictions | Batch or stream decision | Whether temporal evidence warrants an update | n/a: encoders not varied |
| Jhawar and Wang [10] | Deployment structure for fixed sample sets | Oracle action per setting | Whether the observation channel can identify the action | n/a: encoders not varied |
| Kozyrskiy et al. [5] | Two pipeline families, four datasets, label-revealed updates | Dataset or stream accuracy per model–dataset pair | Gains from label-revealed updates under a prequential protocol | Not reported: pipelines varied, unit is the pair |
| Lee et al. [4], NeuroAdapt-Bench | Four foundation models × three adaptation methods × five datasets; split and adaptation hyperparameters fixed | Dataset-level ΔBA over five seeds and three batch sizes | Δ per method–model–dataset cell; 180 metric-level comparisons | Not reported: encoders varied, dataset-level only |
| Arasu and Jahangiri [2] | One pipeline on Sleep-EDF Expanded | Dataset-level, with per-stage metrics | Per-stage levels and Cohen's κ | n/a: single pipeline, levels only |
| Schirmer et al. [12] | Monitoring rule and calibration data | Risk over time | Detection delay and false alarms | n/a: encoders not varied |
| This study | Four encoders; cohort, partition, update rule and scoring time held fixed | Subject, and subject × encoder | Direction of the within-subject class contrast, and how much of it the aggregate change accounts for | Reported: sign differs by encoder on both cohorts |

*Discussed in §2.3.*

**Table 16. PTB-XL: per-subject change by the subject's true class (pp).**

| Encoder | Class 0 (n = 51) mean | Class 0 median | Class 1 (n = 147) mean | Class 1 median |
|---|---|---|---|---|
| CNN | −68.1 / −64.1 / −60.9 / −68.7 | −87.1 / −97.2 / −87.8 / −96.4 | +2.3 / +4.0 / +1.7 / +4.3 | 0.00 in all four |
| EEGNet | −15.6 / −5.5 / −6.8 / −15.1 | −11.0 / 0.00 / 0.00 / −8.2 | +0.2 / −6.1 / −5.1 / +0.4 | 0.00 / −0.5 / 0.00 / 0.00 |
| Mamba | −34.5 / −28.4 / −38.2 / −41.1 | −30.1 / −35.0 / −42.6 / −44.1 | −1.9 / −6.7 / −4.1 / −0.3 | 0.00 in all four |
| MedFormer | +13.4 / +2.8 / +2.1 / +22.0 | 0.00 / 0.00 / 0.00 / +2.1 | −40.4 / −37.7 / −25.0 / −31.1 | −45.6 / −28.7 / 0.00 / −6.5 |

*Discussed in §4.3.1.*

**Table 17. Change relative to the source model by adaptation rule (pp, four-seed means).** Each entry is the adapted BA minus the source BA in the same cell, averaged over folds within a seed and then over four seeds; a negative value means the rule left the cell below its source model. The bottom row counts cells below source. The rules differ in learning rate by design (§3.3), so the contrast is between algorithms, not between objectives alone.

| Dataset | Encoder | Source | `full_ttt` | `tent_ttt` | `shot_ttt` | `adapter_ttt` |
|---|---|---|---|---|---|---|
| PTB-XL | CNN | 80.7 | −31.6 | −17.7 | +2.4 | −1.6 |
| PTB-XL | EEGNet | 73.7 | −6.2 | −23.1 | +2.2 | −0.2 |
| PTB-XL | Mamba | 73.6 | −19.6 | −5.7 | +6.0 | +2.6 |
| PTB-XL | MedFormer | 78.9 | −8.3 | −4.4 | +4.4 | +3.4 |
| Siena | CNN | 66.8 | −16.7 | −16.4 | −9.9 | +0.2 |
| Siena | EEGNet | 60.4 | −0.1 | −6.6 | −2.0 | +0.1 |
| Siena | Mamba | 57.2 | +0.6 | +0.2 | −0.02 | +0.7 |
| Siena | MedFormer | 78.0 | −29.6 | −24.0 | −15.0 | −0.1 |
| Sleep-EDF | CNN | 43.0 | −23.6 | −22.8 | −22.2 | −2.4 |
| Sleep-EDF | EEGNet | 54.4 | −25.2 | −33.9 | −15.9 | −7.5 |
| Sleep-EDF | Mamba | 49.3 | −17.6 | −0.3 | −0.2 | −0.7 |
| **Cells below source** | | | **10 / 11** | **10 / 11** | **7 / 11** | **6 / 11** |

*Discussed in §4.4.*

**Table 18. Subject-level change under each adaptation rule (pp).** Each value is the rule's subject-level balanced accuracy minus the source value for the same subject, averaged over the four splits before any test. Worse / equal / better counts subjects whose four-split mean change is negative, exactly zero and positive. Mean abs δ is the average size of the change regardless of sign. Net / gross is the sum of the changes divided by the sum of their absolute values: near ±1 the subjects moved in one direction, near 0 gains and losses cancelled. Intervals are percentile subject-cluster bootstrap intervals from 50,000 resamples of subjects, conditional on these four splits and source models. p is the two-sided exact sign test on the worse/better counts, Holm-corrected across the 44 cell–rule tests of the third family (11 cells × 4 rules), not across the eight rows shown; all 44 are in the reproducibility package. Subjects with a change of exactly zero carry no sign and are excluded from that test. The per-seed tests of §4.1 are a different family applied to a different quantity, so a cell can be significant there and not here; Siena × CNN under `full_ttt` is that case.

| Cell | Rule | n | Worse / equal / better | Median δ (pp) | Mean abs δ (pp) | Net / gross | 95% CI (pp) | p (Holm) |
|---|---|---|---|---|---|---|---|---|
| Sleep-EDF × CNN | Full TTT | 20 | 20 / 0 / 0 | −24.2 | 23.3 | −1.00 | [−26.3, −20.2] | <0.001 |
| Sleep-EDF × CNN | TENT | 20 | 20 / 0 / 0 | −23.0 | 22.7 | −1.00 | [−25.7, −19.9] | <0.001 |
| Sleep-EDF × CNN | SHOT | 20 | 20 / 0 / 0 | −21.5 | 22.1 | −1.00 | [−25.2, −19.1] | <0.001 |
| Sleep-EDF × CNN | Adapter TTT | 20 | 12 / 0 / 8 | −1.6 | 5.0 | −0.48 | [−5.09, +0.14] | 1.000 |
| Siena × CNN | Full TTT | 13 | 11 / 0 / 2 | −23.7 | 19.1 | −0.97 | [−26.0, −11.3] | 0.449 |
| Siena × CNN | TENT | 13 | 12 / 1 / 0 | −16.6 | 17.4 | −1.00 | [−23.0, −11.8] | 0.011 |
| Siena × CNN | SHOT | 13 | 11 / 0 / 2 | −7.8 | 9.9 | −0.92 | [−13.9, −4.4] | 0.449 |
| Siena × CNN | Adapter TTT | 13 | 2 / 4 / 7 | +0.3 | 0.3 | +0.74 | [+0.05, +0.39] | 1.000 |

*Discussed in §4.4.*

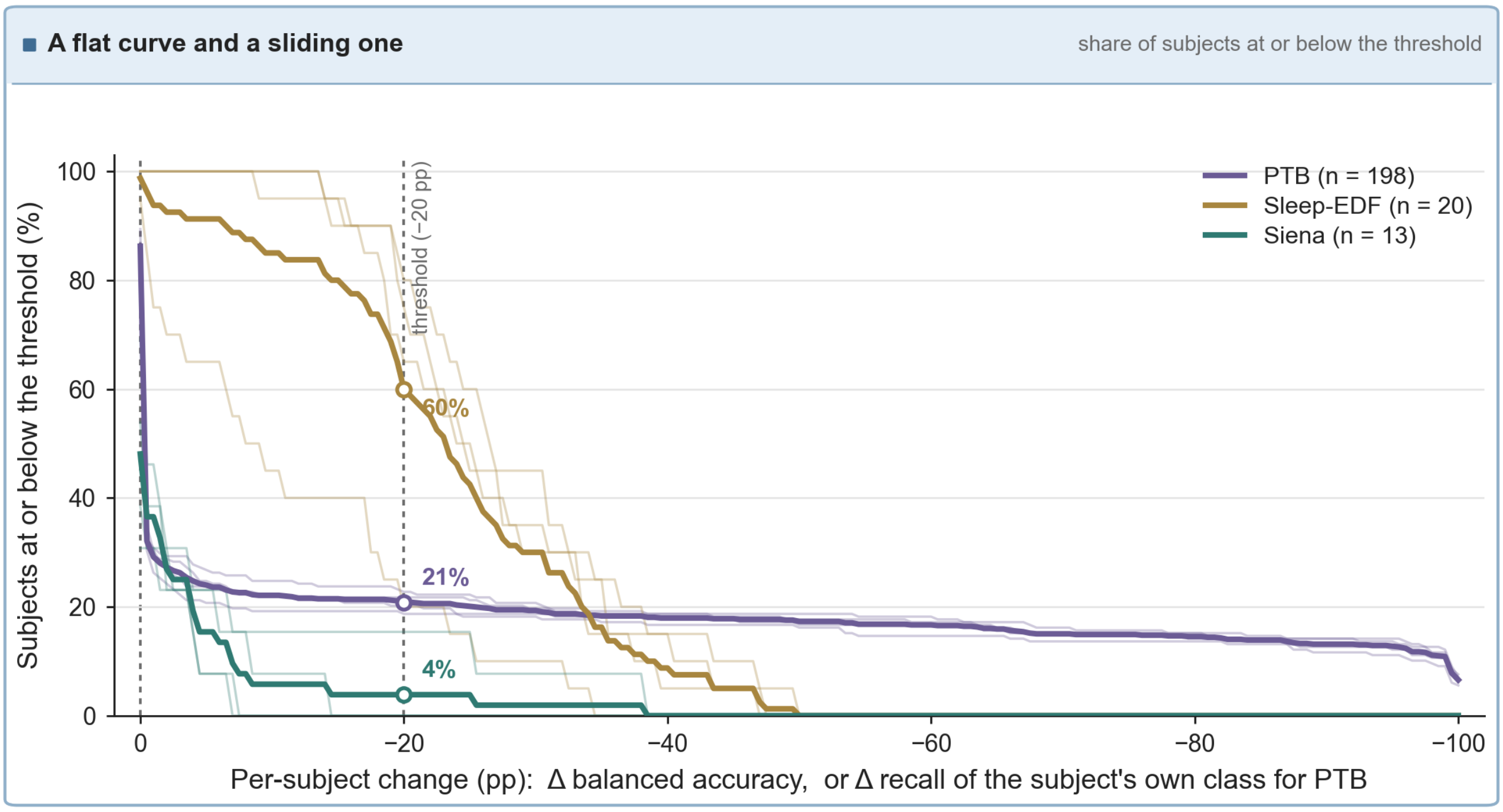

**Figure 9. Cumulative subject shares across deterioration thresholds.** Share of subjects whose per-subject change is at or below the value on the horizontal axis, for PTB-XL with CNN, Sleep-EDF with CNN and Siena with Mamba. Thin lines are the four splits, the thick line their mean; subjects are not pooled across splits. The dashed line marks the threshold of §3.5. The horizontal axis is not the same quantity in all three curves (Figure 2): on PTB-XL it is the change in the recall of the subject's own class, on the other two the change in balanced accuracy, so heights are comparable within a dataset and not across. At zero change the curves read 86.2%, 98.8% and 48.1%; those values count a subject whose change is exactly zero as flagged, which is why they exceed the "share worse" rows of Table 4.

*Discussed in §4.2.*

## S22 Statistical analysis plan

The plan is given in full, not in summary, because the unit of analysis and the correction family differ between analyses, and those differences are what the reporting rests on.

**Table 19. Statistical analysis plan.** No analysis reported here was registered in a public pre-registration. *Design-defined* means the test follows from choices fixed before the runs — the unit, the pairing and the correction family; *exploratory* means the analysis was specified after the results it reports were seen, and it is reported as a description or a sensitivity check rather than as a confirmatory test. *Design-defined, dependence-limited* adds one condition: the dependence audit of S14 raises the smallest attainable two-sided p to within an order of magnitude of the reported one, so that p cannot be read as separated from the dependence bound. The audit covers the five endpoints of S14–S17, and the label is used only where that condition holds. Seeds, folds and stream permutations are not independent subjects: they repartition the same subjects and are never counted as additional units.

| Analysis | Unit | Method | Family | Status |
|---|---|---|---|---|
| Subject-level direction, §4.2 | Subject | Two-sided exact sign test | 11 core cells, within seed | Design-defined |
| Subject-level uncertainty, §4.2 | Subject | Percentile subject-cluster bootstrap, 50,000 resamples | Descriptive | Design-defined |
| Dependence, two levels, §4.2 | Observation; subject | One-way intraclass correlation with fold as cluster; shared-run contrast of subject means with a permutation test; sign test at the effective n implied by the bound; the same audit repeated on the class contrast (§4.3), on the encoder differences (§4.3.3, with an exact run-level sign flip, a run-clustered Wald test and a two-way cluster covariance over subject and run), and on the paired differences of S14–S17 | No correction | Sensitivity |
| Siena class contrast, §4.3 | Subject | Paired within-subject difference; exact sign test against zero, exact sign-flip permutation for pairwise differences | 4 encoders | Design-defined |
| Change by adaptation rule, §4.4 | Subject | Exact sign test | 44 cell–rule tests (11 cells × 4 rules) | Design-defined |
| Stream-construction contrasts, S7–S8 | Subject, paired | Cell-mean differences; variance decomposition | No test | Design-defined |
| Source quality and change, S9 | Subject | Spearman correlation with the adapted-score control | No correction, descriptive | Exploratory |
| Within-stream phases, S10 | Subject | Within-subject early–late contrast; exact sign test | No correction | Exploratory |
| Terminal-state evaluation, S11 | Run (4 seeds × 5 folds) | Paired descriptive difference; no test | No correction | Exploratory |
| Subject jackknife, §4.2 | Subject exclusion | Recomputed estimate and test, 20 and 13 draws | Sensitivity | Exploratory |
| Aggregate change against class contrast, §4.3.6 | Subject × encoder | Nested linear models, $\Delta R^2$ and partial F, Holm across comparisons; leave-one-subject-out refit with the encoder labels permuted within subject by Monte Carlo, 10,000 draws | 11 class-pair comparisons | Exploratory |
| Class-deletion count, §4.1 | Subject-cell, 112 pooled; cut-stream check 13 Siena and 20 Sleep-EDF | Presence of any prediction for a major true class across four seeds; no test | No correction | Exploratory, post hoc |
| Crossed readout interaction, S27 | Subject | Subject-level sign-flip test crossing feature condition and fitted readout | 1 endpoint, Holm m = 3 | Design-defined, resolution-limited |
| Carry contrast of input-stage write-back, S14 | Subject | Paired late-stream difference; two-sided exact sign-flip permutation, one endpoint admitted by a prespecified estimability gate | 1 endpoint, Holm m = 1 | Exploratory, post hoc gate; dependence-limited |
| Representational geometry under an input-stage intervention, S15 | Subject | Paired within-subject Fisher ratio $D = tr(S_B)/tr(S_W)$; exact sign-flip permutation | 1 endpoint, Holm m = 1 | Design-defined |
| Class-level decomposition of that geometry, S15 | Subject × class | Per-class between- and within-class scatter; no test | No correction | Descriptive |
| Fixed-head decoding structure, S15 | Subject | Per-subject macro-recall and confusion structure; no test | No correction | Exploratory |
| Refitted linear readout, S16 | Subject | Cross-subject refit on the training subjects of each fold; paired difference of gaps; exact sign-flip permutation | 1 endpoint, Holm m = 1 | Design-defined |

| Analysis | Unit | Method | Family | Status |
|---|---|---|---|---|
| Affine alignment of the feature geometry, S17 | Subject | Closed-form maps from unlabelled feature statistics; paired difference between alignment arms; exact sign-flip permutation | 1 endpoint, Holm m = 1 | Design-defined |

The Monte Carlo permutation of §3.5 draws 10,000 encoder-label assignments and has an attainable two-sided p-value floor of 1/10001. *Resolution-limited* marks the crossed-readout row: at fold level the five folds give 2^5 = 32 sign assignments, so the smallest attainable two-sided value is 2/32 = 0.0625, and inside the three-contrast Holm family that granularity corresponds to 0.1875. These are resolutions of an enumeration, not a replacement test or a conservative p-value. The S14 row carries an admissibility criterion resolved after the run, and the class-deletion row uses an eligibility rule fixed after an earlier count was found to be contaminated; both statuses are recorded above.

## S23 Robustness of the second-cohort contrasts

§4.3.5 cites four checks on the robustness of these contrasts; they are given here. All were computed on the 21 contrasts of the prespecified family, which contains the six that survive the run-clustered gate.

Resampling subjects gives 95% intervals that exclude zero for 17 of the 21, with $|d_z|$ between 0.15 and 1.79. Removing any single subject never reverses the sign of the mean difference (21 of 21), and no subject contributes more than 20% of the summed absolute difference, so the difference is not carried by one participant. Nor is it an artefact of small class cells: across the 15 contrasts where the correlation is computable, the per-subject difference shows no association with the size of that subject's smaller class (Spearman, none significant).

## S24 Dependence within the shared stream

§4.2 reports that the subject-level result survives the dependence induced by the shared stream; the measurements behind that statement are given here. The question is not whether subjects are independent — they are not — but how much information the archived runs actually carry.

The archive supports one independent unit: the subject. Seeds re-partition the same subjects and retrain the source model on overlapping samples; folds and orders reuse the same samples. Counting fold runs therefore cannot produce independent observations, and the subject-level test is applied within each seed and never pools across them (§3.5), so the repetition of subjects across seeds does not enter the reported p-values. Counting each run once, the three Sleep-EDF cells and Siena × MedFormer have every fold mean negative in each of the four seeds separately (20 of 20), Siena × CNN 19 of 20, and each PTB-XL cell at least 30 of 32; the two Siena cells that are near-null at the subject level are near-null here too (S1). One cell separates the two readings: PTB-XL × Mamba is not significant in any seed (smallest uncorrected p = 0.012) yet all 32 fold means are negative, and it stays non-significant, because a direction count cannot overrule the one test whose unit is independent.

That correlation is observation-level, while the sign test is computed on twenty subject means, and the run term reaches the latter only through the runs two subjects share. The design keeps that number small: folds are redrawn for each seed, so a subject's four changes fall in four different runs, and only 1 of 20 subjects keeps the same fold across all four seeds (1 of 13 on Siena, 2 of 198 on PTB-XL); two subjects share 0.63 of four runs on average on Sleep-EDF and 0.48 on PTB-XL.

This paper tested for that correlation instead of deriving it: for every pair of subjects the analysis formed the product of their centred four-seed means and split the pairs by how many runs they share. On Sleep-EDF × CNN the sharing pairs are not more alike: the correlation is +0.01 over the 94 pairs that share a run against −0.11 over the 96 that share none, and reassigning subjects to folds within each seed, with fold sizes fixed, gives p = 0.23. Ten of the eleven cells give p > 0.10; the exception is PTB-XL × EEGNet (p = 0.016, on a correlation difference of 0.03 that leaves its verdict unchanged).

The most adverse reading the variance decomposition admits treats the whole estimated run variance as a shift shared by every subject in the run. Under it the effective sample size of the Sleep-EDF × CNN sign test is 7.4 rather than 20 and its p falls from $1.9 \times 10^{-6}$ to 0.016, which after Holm across the eleven cells becomes 0.094 rather than $1.9 \times 10^{-5}$. That is the only cell whose adjusted verdict moves: under the same bound the other five cells that survive correction stay at or below 0.002, the largest being Siena × MedFormer, and the five that were not significant remain so. It is reported as the most adverse reading this decomposition admits, not as the estimate: the pairing above does not support it, and with 20 subjects it cannot exclude it either. What the cell supports independently of the correction is the direction: 20 of 20 subjects fell, under each of the three encoders.

Resampling fold runs instead of subjects gives −28.1 to −18.4 pp for that cell and moves the other intervals by at most 2.5 pp. Both audits are recomputable from the archived predictions (`v2_dependence_audit.json`, `v2_dependence_layers.json`).

## S25 PTB-XL: reading Table 16 by split

Table 16, printed in S21, gives the per-subject change by the subject's true class. The reading of it by split, and the group-level qualifications that go with it, are given in full below.

Each PTB-XL subject contributes samples from only one true class. Table 16 therefore compares two class-associated subject groups, not class-specific changes within the same subject; values are shown by encoder and split.

Each cell lists the four splits in seed order; means and medians are computed within each split. Averaging each subject's result across splits and counting those below the harm threshold gives 43 of 51 class-0 subjects under CNN and 15 of 51 under MedFormer; for class 1 the counts are 0 of 147 and 91 of 147.

For CNN, sample-level class-0 recall falls from 66.5% to 0.08%, while class-1 recall changes little, so balanced accuracy falls by 31.6 points against a 14.6-point subject-equal mean; 147 of the 198 subjects belong to the less affected group. Because each subject contributes only one class, the result supports a large difference between the two subject groups but cannot attribute it to class independently of group-level subject characteristics.

The class-associated group contrast also depends on the encoder. MedFormer shows a different pattern: the class-0 group mean is positive in all four splits while class-1 recall falls by 25–40 points. A positive class-0 group mean provides descriptive support for the narrower group-level reading of no overall deterioration in that group, but it does not establish improvement for every subject: averaged per subject across splits, its median is −0.2 pp and 15 of 51 subjects still fall by more than 20 pp. The group-level result is therefore described as no overall deterioration in the class-0 group, not as a class-specific improvement.

The label files do not preserve recoverable clinical semantics, so classes are identified by index only.

## S26 Dependence of the class contrast, and the refit

This section collects the two audits the main text summarises: the dependence of the class contrast, and the refit that asks whether the increment survives a held-out subject.

The class contrast carries its own dependence, and it needed its own audit, because it is not the quantity §4.2 tested. Balanced accuracy is the mean of the per-class recalls, so the change in BA is $(\Delta R_0 + \Delta R_1)/2$ while the contrast is $\Delta R_0 - \Delta R_1$; two such combinations need not share a dependence structure. The observation-level correlation with fold as cluster is 0.713 for CNN, 0.522 for Mamba, 0.198 for EEGNet and 0.748 for MedFormer. Here the shared-run structure is detectable for two of the four encoders: Siena subjects that share a run have more alike contrasts than subjects that share none ($\Delta r = +0.39$, $p = 0.03$ for CNN, 0.03 to 0.05 across permutation seeds; +0.66, $p = 0.002$ for Mamba; −0.16, $p = 0.75$ for EEGNet; +0.32, $p = 0.09$ for MedFormer). The bound sets the effective sample size at 7.0, 8.7, 11.1 and 6.3. Applying it moves the four Holm-adjusted contrasts from 0.010, 0.010, 0.001 and 1.00 to 0.250, 0.035, 0.016 and 1.00, so two survive rather than three. That is a bound and not an estimate: a cluster-robust Wald test on the same contrast, with the run as cluster, gives 0.026, 0.0001, 0.0001 and 0.175 after Holm, so three survive and the CNN contrast is the one whose verdict depends on which of the two readings is used. This paper keeps the reversal, which rests on the ordering of the four means and on Mamba's 13 of 13 subjects, and reports the CNN contrast as a direction with an effect size rather than as a contrast that survives every reading.

**Whether the increment carries to a held-out subject.** This paper refitted both models leaving one subject out — all four splits of that subject leave together — and predicted that subject's within-subject deviations, with the class contrast as y and the aggregate change as x, both centred within subject. Both fits carry an intercept and are scored against the same total sum of squares, so the two $R^2$ values are comparable, and the null is a deviation of zero. The median in-sample increment of 0.34 fell to 0.30, positive in 10 of the 11 comparisons; permuting encoder labels within each subject puts the null median between −0.07 and −0.11 over 10,000 draws, negative because the larger model is penalised out of sample when there is nothing to fit. The observed increment exceeds the 97.5th percentile of that null in 9 of 11 comparisons, and 9 remain below .05 after Holm correction; the two that do not are W vs N2, whose mean contrast here is −0.5 pp, and W vs N3.

The refit withholds the subject from the regression, not from the models that produced the numbers being regressed. Each y and x is a difference between two prediction sets from a source model trained on one fold's complement, so a training subject outside the held-out subject's fold was scored by a model that had it in its training or validation split — 83 to 85% of training rows. The clean rows come only from the one or two (Siena) or three (Sleep-EDF) subjects sharing the held-out subject's fold in a given seed, and none shares that fold in all four, so no refit here is source-model-disjoint (0 of 11 comparisons; `v2_loso_source_overlap.json`).

Restricting the training set to subjects sharing no run with the held-out one, six to eleven per comparison, drops the median from 0.30 to 0.02. The fall is not a sample-size effect: a size-matched random draw gives 0.25, and the restricted value is lower in 11 of 11 comparisons, by 0.26 at the median. But sharing no run means sharing its fold in no seed, which is exactly when a training subject's model was trained on the held-out subject; the restriction removes the same-fold subjects and with them the only clean rows, so the drop and the rise in overlap are one event. Four comparisons keep an uncorrected permutation p below .05 under the restriction, and after Holm correction two remain, N2 vs REM at .042 and N3 vs REM at .0011 (Table 12); both stay positive under every refit, whereas Siena class 0 vs 1 falls to +0.02. For ten of the eleven the increment therefore rests on the training subjects that shared the held-out subject's fold, and those are the only subjects whose models had not seen the held-out subject, so the two properties cannot be separated here; N3 vs REM is the exception, keeping +0.54 of +0.56 once they are removed.

Three further limits keep this from being read as evidence of transfer. The eleven comparisons are not eleven cohorts: ten are the same twenty Sleep-EDF subjects asked about different stage pairs, so the count of positive gains is not a success rate. The permutation null is joint exchangeability across encoders within a subject, which is stronger than the incremental question because it also breaks the pairing between contrast and aggregate change. And the test runs inside the three datasets that produced it, with four implementations and thirteen and twenty subjects (nineteen for three pairs, where one subject's class counts fall below the threshold); it says nothing about a new dataset, centre or device.

**A matched comparison.** The eleven comparisons are between class pairs, so they cannot separate the aggregate from the pair. Subject and pair were held fixed while only the encoder varied; the subjects whose three encoder values span at most 5 pp were kept, and for each of their ten stage pairs the range of the contrast across the three encoders was computed. Five of the twenty Sleep-EDF subjects qualify, giving fifty subject–pair cells, with a median cross-encoder range of 50.5 pp (IQR 39.1 to 74.7) and 94% of the cells reaching 20 pp. The window is not doing the work — at 4 pp two subjects qualify and the median is 49.3 pp, and widening to 20 pp admits sixteen at 50.0 pp — and the subset is not the mildly degraded one, its mean aggregate change being −22.0 pp against −22.1 pp for the other fifteen. Two limits: a subject's ten pairs are five free values rather than ten independent observations, and read per class the range has a median of 36.8 pp (IQR 25.8 to 43.3), smaller and free of that redundancy; and five subjects is five, so these values describe these subjects rather than estimate a population quantity. Siena cannot enter the comparison, since no Siena subject's encoder spread is 10 pp or less (median 29.5 pp), and its unmatched contrast range is a different quantity.

## S27 A crossed readout and a check in held-out subjects

S16 refitted the readout and left a gap of 0.0117 between the two conditions. That comparison changes the feature condition and the readout together. The two analyses collected here separate them by crossing both, and then ask whether the readout-conditioned class-level pattern appears in subjects who took no part in fitting. Both hold the encoder fixed at Mamba in the Sleep-EDF cell, so the quantity is a classifier–feature-condition interaction; §4.3 measures the within-subject class contrast across encoders.

**Crossing the two conditions with the two readouts.** The design uses the same 20 subjects, five folds and 64-dimensional readout input as S15–S17. Two feature conditions, `Z_L` (label-adapted features) and `Z_S` (stream-adapted features), were combined with two linear readouts, `H_L` and `H_S`, each refitted on the labelled training features of its own condition. The diagonal cells A (`H_L` with `Z_L`, 0.5048) and D (`H_S` with `Z_S`, 0.4931) reproduce S16; the crossed cells B (`H_L` with `Z_S`, 0.3882) and C (`H_S` with `Z_L`, 0.3951) are new. Macro-recall is averaged within subject and then over the 20 subjects.

`H_L` scores above `H_S` on `Z_L` (A − C = +0.1097) and below it on `Z_S` (B − D = −0.1049). Writing `Delta_int` for the contrast (A − B) − (C − D), the interaction was +0.2146, positive in 20 of 20 subjects (95% CI +0.1900 to +0.2373). The two marginal contrasts, `M_Z` = −0.0093 and `M_H` = −0.0024, were indistinguishable from zero under the prespecified classification. A separately frozen specification made `Delta_int` its primary endpoint, with a three-contrast Holm family of its own; the subject-level sign-flip test gives Holm-adjusted $p = 5.72 \times 10^{-6}$ under joint sign exchangeability. Under that classification the two readouts rank the two feature conditions in opposite order, while the average feature-condition effect is indistinguishable from zero.

Read at the fold level, the subject-level test has 2^5 = 32 sign assignments, so the smallest attainable two-sided value is 2/32 = 0.0625; inside the three-contrast Holm family that granularity corresponds to 0.1875. These figures describe the resolution of the enumeration, not a replacement test or a conservative p-value. The 20-of-20 direction count and the sign-flip p-value carry the same sign information.

**A check in subjects who entered no fit.** In the first analysis each subject was a test subject in one fold, with 16 training subjects per fold and 12 of those 16 shared between any two folds. The check asks whether the crossover appears when the test subjects have no role in any fit. Of the 52 subjects with E0-PSG recordings in the Sleep-EDF cohort, the 20 development subjects were reserved for fitting and the remaining 32 were screened; four (original identifiers 64, 72, 73 and 74) were excluded because the `N3` class had no rows for them, leaving 28 subjects and 13,018 epochs. `H_L` and `H_S` were fitted on all 20 development subjects, and the 28 subjects were scored under five pairs of donor states and the resulting five pairs of readouts.

The four cell means were A 0.4282, B 0.3761, C 0.3160 and D 0.4268. `Delta_int` was positive in 28 of 28 subjects, with mean +0.1628, and the relative-advantage reversal — A above C and B below D for the same subject — held in 25 of 28 (0.8929). The matched-advantage count, a further endpoint of the frozen reading, was 21 of 28 (0.75). The frozen reading classified this as R1, its criterion being a positive mean together with a reversal proportion of at least 0.80; the reading, the threshold and the four outcome classes were fixed before this cohort was evaluated, and the reading is descriptive. Across the five donor folds the reversal counts were 24, 22, 21, 25 and 22 of 28, with fold-wise interaction means of 0.1837, 0.1484, 0.1624, 0.1581 and 0.1612, so no single fold carries the result. Those five folds reuse the same 28 subjects under five donor-state pairs, and the fold-wise numbers are a descriptive consistency check.

All 28 subjects share the same five donor states and the same five fitted readouts, so uncertainty in fitting those readouts is common to the whole check. The first analysis fitted each readout on 16 training subjects within a fold, whereas this check fitted them on the 20 development subjects; the two interaction means describe different fitting designs, and their four cell values are separate descriptions. The p-value of the first analysis keeps its own specification and its own joint-sign-exchangeability assumption. Both analyses vary the readout and the feature condition at one encoder, leaving the cross-encoder contrast of §4.3 as a different measurement.